\documentclass[11pt]{article}

\usepackage[margin=1in]{geometry}
\usepackage{setspace}
\usepackage[T1]{fontenc}

\usepackage{newtxtext,newtxmath}
\usepackage{bm}

\usepackage{multirow}
\usepackage{longtable}
\usepackage{booktabs}
\usepackage{pifont}
\newcommand{\cmark}{\ding{51}}
\newcommand{\xmark}{\ding{55}}

\usepackage{tabularx}
\usepackage{makecell}
\usepackage{caption}
\usepackage{enumitem}
\usepackage{float}     
\usepackage{placeins}  
\usepackage[dvipsnames]{xcolor}

\usepackage{tikz}
\usetikzlibrary{positioning, fit, arrows.meta}
\usepackage{tcolorbox}
\tcbuselibrary{breakable}  
\tcbset{colback=white,colframe=black!15,colbacktitle=yellow!20,coltitle=black}
\newcounter{eqmbox}[subsection]

\usepackage{authblk}

\usepackage{natbib}
\bibpunct[, ]{(}{)}{;}{a}{,}{,}
\usepackage[colorlinks=true,
            linkcolor=RoyalBlue,
            citecolor=RoyalBlue,
            urlcolor=RoyalBlue,
            filecolor=RoyalBlue]{hyperref}

\newif\ifanon
\anonfalse

\begin{document}

\title{Measuring Judgment Quality in Natural-Language Explanations:\\
       Evidence from Forecasting Tournaments\thanks{Karvetski and Huang are joint first authors. This research was funded by a grant from Open Philanthropy (now known as Coefficient Giving). The views expressed in this paper do not necessarily reflect the views of the Federal Reserve Bank of Chicago or the Federal Reserve System. We thank Jimmy Ba, David Budescu, Yueh-Han Chen, Roger Grosse, Zhijing Jin, Sanmi Koyejo, Hugo Larochelle, Paul Milgrom, Long Phan, Jason Plaks, and Alex Spangher for helpful discussions. Any remaining errors are our own. Corresponding author: Ezra Karger, \href{mailto:ezra.karger@chi.frb.org}{\texttt{ezra.karger@chi.frb.org}}.}}

\ifanon
  \author{}
  \date{June 2026}
\else
  \author[1,2]{Christopher W. Karvetski}
  \author[1,3,4,5]{Sheldon S. Huang}
  \author[1]{Simas Ku\v{c}inskas}
  \author[1]{Nadja Flechner}
  \author[3]{Jingyu Hu}
  \author[1,6]{Philip Tetlock}
  \author[1,7]{Ezra Karger}

  \affil[1]{Forecasting Research Institute}
  \affil[2]{Good Judgment Inc}
  \affil[3]{University of Toronto}
  \affil[4]{Vector Institute for Artificial Intelligence}
  \affil[5]{Stanford University}
  \affil[6]{School of Arts and Sciences \& Wharton, University of Pennsylvania}
  \affil[7]{Federal Reserve Bank of Chicago}

  \date{June 2026}
\fi

\maketitle

\begin{abstract}
\noindent Decision-makers routinely rely on expert judgments accompanied by written explanations, yet explanation quality is difficult to measure at scale. Forecasting tournaments offer a natural testing ground: probabilistic judgments are paired with natural-language rationales and scored against realized outcomes. We introduce \emph{Explanation Quality Markers} (EQMs), a set of sixty theory-guided reasoning patterns scored by large language models (LLMs). In a pre-registered analysis of over 55,000 forecast--rationale pairs from a multiyear forecasting tournament, EQMs predict accuracy at both the forecast and forecaster levels, consistently outperforming pre-LLM text-analysis methods. More than 90\% of statistically significant pattern-level EQM--accuracy correlations match our directional hypotheses. The signal is asymmetric: EQMs identify likely underperformers more reliably than they distinguish the very best forecasters. Benchmarked against traditional indicators of forecasting skill, EQMs are the strongest predictor at the forecast level and competitive at the forecaster level, though weaker than prior accuracy. Human ratings of rationale quality are less consistently correlated with accuracy and place disproportionate weight on rationale length. Results transfer to an independent forecasting study. EQMs provide a scalable, interpretable method for extracting judgment-relevant information from written explanations.


\end{abstract}

\section{Introduction}

Decision-makers routinely receive expert judgments accompanied by written explanations. For example, an intelligence analyst may forecast the likelihood of a geopolitical event alongside a memo outlining the reasoning. An investment committee may review a recommendation backed by a written thesis. A project manager may weigh competing assessments of a project’s feasibility, each supported by a narrative argument. Decision-makers use such explanations, often implicitly, to assess which judgments to trust or act on.

Yet explanation quality is difficult to measure systematically, especially at scale, and therefore little is known about how explanations relate to judgment accuracy. Prior research has identified several hallmarks of sound probabilistic reasoning, including attention to base rates \citep{kahneman2011thinking}, openness to disconfirming evidence \citep{suedfeld1977integrative,suedfeld1992conceptual,baron2008thinking}, and resistance to common cognitive biases \citep{kahneman1973prediction,tversky1974judgment,fischhoff1977confidence,nickerson1998confirmation}. However, measuring whether a given explanation exhibits these features has historically required either labor-intensive expert coding \citep{bakerbrown1992conceptual}, coarse automated proxies \citep{pennebaker2015liwc,boyd2022liwc22}, or measurement tools built around a single theoretical construct \citep{conway2014automated}. Although decision-makers can often tell when an explanation sounds detailed or persuasive, these qualities need not track judgment accuracy itself. In most domains, this problem is hard to study directly because there is no clear ground truth for whether a judgment is good or bad.

Forecasting tournaments offer a natural setting to study this problem empirically. In such tournaments, individuals submit probabilistic forecasts about real-world events, often accompanied by written rationales, and the outcomes are subsequently observed. This makes it possible to compare properties of written explanations with accuracy in a way that is difficult to achieve in other domains \citep{tetlock2005expert,tetlock2014tournaments,tetlock2015superforecasting}. Over the past decade, tournament-based research has produced substantial advances in training forecasters, structuring teams, and aggregating predictions \citep{mellers2014strategies,satopaa2014combining,chang2016developing,chang2017accountability,horowitz2019teams}. However, these advances have mostly relied on numeric data, while the text contained in rationales has received far less systematic attention.

Early efforts to analyze forecasting rationales used conventional natural language processing (NLP) tools, including linguistic dictionaries, readability measures, and sentiment scores \citep{schwartz2017assessing,horowitz2019teams,zong2020measuring}. These studies demonstrated that rationales contain useful information, but the analytical approach was largely atheoretical. A notable exception is \cite{karvetski2021ijf}, which drew on the distinction between inside- and outside-view reasoning \citep{kahneman2011thinking} to build a classifier for comparison-class reasoning; the study showed that this pattern was among the top correlates with forecaster accuracy. Extending such theory-driven measurement to a broader set of reasoning constructs, however, remained infeasible with pre-LLM methods.

Recent advances in large language models (LLMs) change this picture \citep{gilardi2023chatgpt, ziems2024can}. Unlike dictionary-based or feature-engineered approaches, LLMs can interpret the semantic and contextual meaning of unstructured text, making it feasible to score complex reasoning patterns in written explanations at scale \citep{rathje2024gpt}. We use this capability to introduce \emph{Explanation Quality Markers} (EQMs), a set of sixty reasoning patterns drawn from the judgment and decision-making literature and scored by LLMs.

The sixty EQM patterns are organized into six theory-informed families: biases and emotion; analytical reasoning; alignment and discipline; evidential sourcing; integrative reasoning and cognitive framing; and attitude and uncertainty management. Each pattern is defined in natural language and scored on a three-point scale by an LLM (GPT-4o for most of our results).

We test the framework on more than 55,000 forecast--rationale pairs from the Aggregative Contingent Estimation (ACE) tournament, one of the largest collections of probabilistic forecasts paired with written rationales \citep{mellers2014strategies,mellers2015drivers,mellers2015superforecasters}. For each rationale, we map the sixty EQM scores to a single composite score using LASSO and correlate it with accuracy. We benchmark this composite score against one constructed in the same way from the pre-LLM methods discussed above. (Figure~\ref{fig:eqm-pipeline} in Section~\ref{sec:methods} summarizes the LLM scoring pipeline.)

We organize the analysis into four studies. Study 1 tests our core pre-registered hypotheses; Studies 2--4 are extensions that benchmark effect sizes, compare EQMs against human ratings, and test out-of-sample transfer. Study 1 finds that EQM composite scores substantially outperform pre-LLM methods in predicting accuracy at both the forecast ($r = .19$ vs. $r = .06$, $p < .001$) and forecaster ($r = .51$ vs. $r = .39$, $p < .001$) levels. Over 90\% of statistically significant pattern-level EQM–accuracy correlations match our directional hypotheses. The predictive signal is asymmetric: EQMs flag likely underperformers more effectively than they spot top forecasters. Sensitivity checks confirm that the scores are stable across LLMs and robust to prompt variations.

Study 2 benchmarks EQMs against traditional indicators of forecasting skill in a season-over-season design to better understand effect sizes. At the forecast level, EQMs are the strongest available predictor, exceeding even prior-year accuracy. At the forecaster level, prior-year accuracy performs best, but EQMs remain competitive with behavioral indicators such as update size. We also show that forecaster-level EQM scores can be used to improve crowd aggregation.

Study 3 compares EQMs to human ratings of rationale quality collected during the ACE tournament. When tested on the same subsample of the  ACE dataset, human ratings are more weakly related to accuracy than EQMs, both at the forecast level ($r = .07$ vs. $r = .23$) and at the forecaster level ($r = .40$ vs. $r = .50$). In addition, human ratings place disproportionate weight on rationale length: the square root of word count alone correlates with average human ratings at $r=.62$, although rationale length is essentially uncorrelated with forecast-level accuracy.

Finally, Study 4 tests whether the results transfer out of sample. EQM composite scores trained on ACE data carry meaningful signal at both the forecast ($r = .16$ out-of-sample vs. $r = .19$ in ACE) and forecaster ($r = .46$ out-of-sample vs. $r = .51$ in ACE) levels when applied to an independent forecasting dataset collected after the LLM knowledge cutoff. This transfer mitigates the concern that LLM scoring might exploit memorized data. In this out-of-sample setting, human helpfulness ratings are comparable to EQMs at the forecast level but substantially weaker at the forecaster level. The forecast-level comparison thus appears context-dependent, whereas the EQM advantage at the forecaster level is consistent across datasets.

Overall, we make three contributions. First, we introduce EQMs as a theory-guided measurement framework for evaluating written explanations, and we use LLMs to enable scoring at scale. Second, we show that this approach is more predictive of accuracy than pre-LLM methods and human ratings. Third, we demonstrate the practical relevance of using EQMs to screen judgments. The method requires only a single written rationale and no track record.

The remainder of the paper is organized as follows. Section~\ref{sec:methods} describes the data, the EQM pattern set, and the scoring and modeling procedures. Section~\ref{sec:results} presents the four studies. Section~\ref{sec:conclusions} concludes and discusses limitations and potential applications of the EQM framework.

\section{Data and Methods}\label{sec:methods}

We pre-registered the main analysis in Study 1 \citep[our Section~\ref{subsec:hypothesis_testing}]{karvetski2025llm}, specifying data filtering, modeling, and hypothesis-testing procedures for both the EQM and pre-LLM frameworks. The pre-registration allowed for sensitivity, robustness, and exploratory follow-up analyses conditional on EQMs outperforming the pre-LLM methods. The sensitivity and distributional analyses in the remainder of Study 1 (Sections~\ref{subsec:signal_distribution} and \ref{subsec:EQM_sensitivity}) and the supplementary analyses in Studies 2--4 were carried out under this provision. The correspondence between our pre-registration and the paper is detailed in Appendix~\ref{appendix:Pre-reg}. 

\subsection{Data}\label{subsec:data_filtering}

Studies 1--3 use forecast and rationale data collected during the Aggregative Contingent Estimation (ACE) forecasting tournament \citep{mellers2014strategies,tetlock2014tournaments}\footnote{There were originally five competing teams in ACE, but we use only the data from the winning team, the Good Judgment Project (GJP).}, while Study 4 features an independent dataset, which is described in Section~\ref{subsec:team_dynamics}. The ACE geopolitical forecasting tournament was sponsored by the Intelligence Advanced Research Projects Activity (IARPA) and took place from 2011 until 2015, with four seasons of forecasting. The ACE dataset spans a wide range of questions on global political, economic, and security topics, with forecasts produced by a heterogeneous pool of forecasters. It remains one of the most cited judgmental forecasting tournaments that includes forecasts, rationales, human ratings of rationale quality---which feature in Study 3---as well as interventions such as forecaster training, teaming, and the identification and cultivation of superforecasters \citep{tetlock2015superforecasting}. 

In the ACE tournament, geopolitical forecasting questions were posted throughout each season and remained open for weeks or months until resolution \citep{mellers2014strategies}. Some forecasters made forecasts in all four seasons, while others only made forecasts in a single season. Forecasters could submit an initial probability estimate at any point while a question was open, and were permitted to update their forecasts as new information became available. Forecasters were encouraged to provide written rationales explaining their judgments, although submitting a rationale was not mandatory. 

The full ACE dataset contains 424,764 rationales across 498 closed forecasting questions and four seasons. In line with our pre-registered filters, restricting the data to (1) the first rationale provided by each forecaster for each question; (2) those of ten words or more; and (3) associated with binary questions (N = 368) yields a final study sample of 55,463 rationales and corresponding forecasts generated by 3,533 unique forecaster-season combinations. Among these, 1,770 forecaster-season pairs met minimal activity thresholds and thus qualified for forecaster-level analyses described below. More details and support for the filters and thresholds are provided in Appendix~\ref{appendix:FilteringScoring}.

\subsection{Accuracy Measures}\label{subsec:accuracy_measures}

We pre-registered two metrics for measuring forecasting accuracy, corresponding to two different units of analysis: \textit{forecast-level} and \textit{forecaster-level} accuracy. 

The forecast-level metric, \textit{Brier Score Differential} or \ensuremath{BS_{\text{diff}}}, measures how well an individual forecast performs relative to the \textit{local crowd median} at the time the forecast was made\footnote{The \textit{local crowd median} was calculated using the median of the preceding 11 forecasts from different forecasters. See Appendix~\ref{appendix:FilteringScoring} for the details.}. For each binary forecast, the forecaster’s Brier score \citep{brier1950verification} was computed as:
$$
BS = (\text{forecast} - \text{outcome})^2,
$$
where the outcome variable is 0 if the event did not occur and 1 if it did occur. The Brier score of the local crowd median was similarly calculated as:
$$
BS_\text{agg} = (\text{aggregate forecast} - \text{outcome})^2
$$
\ensuremath{BS_{\text{diff}}} is then defined as the difference between these two scores:
$$
BS_{\text{diff}} = BS_\text{agg} - BS.
$$

Positive values of \ensuremath{BS_{\text{diff}}} indicate that the forecaster outperformed the local crowd, while negative values indicate that the forecaster was less accurate than the local crowd. Subtracting the local crowd median helps control for the varying difficulty of different questions without introducing additional complexity.

The second accuracy metric, \textit{Averaged Normalized Brier Score} or \ensuremath{ABS_{\text{norm}}}, captures forecaster-level accuracy: how well a forecaster performed across multiple questions within a season, and was calculated only for forecasters above minimal activity thresholds within a season.\footnote{We compute \ensuremath{ABS_{\text{norm}}} for all forecasters with a single score within a season, but the forecaster-level correlations with \ensuremath{ABS_{\text{norm}}} in Section~\ref{subsec:hypothesis_testing} are calculated using only the subset of forecasters who made forecasts on at least 10 different questions within a season, each accompanied by a rationale of at least 10 words. This pre-registered threshold ensures that the measure reflects sustained forecasting activity rather than isolated contributions. Additional details on the daily scoring, imputation procedure, and normalization are provided in Appendix~\ref{appendix:FilteringScoring}.} For each question, we compute a normalized question-level Brier score, where positive values indicate above-average accuracy relative to other forecasters on the same question. We then average these normalized scores across all questions on which a forecaster participated in a given season. This aggregation reduces the noise inherent in any single realized outcome and approximates how forecasters would rank on the tournament's season-level leaderboard. Additional details on the daily scoring, imputation procedure, and normalization are provided in Appendix~\ref{appendix:FilteringScoring}.

\subsection{Explanation Quality Markers}

\begin{figure}[t!]
\centering
\begin{tikzpicture}[
  font=\footnotesize,
  >={Stealth[length=2.5mm,width=2mm]},
  arrow/.style={->, line width=0.5pt, draw=black!60},
  node distance=4mm,
]
\definecolor{inputBg}{HTML}{F4F2EC}
\definecolor{inputBorder}{HTML}{888780}
\definecolor{llmBg}{HTML}{F0EEFE}
\definecolor{llmBorder}{HTML}{7F77DD}
\definecolor{lassoBg}{HTML}{E8F6F0}
\definecolor{lassoBorder}{HTML}{1D9E75}
\definecolor{valBg}{HTML}{FBF0EB}
\definecolor{valBorder}{HTML}{D85A30}
\definecolor{posGreen}{HTML}{0F6E56}
\definecolor{negRed}{HTML}{A32D2D}
\tikzset{
  bigbox/.style={
    rectangle, rounded corners=3pt, draw, line width=0.4pt,
    inner sep=6pt, align=left, text width=\dimexpr\linewidth-16pt\relax
  },
  halfbox/.style={
    rectangle, rounded corners=3pt, draw, line width=0.4pt,
    inner sep=6pt, align=left, text width=\dimexpr0.46\linewidth-12pt\relax
  },
  patbox/.style={
    rectangle, rounded corners=2pt, draw=llmBorder, fill=white,
    line width=0.3pt, inner sep=4pt, align=left,
    text width=\dimexpr0.50\linewidth-12pt\relax
  },
  patboxEmpty/.style={
    rectangle, rounded corners=2pt, draw=llmBorder, fill=white,
    line width=0.3pt, dashed, inner sep=4pt, align=center,
    text width=\dimexpr0.50\linewidth-12pt\relax
  }
}
\node[bigbox, fill=inputBg, draw=inputBorder] (input) {%
  \textbf{Forecast + rationale}\\[2pt]
  Question: Will six-party talks with North Korea resume before 1 January 2014?\\
  Forecast: 2\%\\
  Rationale: ``\textit{Time is rapidly running to actually have the talks. The US has continued to make discussion of nukes part of the discussion\ldots}''
};
\node[halfbox, fill=llmBg, draw=llmBorder,
      below=6mm of input.south west, anchor=north west] (llm) {%
  \textbf{LLM scoring}\\[2pt]
  60 patterns scored 0/1/2 in a single API call; JSON output:\\[5pt]
  \texttt{\{}\\
  \texttt{\ \ "Statistical\_Reasoning": 0,}\\
  \texttt{\ \ "Fact\_Based": 1,}\\
  \texttt{\ \ "Guessing": 0,}\\
  \texttt{\ \ \ldots 57 more}\\
  \texttt{\}}
};
\node[patbox, anchor=north west] (pat1) at ([xshift=4mm]llm.north east) {%
  \textbf{Statistical Reasoning}\hfill\textcolor{posGreen}{hypothesis: $+$}\\[1pt]
  Looks to past occurrences or data; forms a comparison class and derives a base rate\ldots
};
\node[patbox, below=2mm of pat1.south west, anchor=north west] (pat2) {%
  \textbf{Fact Based}\hfill\textcolor{posGreen}{hypothesis: $+$}\\[1pt]
  Rationale consists mostly of fact-based statements with verifiable details, measurable data, or cited sources
};
\node[patbox, below=2mm of pat2.south west, anchor=north west] (pat3) {%
  \textbf{Guessing}\hfill\textcolor{negRed}{hypothesis: $-$}\\[1pt]
  Forecaster states that the forecast is primarily based on a guess
};
\node[patboxEmpty, below=2mm of pat3.south west, anchor=north west] (pat4) {%
  \textit{57 more patterns}
};
\node[fit=(llm)(pat4), inner sep=0pt] (middle) {};
\node[bigbox, fill=lassoBg, draw=lassoBorder,
      below=6mm of middle.south, anchor=north] (lasso) {%
  \textbf{Composite score}\\[2pt]
  60 EQM pattern scores converted into a single composite score via LASSO
};
\node[bigbox, fill=valBg, draw=valBorder,
      below=6mm of lasso] (val) {%
  \textbf{Correlate with realized accuracy}\\[2pt]
  Evaluated at both the forecast level and the forecaster level
};
\node[below=3mm of val, inner sep=0pt] (spacer) {};
\draw[arrow] (input.south) -- (llm.north);
\draw[arrow] (llm.south) -- (llm.south |- lasso.north);
\draw[arrow] (lasso.south) -- (val.north);
\end{tikzpicture}
\caption{\emph{Overview of the EQM scoring pipeline}. A forecast--rationale pair is scored by an LLM against 60 theory-derived patterns that are defined in natural language, each with a directional hypothesis. The 60 scores are converted into a single composite quality score via LASSO regression and validated against realized forecasting accuracy.}
\label{fig:eqm-pipeline}
\end{figure}
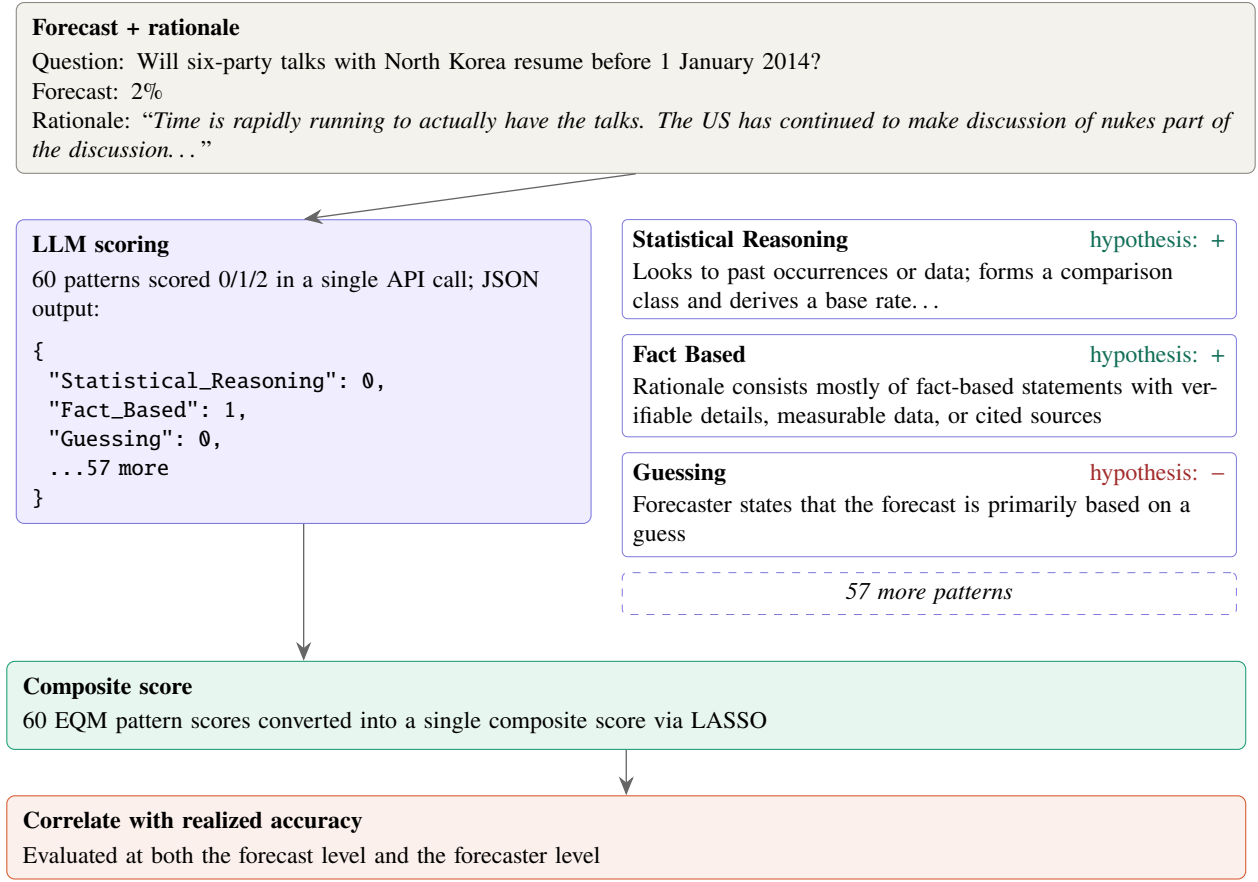

We next describe the full EQM scoring pipeline; see Figure~\ref{fig:eqm-pipeline} for an overview.

\subsubsection{EQM Pattern Set Development}\label{sec:EQMpatterns}

To develop the EQM pattern set, the authors jointly specified a set of sixty patterns with corresponding hypotheses, finalized at the time the pre-registration was submitted. Each pattern consisted of a pattern name, a natural-language pattern description, and a directional hypothesis specifying the expected association with accuracy (positive, negative, or neutral). The goal was to construct a set of candidate markers spanning two broad strands: prescriptive forecasting practices and broader theory-guided constructs. 

The EQM pattern set emerged primarily from three partly overlapping sources. First, we included patterns tied to the forecasting tournament setting, encompassing task structure, scoring, training materials, cognitive batteries, and other design features. These patterns captured behaviors that forecasting research and training link directly to accuracy in tournament settings. Second, we included reasoning behaviors documented in prior forecasting-specific literature, including but not limited to constructs captured by the pre-LLM models applied to forecasting rationales. Third, we included broader constructs from judgment and decision making, psycholinguistics, and cognitive psychology, especially cognitive biases and theories of mental framing and representation whose relationship to forecast accuracy was less direct but theoretically plausible. 

We deliberately kept the patterns content-general rather than topic-specific, so that the same scoring framework could be applied across different forecasting questions and, potentially, to other forms of written analytic judgment. The grouping into six families came after the full pattern set was created. We therefore treat the sixty-pattern set as a first-generation EQM taxonomy rather than as a closed or exhaustive list. Table~\ref{tab:eqm-family-map} lists the full set of EQM pattern names by family, along with the primary motivation for each family; Appendix~\ref{appendix:NSPatterns} provides the full natural-language descriptions, directional hypotheses, and supporting references.

\begin{table}[H]
\centering
\scriptsize
\caption{EQM pattern families and pattern names}
\label{tab:eqm-family-map}
\renewcommand{\arraystretch}{1.15}
\setlength{\tabcolsep}{3.5pt}
\begin{tabularx}{\textwidth}{
  >{\raggedright\arraybackslash}p{1.9cm}
  >{\raggedright\arraybackslash}p{6.35cm}
  X
}
\toprule
\textbf{Family} & \textbf{Primary motivation} & \textbf{Patterns} \\
\midrule

Bias and Emotion
& Draws mainly from broader judgment theory and forecasting research; captures cognitive, affective, and motivational traps that may distort evidence use.
& Clustering Illusion; Confirmation Bias; Extreme Confidence; Speculative Terms; Gut Based; Inside View; Personal Anecdote; Simplification Bias; Positive Emotion; Negative Emotion; Political Preference. \\

Analytical Reasoning
& Draws mainly from prescriptive forecasting practices and forecasting-specific literature; captures structured reasoning used to compare, decompose, explain, and test assumptions.
& Analogies; Best Practices; Causal Reasoning; Domain Expertise; Historic Expertise; Fermi Breakdown; Scenario Reasoning; Second Level Reasoning; Statistical Causal Blend; Statistical Polling; Statistical Reasoning; Statistical Sensitivity Analysis; Wildcards. \\

Alignment and Discipline
& Draws mainly from tournament mechanics, task structure, scoring, and forecaster training; captures forecasting discipline around timing, updating, resolution criteria, and forecast--rationale coherence.
& Forecast and Rationale Align; Forecast and Rationale Misalign; Forecaster Error; Levels of Confidence; Resolution Criteria Deep Dive; Timing; Conditions to Update; Updating; Adjusting Down; Adjusting Up. \\

Evidential Sourcing
& Draws mainly from forecasting practice, tournament design, and evidential assessment; captures how forecasters find, evaluate, and incorporate information and other forecasters' views.
& Authority Based; News or Data Easily Accessible; News or Data Remote; Source at Face Value; Source Credibility and Reliability; Teaming Agree; Teaming Disagree; Teaming Insight. \\

Integrative Reasoning and Cognitive Framing
& Draws mainly from judgment and decision making, psycholinguistics, and cognitive psychology; captures mental framing, representation, attribution, and integration in written explanations.
& Abstract Reasoning; Concrete Reasoning; Descriptive Action Verbs; Interpretive Action Verbs; State Verbs; Situational Factors; Dispositional Factors; Fact Based; FOG Based; Dialectical Reasoning; Elaborative Reasoning; First Person; Third Person. \\

Attitude and Uncertainty Management
& Draws mainly from forecasting practice and administered cognitive batteries; captures epistemic stance, motivation, persistence, and responses to uncertainty.
& Actively Open Minded; Grit; Guessing; Inevitable or Impossible; Wallowing in Uncertainty. \\

\bottomrule
\end{tabularx}

\vspace{0.35em}
\begin{minipage}{0.98\textwidth}
\scriptsize
\textit{Notes.} Rows correspond to the six EQM pattern families used throughout the paper. \emph{Primary motivation}: the main source or rationale for including each family in the EQM pattern set. \emph{Patterns}: EQM pattern names assigned to each family. Appendix~\ref{appendix:NSPatterns} provides the full natural-language descriptions, directional hypotheses, and supporting references.
\end{minipage}
\end{table}

\subsubsection{EQM Example Patterns}\label{sec:EQMexamplepatterns}

One benefit of the EQM framework is that its patterns are defined in natural language. For example, the EQM pattern \textit{Statistical Reasoning}---part of the \textit{Analytical Reasoning} family---had the following pattern description:

\begin{quote}
    {\it The forecaster looks to past occurrences or data and uses logical or statistical reasoning to inform their forecast. This could include forming a comparison class and deriving a base rate, using basic statistics, or building a statistical model.}
\end{quote}
The associated hypothesis for \textit{Statistical Reasoning} was that higher scores would be \textit{positively} associated with forecasting accuracy.

A second example is the pattern \textit{Simplification Bias}---part of the \textit{Bias and Emotion} family---which had the following pattern description:

\begin{quote}
    {\it The forecaster is over-simplifying a complex forecasting dynamic into a simplistic viewpoint.}
\end{quote}
This pattern flagged instances in which the forecaster reduced a multi-factor or uncertain situation into a single-causal or overly certain frame, connecting to long-standing concerns about cognitive miserliness and low-effort processing \citep{stanovich2018miserliness}. The associated hypothesis was that \textit{Simplification Bias} would be \textit{negatively} associated with forecasting accuracy. The full set of pattern descriptions, directional hypotheses, family assignments, and supporting references is provided in Appendix~\ref{appendix:NSPatterns}.

\subsubsection{LLM Scoring}\label{sec:pattern_scoring}

Each EQM pattern was scored on a three-point scale:
\begin{itemize}[leftmargin=2em]
    \item 0 = No evidence of the pattern present in the rationale;
    \item 1 = Some evidence of the pattern present, but incomplete or mixed;
    \item 2 = Clear, unambiguous evidence of the pattern present.
\end{itemize}

The EQM pattern set was scored using OpenAI’s GPT-4o model (\textit{gpt-4o-2024-08-06})\footnote{The EQM methodology is not tied to a specific LLM. In general, any model supporting batch processing and structured JSON outputs should work. We confirm this by replicating the scoring with Google's Gemini 2.5 Pro; see Appendix~\ref{appendix:sensitivity} for details.} with temperature set to zero to reduce output variability \citep{atil2025non}. Each forecast-rationale pair was inserted into a standardized zero-shot prompt along with the full list of pattern names and descriptions and the scoring rubric above; see Appendix~\ref{appendix:NSPrompt} for an example prompt. We scored all 60 EQM patterns in a single API call, without reference to external examples. The co-presence of the full pattern set provides disambiguating context, particularly for conceptually related or contrasting patterns (e.g., \emph{Fact Based} vs. \emph{FOG Based}, the latter denoting \textit{\textbf{F}act-deficient \textbf{O}bscuring \textbf{G}eneralities}) and is substantially more cost-efficient than scoring each pattern in isolation. Each response included the 60 pattern names with scores and an accompanying one-sentence description of the overall reasoning, all of which were returned in structured JSON format.

The LLM was given substantial flexibility in detecting and scoring these patterns within individual rationales. Rather than supplying keyword lists, pre-coded examples, or a detailed code book \citep{stavropoulos2024wisdom}, we relied on the LLM to interpret the natural-language pattern descriptions and infer each pattern's presence through semantic understanding.

Scoring all 55,463 rationales with GPT-4o cost approximately \$400 in API fees (roughly \$0.007 per rationale). Each rationale was scored for all 60 patterns in a single call ($\sim$3,000 input and $\sim$700 output tokens), submitted asynchronously via OpenAI's Batch API. With a current frontier model such as Anthropic's Claude Opus 4.7, the same procedure would cost approximately \$1,000 (roughly \$0.018 per rationale), using batch pricing of \$2.50/\$12.50 per million input/output tokens \citep{anthropic2026opus47}.

\subsection{Pre-LLM Scoring}

Each forecast-rationale pair was also processed with pre-LLM methods to produce a vector with 122 dimensions, in addition to word count. These models included: (1) \textit{Comparison Class Usage}, a token-based Random Forest model developed and validated in \cite{karvetski2021ijf} to detect the presence of outside-view reasoning, further denoted as \textit{CC\_RF}; (2) three automated complexity measures that capture the extent to which forecasters integrate multiple perspectives—\textit{Integrative Complexity} (henceforth denoted \textit{IC}), \textit{Dialectical Complexity} (henceforth denoted \textit{DIAL}), and \textit{Elaborative Complexity} (henceforth denoted \textit{ELAB})—derived from the AutoIC tool \citep{conway2014automated, conway2008twoways}\footnote{\textit{DIAL} captures the integration of divergent perspectives (e.g., discourse markers like "however"), while \textit{ELAB} reflects the integration of reinforcing perspectives (e.g., "in addition"), and \textit{IC} captures a mix of the two.}; and (3) all psycholinguistic variables from the latest (2022) version of LIWC, which includes a broad array of psycholinguistic categories such as emotional tone, cognitive processes, temporal focus, and social orientation \citep{boyd2022liwc22}. In previous comparisons with forecasting accuracy \citep{karvetski2021ijf,zong2020measuring}, the scores from \textit{CC\_RF} and \textit{DIAL} were the strongest predictors of forecaster-level accuracy (i.e., \ensuremath{ABS_{\text{norm}}}), but the other integrative complexity scores as well as LIWC categories also showed significant correlations. Correspondingly, the pre-LLM methods set a high bar for evaluating the EQM patterns.

\subsection{Composite Scores}\label{subsec:predictive_modeling}

Following the pre-registration, we map each forecast–rationale pair's scores (60 EQM patterns or 122 pre-LLM features, plus word count) to a single composite quality score via LASSO \citep{hastie2009elements}. The full training pipeline (data splits, fold construction, and parameter selection) is held identical across frameworks to make sure that any difference in performance reflects predictor signal rather than modeling choices. Full details are provided in Appendix~\ref{sec:composite_scoring}.

Our pre-registered evaluation metric is the Pearson correlation between each composite score and the two accuracy measures (\ensuremath{BS_{\text{diff}}} and \ensuremath{ABS_{\text{norm}}}). We test whether the EQM and pre-LLM correlations differ using a standard test for dependent correlations \citep{steiger1980tests,lee2013corrtest}. One final pre-registration decision was to remove any variable from either the pre-LLM methods or EQM pattern set from the analysis that had the same score on 99\% of rationales, in order to filter out variables with negligible variability and reduce their impact in the model-fitting.

\subsection{Main Hypotheses}\label{subsec:main_hypotheses}

The first part of Study 1 tests the following pre-registered, primary hypotheses:

\begin{enumerate}
    \item The EQM composite scores will predict \textit{forecast-level accuracy} (\ensuremath{BS_{\text{diff}}}) better than the pre-LLM composite scores.
    
    \item The EQM composite scores will predict \textit{forecaster-level accuracy} (\ensuremath{ABS_{\text{norm}}}) better than the pre-LLM composite scores.
    
\end{enumerate}

Per the pre-registration, conditional on EQMs outperforming the pre-LLM methods, we conducted exploratory follow-up analyses, including the sensitivity and distributional analyses in Sections~\ref{subsec:signal_distribution} and \ref{subsec:EQM_sensitivity} and the extensions in Studies 2–4.  The full correspondence between pre-registration and paper is in Appendix~\ref{appendix:Pre-reg}.

\section{Results}\label{sec:results}

\subsection{Study 1: EQMs and Forecasting Accuracy}

\subsubsection{Introduction}\label{subsec:study1intro}

Study 1 tests the central hypothesis of the paper: LLM-scored Explanation Quality Markers predict accuracy better than pre-LLM text-analysis methods. The study proceeds in four steps. We begin with descriptive statistics on rationale length, EQM prevalence, and accuracy in the ACE sample. We then test the two pre-registered hypotheses by comparing how well EQM and pre-LLM frameworks predict accuracy at the forecast and forecaster levels.  Third, we examine if the  EQM signal is asymmetric. Finally, we report sensitivity and reliability analyses. The first two steps were pre-registered; the latter two were exploratory follow-ups conducted under the pre-registered provision for additional analyses conditional on the main hypotheses being supported.

\subsubsection{Descriptive Statistics}\label{subsec:descriptive_stats}

Following the pre-registration, we drop EQM patterns and pre-LLM features that appeared in fewer than 1\% of rationales, removing 10 EQM patterns and 9 pre-LLM features (all from LIWC) from subsequent analysis.\footnote{Removed EQM patterns: \textit{Fermi Breakdown}, \textit{Forecaster Error}, \textit{Grit}, \textit{Levels of Confidence}, \textit{Personal Anecdote}, \textit{Political Preference}, \textit{Positive Emotion}, \textit{Source Credibility and Reliability}, \textit{Statistical Sensitivity Analysis}, and \textit{Teaming Disagree}. Removed LIWC categories: \textit{fatigue}, \textit{filler}, \textit{memory}, \textit{mental}, \textit{nonflu}, \textit{sexual}, \textit{substances}, \textit{swear}, and \textit{wellness}.}

The mean forecast-level accuracy metric, \ensuremath{BS_{\text{diff}}}, was $-0.04$ ($N = 55{,}463$, 95\% CI = [$-0.05$, $-0.04$]). 43\% of forecasts beat the local crowd median, 9\% tied, and 48\% underperformed. The mean forecaster-level accuracy metric, \ensuremath{ABS_{\text{norm}}}, was $0.13$ ($N = 1{,}770$, 95\% CI = [$0.12$, $0.15$]) among forecasters with at least ten questions with rationales of ten words or more.\footnote{The zero point of \ensuremath{ABS_{\text{norm}}} is the average of all forecasters on the same questions, including those without rationales. Forecaster-season observations with at least one qualifying rationale averaged $0.06$ ($N = 3{,}533$, 95\% CI = [$0.05$, $0.08$]), while the main sample with at least ten qualifying rationales averaged $0.13$. This finding suggests a modest positive association between forecasting and rationale-writing activity and standardized accuracy.}

\begin{figure}[t!]
  \centering
  \includegraphics[width=\textwidth]{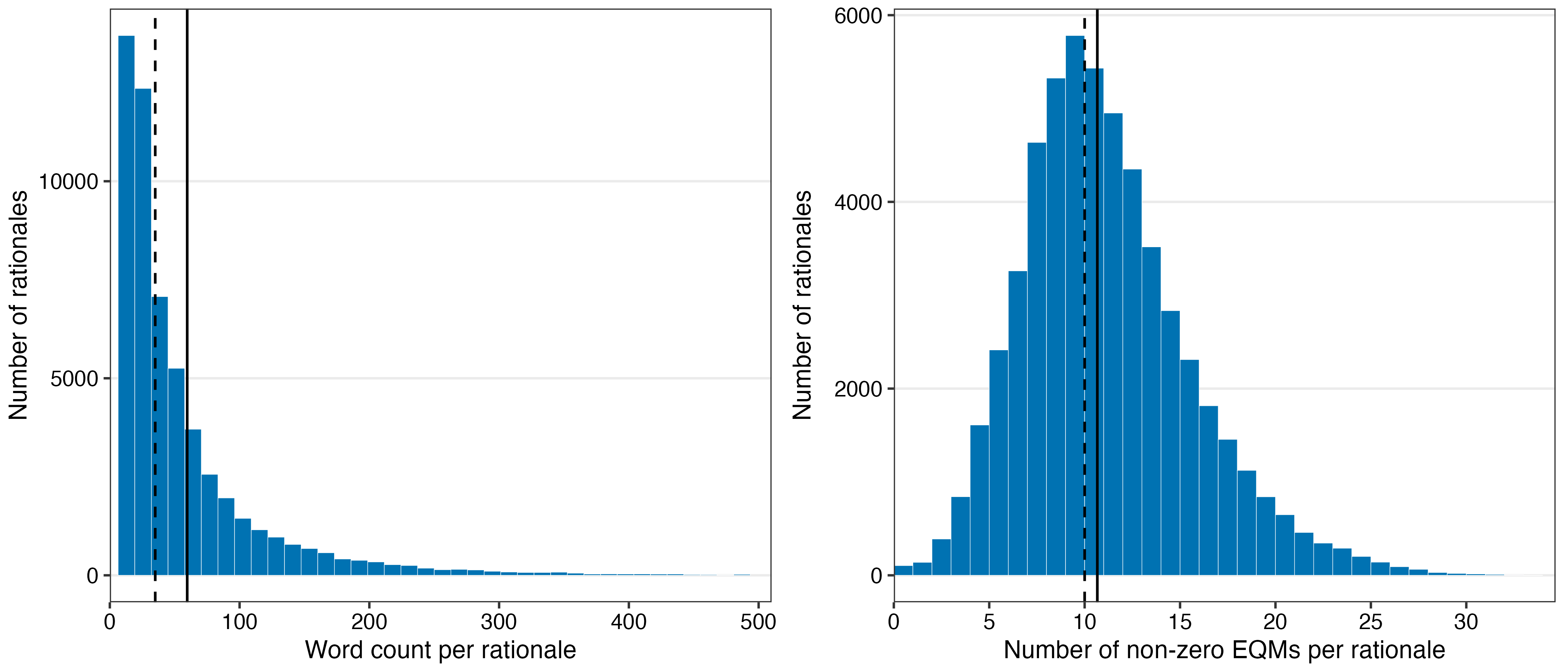}
  \caption{
    \emph{Distributions of rationale length and number of EQM patterns per rationale}. Main ACE analysis sample ($N = 55{,}463$).
    \emph{Left panel}: word count per rationale. \emph{Right panel}: number of EQM patterns with a score greater than zero. Solid vertical lines denote means; dashed vertical lines denote medians.
  }
  \label{fig:hist_wordcount_nonzeroeqm}
\end{figure}

The left panel of Figure~\ref{fig:hist_wordcount_nonzeroeqm} shows the distribution of rationale word counts. Rationales were right-skewed, with a mean of 59.6 words, a median of 35, and a standard deviation of 78.9. Most rationales were concise, though a long right tail reflects a small subset of substantially longer rationales. The right panel of Figure~\ref{fig:hist_wordcount_nonzeroeqm} displays the number of non-zero EQM patterns per rationale. This distribution was more symmetric and concentrated, with a mean of 10.7 EQM patterns per rationale (SD = 4.5, median = 10). Thus, most rationales contained multiple identifiable reasoning patterns rather than a single dominant one.

\begin{figure}[t!]
  \centering
  \includegraphics[width=\textwidth]{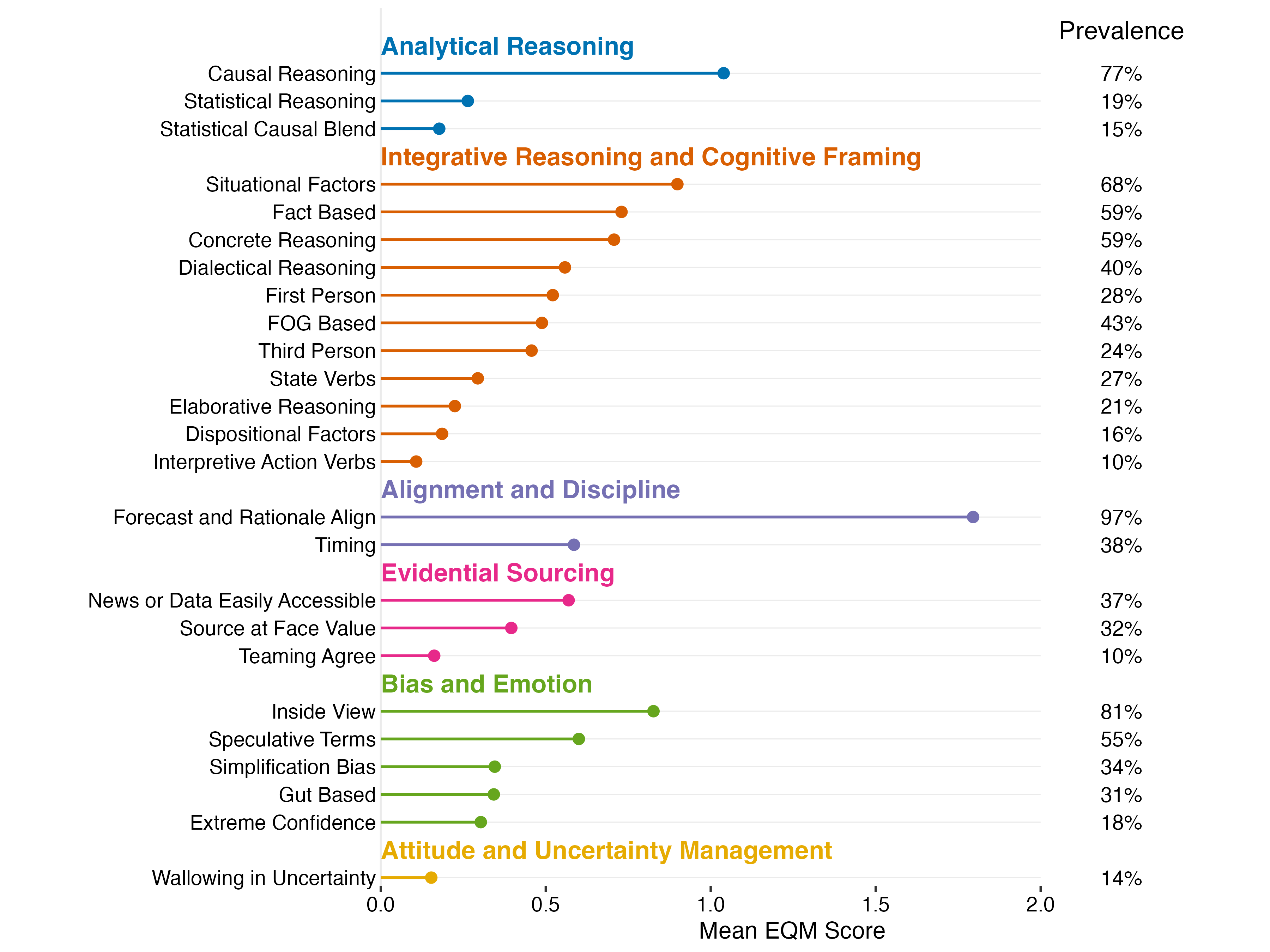}
  \caption{
    \emph{Mean scores and prevalence of high-prevalence EQM patterns}. Main ACE analysis sample ($N = 55{,}463$); patterns with prevalence below 10\% are omitted.
    \emph{Bar length}: mean EQM score across all rationales (including zeros). \emph{Bar color}: EQM family, with family labels shown inline. \emph{Right column labels}: prevalence, defined as the percentage of rationales with a score greater than zero for the given EQM pattern.
  }
  \label{fig:top_eqms_mean_prev}
\end{figure}

Figure~\ref{fig:top_eqms_mean_prev} filters to those EQM patterns that appeared in at least 10\% of rationales, groups them by family, and finally ranks EQM patterns within each family by mean score. Since EQM patterns are scored on a 0--2 scale, mean scores reflect both how often a pattern appears and how strongly it is scored when present. Three observations stand out. First, the \textit{Integrative Reasoning and Cognitive Framing} family contributes the largest number of high-prevalence patterns, suggesting that forecasting rationales commonly involve multiple forms of framing, abstraction, and interpretation. Second, statistical reasoning is rare relative to causal reasoning: \textit{Causal Reasoning} appears in 77\% of rationales, while \textit{Statistical Reasoning} appears in only 19\%. Third, bias-related patterns are also fairly common, indicating that potentially problematic reasoning styles are prevalent in our sample. The most prevalent pattern overall is \textit{Forecast and Rationale Align}, which appears in nearly all rationales.

\FloatBarrier

\subsubsection{Hypothesis Testing}\label{subsec:hypothesis_testing}

Table~\ref{tab:hypothesis_tests} reports the central pre-registered comparison between EQMs and pre-LLM text-analysis methods.

\begin{table}[t!]
\centering
\begin{minipage}{0.90\textwidth}
\centering
\footnotesize
\caption{Main hypothesis tests comparing EQM and pre-LLM composite scores}
\label{tab:hypothesis_tests}
\renewcommand{\arraystretch}{1.1}
\setlength{\tabcolsep}{4pt}
\begin{tabular*}{\textwidth}{@{\extracolsep{\fill}}llccccc@{}}
\toprule
\textbf{Level} & \textbf{Metric} & \textbf{$n$} & \textbf{EQM} & \textbf{Pre-LLM} & \textbf{Between} & \textbf{EQM Better} \\
\midrule
Forecast
& $BS_{\text{diff}}$
& 55{,}463
& $r=.19,\ p<.001$
& $r=.06,\ p<.001$
& $r=.16,\ p<.001$
& $p<.001$ \\

Forecaster
& $ABS_{\text{norm}}$
& 1{,}770
& $r=.51,\ p<.001$
& $r=.39,\ p<.001$
& $r=.69,\ p<.001$
& $p<.001$ \\
\bottomrule
\end{tabular*}

\vspace{0.35em}
\scriptsize
\textit{Notes.} Rows correspond to different accuracy measures (forecast and forecaster level, respectively). \emph{EQM}: correlation between the EQM composite score and the relevant accuracy measure. \emph{Pre-LLM}: correlation between the pre-LLM composite score and the relevant accuracy measure. \emph{Between}: correlation between the EQM and pre-LLM composite scores. \emph{EQM Better}: $p$-value from a test of dependent correlations for whether EQMs outperformed pre-LLM methods.
\end{minipage}
\end{table}

At the forecast level ($N = 55{,}463$), EQM composite scores were positively correlated with forecast-level accuracy, \ensuremath{BS_{\text{diff}}}, with a correlation of $r = .19$ ($p < .001$). The EQM correlation was more than three times larger than the pre-LLM correlation of $r = .06$ ($p < .001$). A test of dependent correlations rejected equality at $p < .001$. At the forecaster level ($N = 1{,}770$), EQM composite scores correlated with forecaster-level accuracy, \ensuremath{ABS_{\text{norm}}}, at $r = .51$ ($p < .001$) versus $r = .39$ ($p < .001$) for the pre-LLM methods, with the two correlations also statistically significantly different ($p < .001$). The two composite scores were themselves correlated ($r=.16$ at the forecast level, $r=.69$ at the forecaster level), confirming partial overlap but a distinct signal. 

\textbf{EQM composite score drivers.} At the forecast level, 19 of the 50 retained EQM patterns received nonzero coefficients in the LASSO regression.\footnote{The coefficients from the LASSO models for both EQM and pre-LLM composite scores are provided in Appendix~\ref{app:lasso_coefficients}. For comparability, we present OLS regression coefficients in Appendix~\ref{app:ols_coefficients}.} The largest positive coefficients for the EQM composite score were for \emph{Forecast and Rationale Align} ($\beta = 0.023$) and \emph{Inevitable or Impossible} ($\beta = 0.011$), while the largest negative coefficients were for \emph{Extreme Confidence} ($\beta = -0.019$) and \emph{Confirmation Bias} ($\beta = -0.010$). At the forecaster level, 20 of the 50 patterns entered the model. The largest positive coefficients were for \emph{Forecast and Rationale Align} ($\beta = 0.046$), \emph{Timing} ($\beta = 0.032$), and \emph{Teaming Agree} ($\beta = 0.025$), while the largest negative coefficients were for \emph{Simplification Bias} ($\beta = -0.036$), \emph{Forecast and Rationale Misalign} ($\beta = -0.031$), and \emph{Inside View} ($\beta = -0.027$). In both models, the signs of the non-zero coefficients were broadly consistent with the directional hypotheses in Appendix~\ref{appendix:NSPatterns}.

{\bf Interpreting effect sizes.} Interpreting effect sizes requires some care in our empirical setting. Forecaster-level correlations are substantially larger than forecast-level correlations because averaging over many forecasts reduces the irreducible outcome noise \citep{morrison1972upper,murphy1973new,srinivasan2025womac}. For example, even a perfectly calibrated forecast of 0.1 will still sometimes resolve positively (outcome = 1), attenuating forecast-level correlations towards zero. For this reason, Study 2 benchmarks the EQM-accuracy correlations against traditional indicators of forecasting skill to put their absolute magnitude in context.

\begin{figure}[t!]
  \centering
  \includegraphics[width=0.95\textwidth]{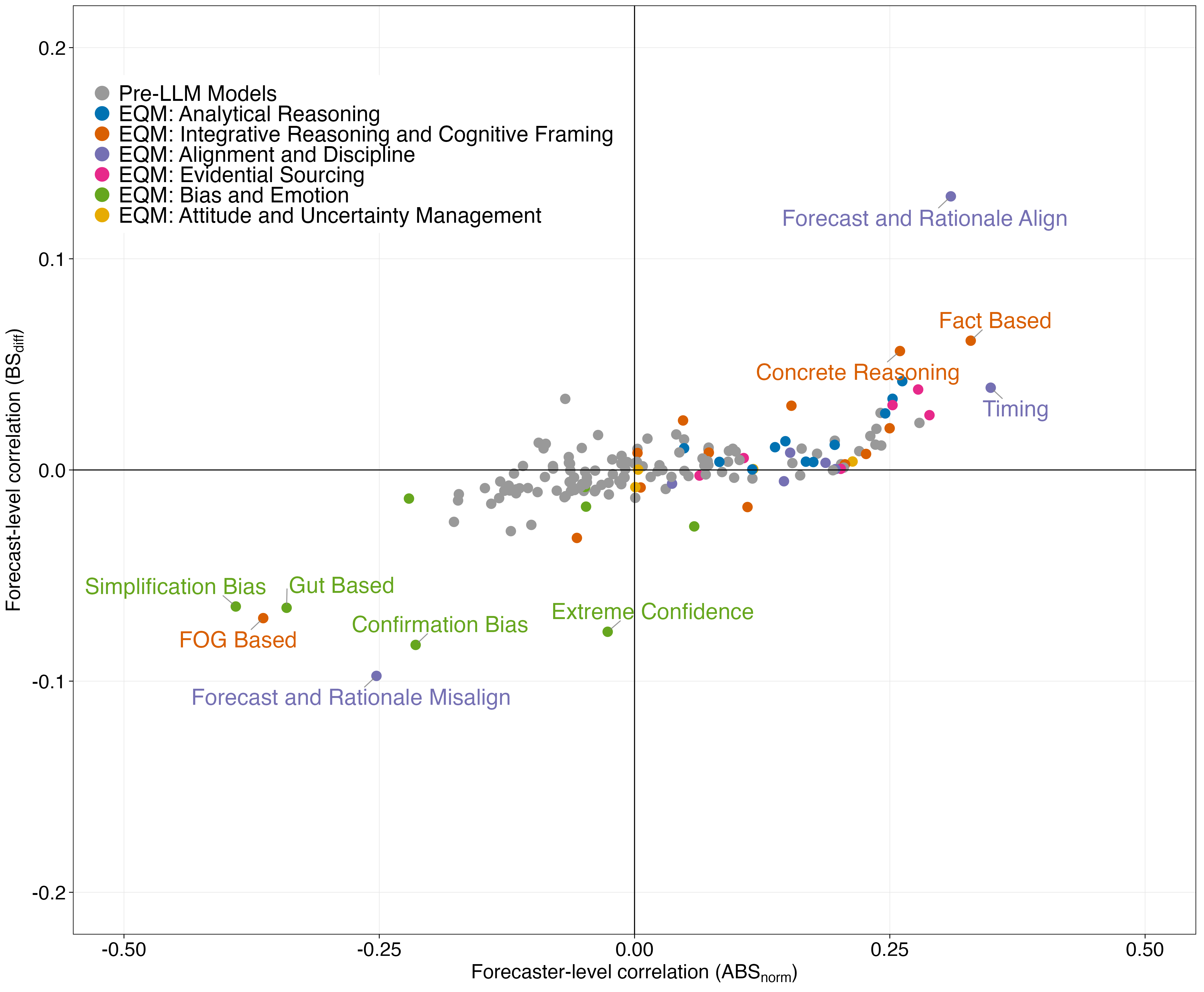}
  \caption{
    \emph{Pattern-level correlations with forecasting accuracy}.
    Each point represents one EQM pattern or one pre-LLM feature.
    \emph{$x$-axis}: correlation with forecaster-level accuracy (\ensuremath{ABS_{\text{norm}}}, $N = 1{,}770$). \emph{$y$-axis}: correlation with forecast-level accuracy (\ensuremath{BS_{\text{diff}}}, $N = 55{,}463$).
    EQM patterns are colored by family; pre-LLM features are shown in gray.
    Labels are provided for selected EQM patterns with large absolute correlations.
  }
  \label{fig:individual_cors}
\end{figure}

{\bf Pattern-level results.} Figure~\ref{fig:individual_cors} plots each pattern's correlation with forecaster-level accuracy (\ensuremath{ABS_{\text{norm}}}, $x$-axis) against its correlation with forecast-level accuracy (\ensuremath{BS_{\text{diff}}}, $y$-axis). EQM patterns are colored by family; pre-LLM features are shown in gray. Patterns in the upper-right quadrant were positively associated with forecast- and forecaster-level accuracy, whereas those in the lower-left quadrant were negatively associated with both. The off-diagonal quadrants contain patterns that were positively associated with one accuracy measure and negatively associated with the other.

Several EQM patterns appeared in the upper-right quadrant, including \textit{Forecast and Rationale Align}, \textit{Fact Based}, \textit{Concrete Reasoning}, \textit{Timing}, and other analytical reasoning and evidential sourcing patterns. The lower-left quadrant was dominated by the bias family---\textit{Gut Based}, \textit{Simplification Bias}, \textit{Confirmation Bias}, and \textit{Extreme Confidence}---as well as \textit{FOG Based} and \textit{Forecast and Rationale Misalign}. Three broad regularities emerged: biases tended to be associated with lower accuracy, structured analytical reasoning with greater accuracy (no analytical pattern was negatively correlated with either accuracy measure), and framing- and alignment-related patterns were mixed, as hypothesized.

In contrast, no pre-LLM feature reached the predictive strength of the top EQM patterns. Thirteen EQM patterns exceeded the strongest pre-LLM feature (LIWC's \textit{negate} with $r = .03$) in absolute correlation with forecast-level accuracy (\ensuremath{BS_{\text{diff}}}), and seven exceeded the strongest pre-LLM feature (\textit{CC\_RF} with $r = .28$) in absolute correlation with forecaster-level accuracy (\ensuremath{ABS_{\text{norm}}}). The EQM pattern set, therefore, provided more diagnostic information than the pre-LLM benchmark not only at the composite level but also at the level of individual patterns. The square root of word count (as suggested by \cite{schwartz2017assessing}), which we use as our main rationale length measure, had a minimal correlation ($|r| < .01$) at the forecast level. At the forecaster level, this metric had a more meaningful but still small correlation with accuracy ($r = .24$), and was surpassed in an absolute sense by three pre-LLM features (including $CC\_RF$, as noted above) and 15 EQM patterns.\footnote{Raw word count behaved nearly identically, with $|r| < .01$ at the forecast level and $r = .19$ at the forecaster level.}

\begin{figure}[t!]
  \centering
  \includegraphics[width=\textwidth]{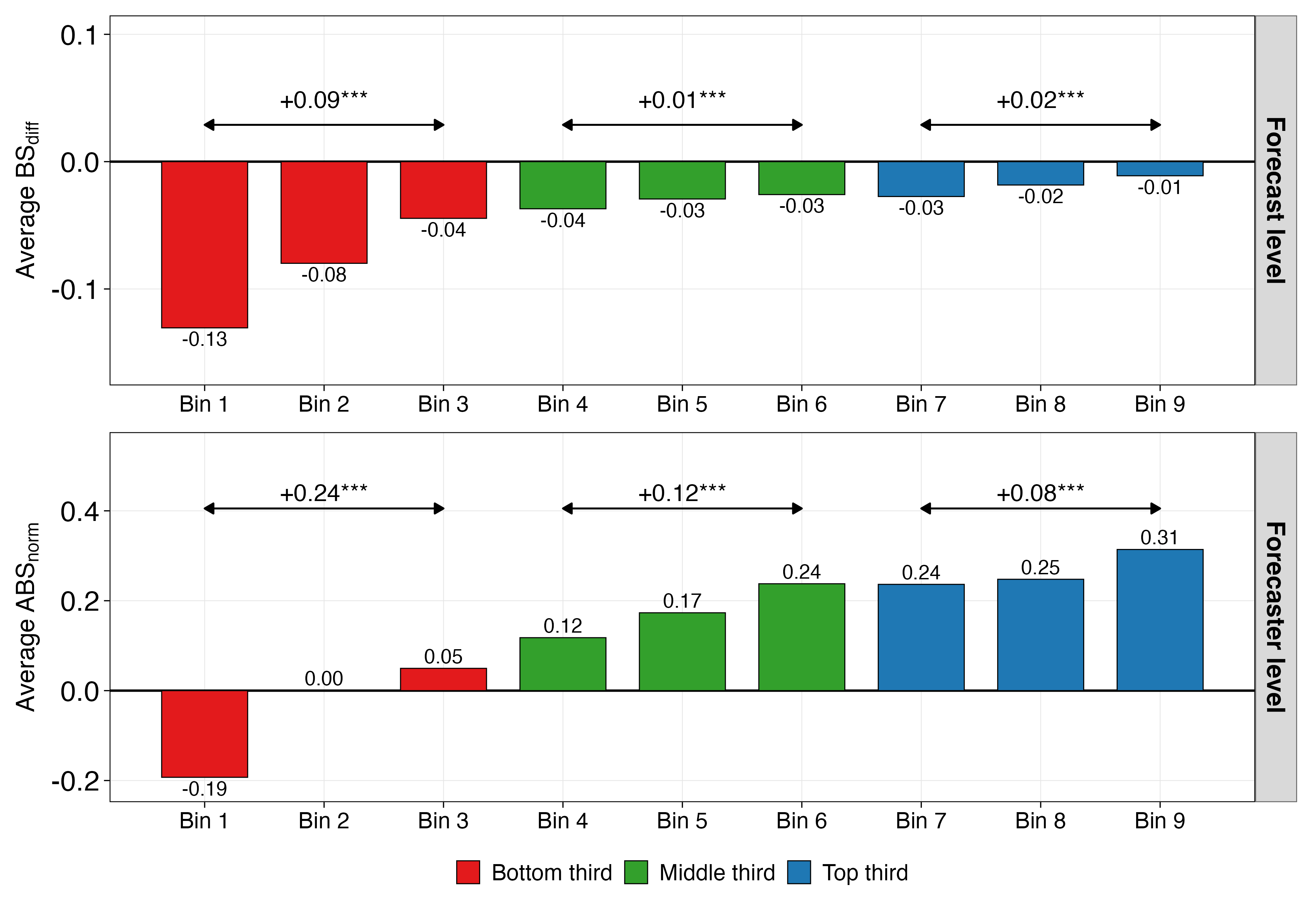}
  \caption{
    \emph{Mean accuracy across the EQM composite score distribution}.
    Observations are sorted into nine equal-sized bins by the EQM composite score. Bars show mean accuracy within each bin, with bins grouped visually into bottom, middle, and top thirds of the EQM distribution. \emph{Top panel}: forecast-level accuracy, measured by \ensuremath{BS_{\text{diff}}} ($N = 55{,}463$). \emph{Bottom panel}: forecaster-level accuracy, measured by \ensuremath{ABS_{\text{norm}}} ($N = 1{,}770$). Positive values indicate better accuracy. Asterisks indicate $t$-tests comparing the outer bins within each third: * $p < .05$, ** $p < .01$, *** $p < .001$.
  }
  \label{fig:bin9_accuracy}
\end{figure}

{\bf Directional consistency.} Pattern-level results were also broadly consistent with our directional hypotheses (see Appendix~\ref{appendix:NSPatterns} for the full list of hypotheses). For the EQM patterns, we considered only those correlations that were statistically significant at $p < .001$. At the forecast level, 23 patterns met this criterion, of which 21 (91\%) matched the hypothesized direction. The only mismatches were for \textit{State Verbs} ($r = .02$) and \textit{Descriptive Action Verbs} ($r = -.02$) that both had negligible effect sizes.
At the forecaster level, 36 EQM patterns were significant at $p < .001$, of which 34 (94\%) aligned with the hypothesized direction. The two exceptions were \textit{Adjusting Up} ($r = .15$) and \textit{Wallowing in Uncertainty} ($r = .12$), both of which had been hypothesized to correlate negatively with forecaster-level accuracy but instead exhibited modest positive associations. Overall, the EQM patterns were associated with both forecast- and forecaster-level accuracy in ways that largely align with our hypotheses,\footnote{This pattern was not driven by the $p<.001$ threshold. At $p < .01$, 38 forecaster-level patterns and 26 forecast-level patterns were statistically significant, with 92\% of each set matching the hypothesized direction. At $p < .05$, 44 forecaster-level patterns and 28 forecast-level patterns were statistically significant, with 86\% and 89\% matching the hypothesized direction, respectively. Even when all 50 EQM patterns were included regardless of statistical significance, 80\% matched the hypothesized direction at both levels of analysis.} and Appendix~\ref{appendix:pattern_correlation_plots} shows that these accuracy-relevant patterns also correlate with one another in theoretically expected ways, further supporting the coherence of the EQM pattern set.

\subsubsection{Predictive Signal Across the EQM Score Distribution}\label{subsec:signal_distribution}

A natural follow-up question is whether EQMs identify forecasts or forecasters associated with high accuracy, screen out those associated with low accuracy, or both. To answer this question, we sorted forecasts and forecasters into nine equal-sized bins by their EQM composite score and examined average accuracy within each bin (Figure~\ref{fig:bin9_accuracy}).

Accuracy rose monotonically with EQM composite scores, but the gains were largest at the bottom third of the distribution. At the forecast level, average \(BS_{\text{diff}}\) rose from \(-0.13\) in Bin 1 to \(-0.04\) in Bin 3, a within-tercile gain of \(0.09\) (\(p < .001\)). Outside the bottom third, the relationship was largely flat, with the middle and top terciles yielding much lower gains of 0.01 and 0.02, respectively.  EQMs thus sharply distinguished the weakest rationales from the rest, while providing only modest additional differentiation among higher-scoring ones.

The forecaster-level results showed a similar asymmetry but with the signal persisting further up the distribution. The bottom tercile again drove the largest gain: average \(ABS_{\text{norm}}\) rose from \(-0.19\) in Bin 1 to \(0.05\) in Bin 3, a within-tercile gain of \(0.24\) (\(p < .001\)). Unlike at the forecast level, however, accuracy continued to rise in the middle and top terciles, with gains of 0.12 and 0.08, respectively. EQMs therefore carried information across the full forecaster-level distribution, although the marginal impact was strongest at the bottom.

\FloatBarrier

\subsubsection{Sensitivity and Reliability Analyses}\label{subsec:EQM_sensitivity}

To assess whether the Study 1 results depend on specific implementation choices, we next varied the LLM, the prompt, the accuracy benchmark, and the modeling approach in turn. The main findings were broadly stable across all four dimensions.

{\bf LLM choice and ensembling.} Re-scoring all rationales with Gemini 2.5 Pro instead of GPT-4o produced a slightly weaker forecast-level correlation ($r = .15$ vs. $r = .19$ for GPT-4o) but a somewhat stronger forecaster-level correlation ($r = .54$ vs. $r = .51$ for GPT-4o). Averaging the two models matched GPT-4o at the forecast level (\( r = .19 \)) and modestly improved performance at the forecaster level (\( r = .55 \)). Thus, ensembling appeared to offer at most a limited improvement.

{\bf Cross-model agreement.} Across all 60 patterns, GPT-4o and Gemini 2.5 Pro agreed at Cohen’s $\kappa = .59$ and $r = .65$, suggesting moderate-to-substantial agreement \citep{landis1977measurement, mchugh2012kappa}. When collapsing scores of 1 and 2 into a single bin, we obtained a Cohen's $\kappa = .56$, $ r = .59$, and an overall agreement percentage of 82\%. Pattern-specific reliabilities are provided in Appendix~\ref{appendix:reliability}. 

{\bf Prompt sensitivity.} We examined prompt-level sensitivity across six prompt variations and two LLMs (GPT-4o and Gemini 2.5 Flash). Identical agreement with the baseline ranged from 82.0\% to 95.3\% for GPT-4o and from 86.0\% to 92.8\% for Gemini 2.5 Flash, with GPT-4o more variable across structural changes such as removing the pattern name from the output. A separate ablation that scored each pattern individually rather than jointly diverged far more sharply---59.6\% identical scores for GPT-4o and 75.0\% for Gemini 2.5 Flash, the largest divergence among configurations tested. One plausible explanation is that seeing related EQMs together helps the model distinguish them; isolating each pattern removes that signal. Detailed prompt-variation tables and the single-pattern analysis are reported in Appendices~\ref{appendix:sensitivity} and \ref{subsec:single_pattern_scoring}.

{\bf Alternative modeling choices.}
Expanding the local crowd benchmark from the median of 11 to median of 21 prior forecasts left the forecast-level correlation unchanged (\( r = .19 \)). Collapsing the three-point scale (0/1/2) into a binary one (0/1) reduced correlations modestly (\( r = .16\text{--}.17 \)), suggesting that the three-level scale contains some but very modest additional information. Allowing nonlinear structure via XGBoost increased the forecast-level correlation slightly (\( r = .22 \)), a small gain that suggests the LASSO benchmark used throughout the paper is, if anything, a conservative estimate of the signal extractable from EQMs.\footnote{The XGBoost implementation followed the same partitioning and held-out prediction logic described in Appendix~\ref{sec:composite_scoring}, but replaced the LASSO tuning parameter with a small boosted-tree hyperparameter grid over tree depth and minimum child weight, using a fixed learning rate of 0.10, 80\% row and column subsampling, and early stopping to select the number of boosting rounds.}

\subsubsection{Conclusion}\label{subsec:study1conclusion}

Study 1 supports the central claim of the paper: LLM-scored EQMs extracted more accuracy-relevant signal from written rationales than pre-LLM text-analysis methods, and the pattern-level correlations were directionally consistent with our theoretical hypotheses. The predictive signal was asymmetric, with EQMs more reliably flagging low-accuracy forecasts and forecasters than identifying the highest-accuracy ones. For decision-makers, EQMs appear more useful for screening out weak judgments than for identifying the very best forecasts.

Three questions follow. First, how large is the EQM signal relative to other indicators a decision-maker might already have? Second, how do EQM composite scores compare to human ratings? Finally, does the framework transfer beyond the specific ACE forecasting tournament? Studies 2, 3, and 4 take up these questions in turn.

\subsection{Study 2: How Useful Are EQMs Relative to Other Indicators of Judgment Quality?}\label{subsec:CaseStudy}

\subsubsection{Introduction}\label{subsec:study2intro}

Study 1 demonstrated that EQMs extract more accuracy-relevant signal from rationales than 
pre-LLM text-analysis methods. The next question is practical: how strong is that signal relative to other indicators a decision-maker might use to evaluate judgments?

Three main families of such indicators have been used in the forecasting literature. The first is \emph{past performance}. Prior accuracy has been found to be the single strongest predictor of future accuracy \citep{mellers2015drivers,mellers2015superforecasters,himmelstein2020forecasting}. However, prior accuracy is only available after a track record has accumulated, a process that can take weeks or months. The second is \emph{individual differences}. For example, cognitive tests and personality measures predict accuracy and can be administered in single-shot elicitation settings, but they require separate elicitation infrastructure and assume that test performance transfers to the judgment task at hand \citep{mellers2017generalizable,himmelstein2024fpt}. The third is \emph{behavioral indicators} such as updating frequency and update size. However, these indicators require judgments over time and hence   do not apply in single-shot elicitation settings \citep{atanasov2020small}. EQMs occupy a different niche: they require only a single written rationale (which is typically elicited anyway), with no track record, no separate testing, and no repeated elicitation. Table~\ref{tab:indicators} summarizes these tradeoffs.

\begin{table}[t!]
\centering
\caption{Practical requirements of common indicators of judgment quality}
\label{tab:indicators}
\small
\begin{tabular}{lccc}
\toprule
 & \multicolumn{1}{c}{No track} & \multicolumn{1}{c}{No separate} & \multicolumn{1}{c}{Single-shot} \\
Indicator & \multicolumn{1}{c}{record needed} & \multicolumn{1}{c}{instrument needed} & \multicolumn{1}{c}{elicitation OK} \\
\midrule
Past performance (e.g., prior accuracy)             & \xmark & \cmark & \cmark \\
Individual differences (e.g., cognitive batteries)  & \cmark & \xmark & \cmark \\
Behavioral (e.g., update size, update frequency)    & \cmark & \cmark & \xmark \\
\addlinespace
{\bf Explanation Quality Markers (EQMs)}                   & \cmark & \cmark & \cmark \\
\bottomrule
\end{tabular}
\end{table}

In Study 2, we benchmark EQMs against indicators from each of these three families. We do so in two ways. First, we compare how strongly each indicator correlates with accuracy at the forecast and forecaster levels, providing a direct read on relative effect sizes. Second, in a more stringent test, we examine whether EQMs can improve crowd aggregation by ranking and selecting forecasts within the pool. Taken together, these exercises help locate EQMs in the broader landscape of judgment-quality indicators.

\subsubsection{Design}\label{subsec:CaseStudyData}

We used a season-over-season design linking adjacent ACE seasons (Seasons 1$\rightarrow$2, 2$\rightarrow$3, and 3$\rightarrow$4). For each pair, the earlier season served as the training season (e.g., for computing prior accuracy and fitting LASSO models), and the latter served as the test season in which accuracy was evaluated. 

We benchmarked EQMs against four alternatives: prior-season accuracy, current-season behavioral indicators capturing engagement and responsiveness (average forecast count per question and average update size), the pre-LLM composite score, and current-season word count.\footnote{All measures were hypothesized to be positively correlated with our dependent accuracy measures, except for update size. \cite{atanasov2020small} found that more accurate forecasters tend to make more frequent but smaller updates. To orient every correlation in the same direction, we used the absolute value of the correlation between update size and accuracy.} Both EQM and pre-LLM composite scoring models were trained on the prior season following the procedure described in Section~\ref{subsec:exploratory_training} and applied out-of-sample to the current (test) season.\footnote{For simplicity, we use word count as opposed to the square-root of word count given the comparable correlations in Study 1.}

The sample was restricted to forecasters with at least five scored questions in the prior season and at least five test-season questions with rationales of ten or more words. These joint restrictions resulted in $N = 990$ forecasters across seasons 2 through 4, slightly more restrictive than the single-season forecaster-level requirement in Study 1.\footnote{Study 1 required ten forecasts in a season with rationales of ten words or more, yielding $N = 1{,}280$ forecasters across seasons 2 through 4.} To account for differing levels of participation, we used \emph{participation-adjusted} $ABS_{\text{norm}}$, defined as $ABS_{\text{norm}} \times (n/N)$, where $n$ is the number of binary questions forecast by a given forecaster in a season, and $N$ is the total number of binary questions in that season.\footnote{When a forecaster submits a single forecast on each question within a season without updating, update size is undefined. Rather than setting update size to zero for these cases, we imputed it using the intercept (approximately 0.203) from a linear regression of average update size on forecast frequency.} Our results were robust to both relaxing the participation thresholds to zero and tightening them to ten questions per season; the former required additional imputation of missing prior-season statistics. Details for these robustness checks are provided in Appendix~\ref{appendix:Study2AccuracyImpute}.

\subsubsection{Benchmarking Results}\label{subsec:CaseStudyCors}

We report correlations at two levels of analysis. Table~\ref{tab:sos_summary} evaluates indicators against forecast-level accuracy ($BS_{\text{diff}}$), while
Table~\ref{tab:forecaster_sos_5_5} evaluates them against participation-adjusted forecaster-level accuracy ($ABS_{\text{norm}} \times (n/N)$). Findings differ across the two levels.

{\bf Forecast level.} Predictability was modest at the forecast level (Table~\ref{tab:sos_summary}) for every indicator we considered, consistent with the irreducible outcome noise attenuating forecast-level correlations \citep{morrison1972upper,murphy1973new}; see Section~\ref{subsec:hypothesis_testing} for more discussion. Within that low ceiling, the forecast-level EQM composite score was the strongest predictor in all three season pairs (mean $r = .16$), narrowly exceeding prior accuracy and update size (both mean $r = .13$). The pre-LLM composite score (mean $r = .05$) and word count (mean $r = .03$) are largely uninformative. Thus, at the forecast level, the EQM composite score appeared to be the most useful indicator of forecast accuracy, although the absolute level of predictability was modest.

{\bf Forecaster level.} At the forecaster level (Table~\ref{tab:forecaster_sos_5_5}), prior-season accuracy was the strongest predictor of current-season accuracy (mean $r = .70$).
This reproduced a well-established result that forecasting skill is a partly stable individual trait \citep{mellers2015drivers,mellers2015superforecasters,himmelstein2020forecasting}. Of the remaining indicators, average update size (mean $r = .57$) and the EQM composite score (mean $r = .55$) were essentially tied, followed by forecast count (mean $r = .52$), the pre-LLM composite score (mean $r = .43$), and word count (mean $r = .30$). Thus, EQMs did not outperform prior accuracy, but they were competitive with behavioral indicators and clearly improved on the pre-LLM methods.

\begin{table}[t!]
\centering
\footnotesize
\caption{Season-over-season correlations of indicators with forecast-level accuracy}
\label{tab:sos_summary}
\renewcommand{\arraystretch}{1.1}
\setlength{\tabcolsep}{7pt}
\begin{tabular}{l l c c c c}
\toprule
\textbf{Indicator} & \textbf{Calculation Notes} & \textbf{S1$\rightarrow$S2} & \textbf{S2$\rightarrow$S3} & \textbf{S3$\rightarrow$S4} & \textbf{Mean} \\
\midrule
EQM forecast composite
& Forecast-level
& 0.17 & 0.17 & 0.15 & 0.16 \\

Pre-LLM forecast composite
& Forecast-level
& 0.06 & 0.03 & 0.05 & 0.05 \\

Word count
& Forecast-level
& 0.04 & 0.03 & 0.01 & 0.03 \\

\midrule
Past accuracy
& Forecaster-level from previous season data
& 0.14 & 0.13 & 0.12 & 0.13 \\

Forecast count
& Forecaster-level from current season data
& 0.14 & 0.09 & 0.05 & 0.09 \\

Update size
& Forecaster-level from current season data
& 0.15 & 0.12 & 0.11 & 0.13 \\

EQM forecaster composite
& Forecaster-level from current season data
& 0.12 & 0.11 & 0.09 & 0.11 \\

Pre-LLM forecaster composite
& Forecaster-level from current season data
& 0.08 & 0.09 & 0.07 & 0.08 \\
\bottomrule
\end{tabular}

\vspace{0.35em}
\begin{center}
\begin{minipage}{0.92\textwidth}
\scriptsize
\textit{Notes.} Rows correspond to different indicators correlated with \ensuremath{BS_{\text{diff}}}, forecast-level accuracy. Sample is restricted to forecasters with at least five scored questions in the prior season and at least five current-season forecasts with rationales of ten or more words. \emph{Indicator}: the metric correlated with current-season forecast-level accuracy. \emph{Calculation Notes}: whether the indicator is forecast-level or forecaster-level, and the source of the data for forecaster-level indicators. \emph{Si$\rightarrow$Sj}: Pearson correlation of the indicator with forecast-level accuracy during the $j^{th}$ season, using the $i^{th}$ season's data to calculate prior accuracy and train LASSO composite models. \emph{Mean}: average of the three \emph{Si$\rightarrow$Sj} correlations. \emph{EQM forecast composite}: forecast-level EQM composite score of the individual forecast. \emph{Pre-LLM forecast composite}: forecast-level pre-LLM composite score of the individual forecast. \emph{Word count}: number of words in the individual forecast's rationale. \emph{Past accuracy}: forecaster's prior-season participation-adjusted normalized accuracy. \emph{Forecast count}: average number of forecasts per question in the current season. \emph{Update size}: average absolute size of forecast updates in the current season. \emph{EQM forecaster composite}: composite score derived from current-season EQM patterns averaged at the forecaster level. \emph{Pre-LLM forecaster composite}: composite score derived from current-season pre-LLM features averaged at the forecaster level. For the forecast-level composite scores, the LASSO model was fit on previous-season forecast-level data and then applied to the current forecast's EQM pattern or pre-LLM feature scores. For the forecaster-level composite scores, the LASSO model was fit on previous-season averaged forecaster-level data and then applied to current-season averaged EQM pattern or pre-LLM feature scores. Sample sizes are: Season 1$\rightarrow$2, $N = 5{,}851$ forecast-level observations; Season 2$\rightarrow$3, $N = 5{,}188$ forecast-level observations; Season 3$\rightarrow$4, $N = 12{,}219$ forecast-level observations.
\end{minipage}
\end{center}
\end{table}

\begin{table}[t!]
\centering
\footnotesize
\caption{Season-over-season correlations of indicators with forecaster-level accuracy}
\label{tab:forecaster_sos_5_5}
\renewcommand{\arraystretch}{1.1}
\setlength{\tabcolsep}{7pt}
\begin{tabular}{l l c c c c}
\toprule
\textbf{Indicator} & \textbf{Calculation Notes} & \textbf{S1$\rightarrow$S2} & \textbf{S2$\rightarrow$S3} & \textbf{S3$\rightarrow$S4} & \textbf{Mean} \\
\midrule
Past accuracy
& Forecaster-level from previous season data
& 0.72 & 0.68 & 0.71 & 0.70 \\

Forecast count
& Forecaster-level from current season data
& 0.54 & 0.58 & 0.45 & 0.52 \\

Update size
& Forecaster-level from current season data
& 0.59 & 0.62 & 0.50 & 0.57 \\

EQM forecaster composite
& Forecaster-level from current season data
& 0.55 & 0.53 & 0.57 & 0.55 \\

Pre-LLM forecaster composite
& Forecaster-level from current season data
& 0.35 & 0.41 & 0.52 & 0.43 \\

Word count
& Forecaster-level from current season data
& 0.35 & 0.32 & 0.23 & 0.30 \\
\bottomrule
\end{tabular}

\vspace{0.35em}
\begin{center}
\begin{minipage}{0.92\textwidth}
\scriptsize
\textit{Notes.} Rows correspond to different indicators correlated with current-season participation-adjusted forecaster-level accuracy ($ABS_{\text{norm}} \times (n/N)$). Sample is restricted to forecasters with at least five scored questions in the prior season and at least five current-season forecasts with rationales of ten or more words. \emph{Indicator}: the metric correlated with current-season forecaster-level accuracy. \emph{Calculation Notes}: source of the data for forecaster-level indicators. \emph{Si$\rightarrow$Sj}: Pearson correlation of the indicator with forecaster-level accuracy during the $j^{th}$ season, using the $i^{th}$ season's data to calculate prior accuracy and train LASSO composite models. \emph{Mean}: average of the three \emph{Si$\rightarrow$Sj} correlations. \emph{Past accuracy}: forecaster's prior-season participation-adjusted normalized accuracy. \emph{Forecast count}: average number of forecasts per question in the current season. \emph{Update size}: average absolute size of forecast updates in the current season. \emph{EQM forecaster composite}: composite score derived from current-season EQM patterns averaged at the forecaster level. \emph{Pre-LLM forecaster composite}: composite score derived from current-season pre-LLM features averaged at the forecaster level. \emph{Word count}: average number of words in the forecaster's rationales. For the forecaster-level composite scores, the LASSO model was fit on previous-season averaged forecaster-level data and then applied to current-season averaged EQM pattern or pre-LLM feature scores. Sample sizes are: Season 1$\rightarrow$2, $N = 270$ forecasters; Season 2$\rightarrow$3, $N = 209$ forecasters; Season 3$\rightarrow$4, $N = 511$ forecasters.
\end{minipage}
\end{center}
\end{table}

\subsubsection{Crowd Aggregation}\label{subsec:CaseStudyAgg}

We next asked whether EQMs could improve accuracy when used for forecast aggregation. We used each indicator from Section~\ref{subsec:CaseStudyCors} to rank forecasts within a question, select a subset of those forecasts, and compare the resulting aggregate to a baseline that used all forecasts.

{\bf Design.} We first formed the pool of eligible forecasters using the same sample restrictions of at least five scored questions in the prior season and at least five test-season questions with rationales of ten or more words, focusing on questions from seasons 2 through 4. To reduce the impact of variation in pool size and posting timing, we restricted attention to the first thirty forecasts on each question and kept only questions for which those thirty forecasts arrived within two weeks of the question opening. We compared two aggregators (mean and median) crossed with three selection rules: (1) all thirty forecasts (baseline), (2) the top two-thirds (twenty forecasts), and (3) the top third (ten forecasts), with all ties broken randomly. Aggregates were scored using Brier scores and then averaged across questions. Robustness checks under more and less restrictive samples are reported in Appendix~\ref{appendix:ThresholdingAggregations}.

\begin{figure}[t!]
\centering
\includegraphics[width=\linewidth]{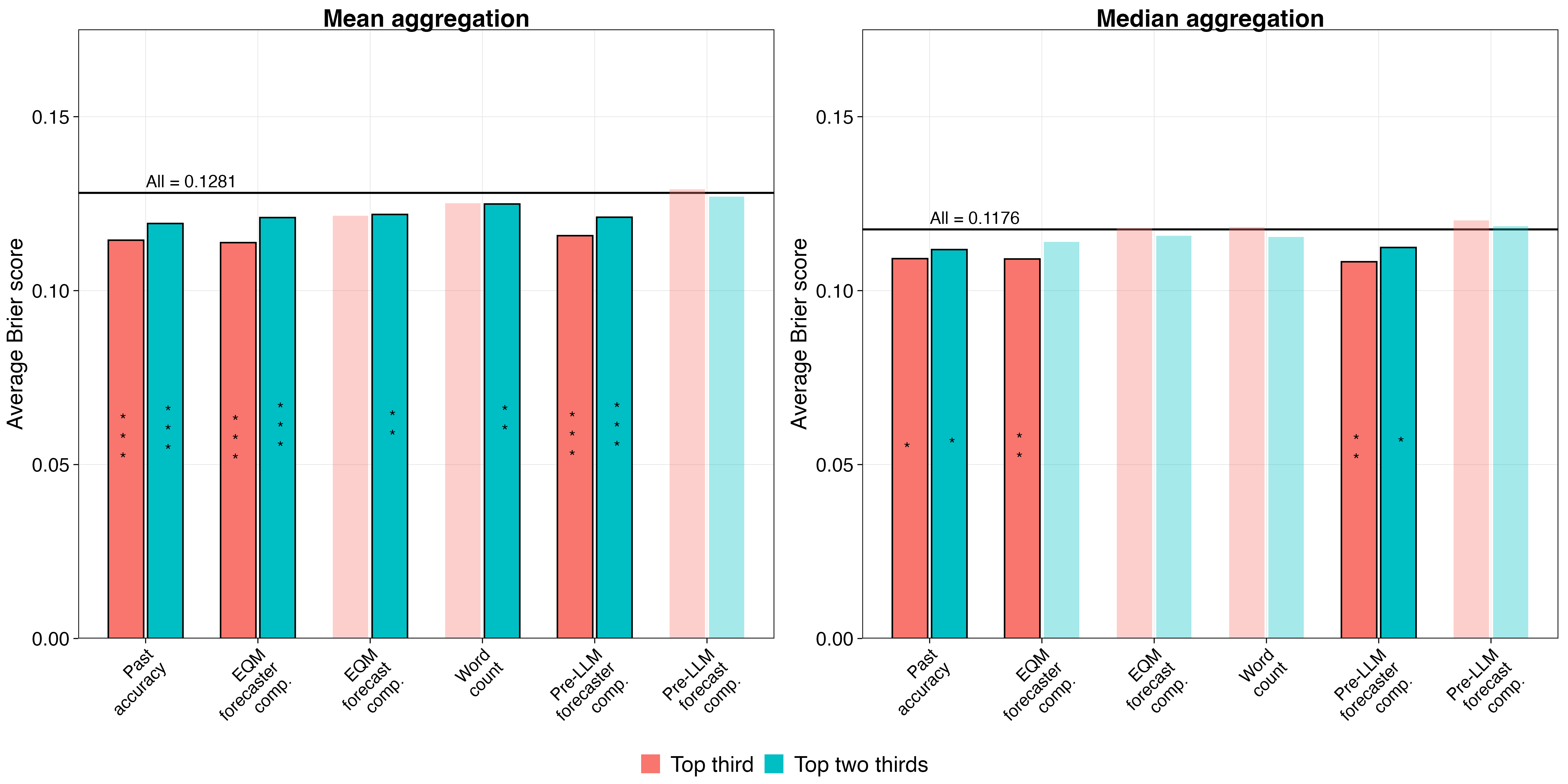}
\caption{\emph{Aggregate Brier scores by indicator and selection rule}. \emph{Left panel}: mean aggregation. \emph{Right panel}: median aggregation. The horizontal line marks the baseline Brier score of the aggregate computed from all thirty forecasts on each question. Bars show the average Brier score after retaining either the top third (ten forecasts) or top two-thirds (twenty forecasts) within each question, ranked by the indicator on the $x$-axis. Lower score is better. Asterisks indicate paired $t$-tests against the baseline: * $p < .05$, ** $p < .01$, *** $p < .001$. \emph{Past accuracy} is the forecaster's prior-season participation-adjusted normalized accuracy ($ABS_{\text{norm}} \times (n/N)$); \emph{EQM forecaster comp.} is the composite score derived from current-season EQM patterns averaged at the forecaster level; \emph{EQM forecast comp.} is the forecast-level EQM composite score of the individual forecast; \emph{Word count} is the number of words of the individual forecast's rationale; \emph{pre-LLM forecaster comp.} is the composite score derived from current-season pre-LLM features averaged at the forecaster level; \emph{pre-LLM forecast comp.} is the forecast-level pre-LLM composite score of the individual forecast. For the forecast-level composite scores, the LASSO model was fit on previous-season forecast-level data and then applied to the current forecast's EQM pattern or pre-LLM feature scores. For the forecaster-level composite scores, the LASSO model was fit on previous-season averaged forecaster-level data and then applied to current-season averaged EQM pattern or pre-LLM feature scores.}
\label{fig:aggregation_selection_combined}
\end{figure}

{\bf Results.} Figure~\ref{fig:aggregation_selection_combined} summarizes the aggregation results. Several key findings emerged. First, median aggregation generally performed better than mean aggregation in this sample, consistent with the fact that the median is less sensitive to outlier forecasts. Accordingly, the mean baseline was easier to improve upon, and several selection rules did so in a statistically significant manner. The median baseline set a higher bar, and fewer selection rules cleared it. Second, the clearest gains resulted from ranking on \emph{forecaster}-level indicators, prior accuracy and EQM forecaster-level composite score in particular. The pre-LLM forecaster-level composite score also produced improvements in several conditions. By contrast, forecast-level indicators did little to improve aggregate accuracy. This is especially striking for the forecast-level EQM composite score, which 
was the strongest predictor of forecast-level accuracy in Section~\ref{subsec:CaseStudyCors} but produced little aggregation benefit, especially when using median aggregation. The pattern is consistent with the asymmetry documented in Section~\ref{subsec:signal_distribution}: forecast-level EQMs identify poor forecasts more reliably than excellent ones. Therefore, selecting forecasts on the basis of forecast-level EQM scores removes weak forecasts but adds little information among the rest. Under median aggregation, some of these weak forecasts already have little impact due to the robustness properties of the median.

\FloatBarrier

\subsubsection{Conclusion}\label{subsec:study2conclusion}

Study 2 clarifies where EQMs fit among practical indicators of judgment quality. At the forecast level, EQMs were the strongest available indicator, outperforming both prior accuracy and the pre-LLM methods, although the absolute level of predictability was low. At the forecaster level, prior accuracy was the strongest predictor of future accuracy. However, the EQM composite score compared favorably with behavioral indicators  and consistently improved on the pre-LLM methods. In the aggregation analysis, the strongest gains resulted from ranking forecasters by forecaster-level indicators, especially prior accuracy and the EQM composite score, whereas forecast-level EQM scores appeared more useful for screening weak individual judgments than for improving aggregates.

\subsection{Study 3: EQMs Versus Human Ratings}\label{subsec:HumansvsEQM}

\subsubsection{Introduction}\label{subsec:HumansIntro}

Studies 1 and 2 established that EQMs extract more accuracy-relevant signal from written rationales than pre-LLM methods, and that they are competitive with several traditional indicators of forecasting skill. A natural next benchmark is \emph{human ratings}. Decision-makers who receive judgments accompanied by written explanations typically rely on their own reading of those explanations to decide which judgments to trust. Are such impressions likely to track judgment accuracy?  

Study 3 addresses this question using a subset of ACE rationales for which contemporaneous human ratings of rationale quality were available. We compare EQM composite scores and average human ratings on three dimensions: (i) how strongly each correlates with forecast- and forecaster-level accuracy; (ii) where in the score distribution the predictive signal is concentrated; and (iii) how reliably each is measured, as captured by inter-rater agreement. Beyond comparing correlations with accuracy, this design allows us to examine whether human raters and EQMs respond to the same features of written explanations, or to different ones.

\subsubsection{Data and Descriptive Statistics}\label{subsec:HumansData}

During the fourth and final ACE tournament season, selected rationales from the main forecasting platform were displayed as ``tips'' to forecasters on a parallel platform, who then read and rated each rationale's usefulness on a five-point scale. In this section, we refer to the rationale writers as \emph{forecasters} and to the readers as \emph{raters}. Raters only saw the rationale but not the associated probability forecast \citep{chang2017accountability,schwartz2017assessing}. Because raters had already forecast on the same questions themselves, they were familiar with the forecasting problems and their context. After submitting a rating, raters could optionally revise their own forecast. Our analysis focuses on the ratings themselves rather than on any subsequent forecast revisions. 

The rating rubric was based on the \textit{CHAMPS KNOW}\footnote{\textit{CHAMPS} stands for \textbf{C}omparison classes, \textbf{H}unt for the right information, \textbf{A}djusting forecasts as evidence changed, the use of \textbf{M}athematical and statistical models, \textbf{P}ost-mortems on forecasting errors, and \textbf{S}elective attention to where one invested time and effort; \textit{KNOW} emphasized contextual political analysis, including identifying \textbf{K}ey players and incentives, understanding \textbf{N}orms and protocols, incorporating \textbf{O}ther perspectives, and considering \textbf{W}ildcards or low-probability events.} framework, an internally developed checklist for sound forecasting reasoning used in Good Judgment training during the ACE tournament \citep{chang2016developing}. Raters evaluated the rationale’s usefulness on a five-point scale ranging from (1) ``\emph{Not at all useful (No use of CHAMPS KNOW)}'' to (5) ``\emph{Extremely Useful (Great Integration of CHAMPS KNOW)}.'' Although these ratings are based on a single, domain-specific rubric, the rubric was developed and refined over four seasons of tournament operation, making it an informative benchmark.

\begin{figure}[t!]
\centering
\includegraphics[width=\textwidth]{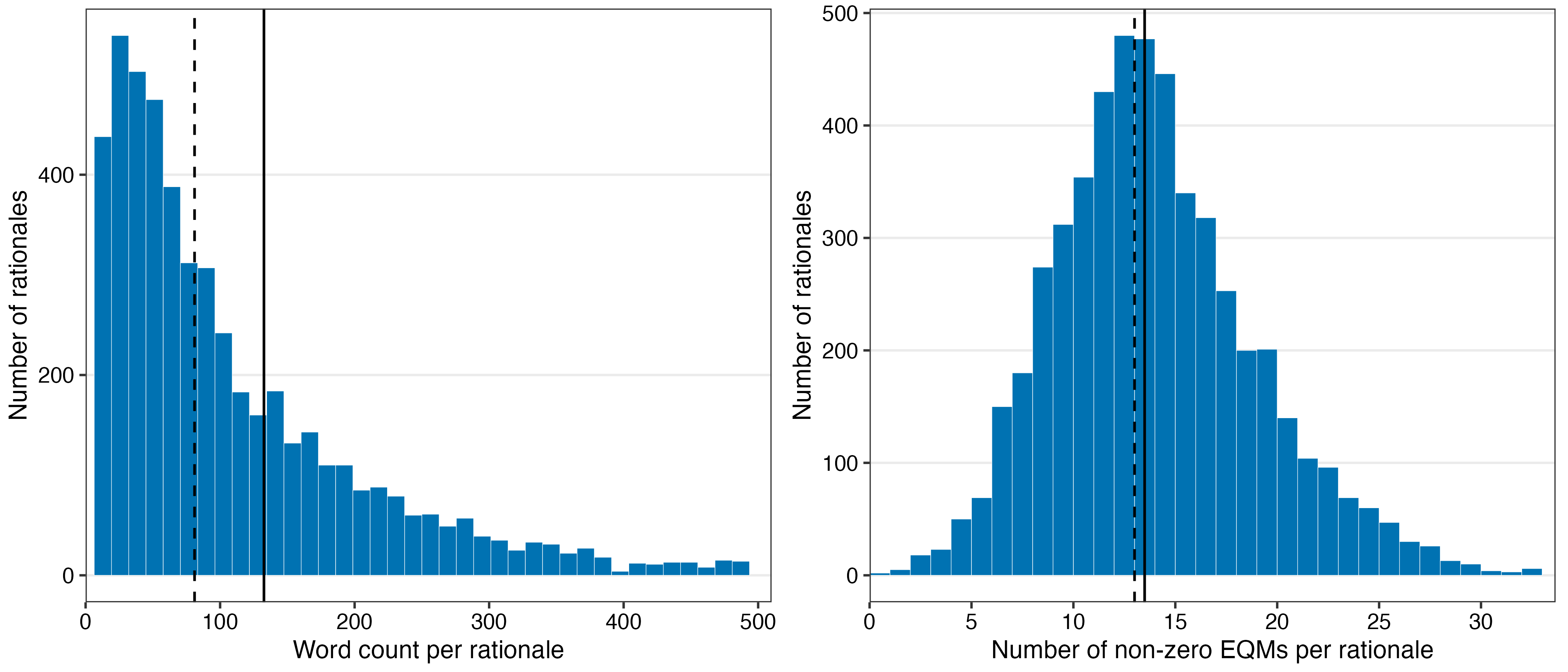}
\caption{
\emph{Distributions of rationale length and number of EQM patterns per rationale}. Rated-rationale sample ($N = 5{,}190$).
\emph{Left panel}: word count per rationale. \emph{Right panel}: number of EQM patterns with a score greater than zero. Solid vertical lines denote means; dashed vertical lines denote medians. See Figure~\ref{fig:hist_wordcount_nonzeroeqm} for the corresponding distributions in the main analysis sample.
}
\label{fig:ratings_stats}
\end{figure}

{\bf Sample construction.} The full ratings dataset contained 105,780 ratings across 9,831 rationales and 133 forecasting questions. We restricted attention to binary questions, rationales of at least ten words, and rationales that received at least three ratings. This filtering yielded 58,104 ratings on 5,190 rationales across 87 questions. Compared with the main analysis sample ($N = 55{,}463$), rated rationales were substantially longer and contained more EQM patterns. As shown in Figure~\ref{fig:ratings_stats}, mean rationale length in the rated sample was 132.6 words versus 59.6 words in the main sample ($d = 0.79$), while rated rationales contained an average of 13.5 non-zero EQM patterns versus 10.7 in the main sample ($d = 0.63$); both differences were highly statistically significant ($p < .001$). Forecast-level accuracy, however, was nearly identical. The mean $BS_{\text{diff}}$ score in the rated sample was $-0.06$ ($N = 5{,}190$, 95\% CI = [$-0.06$, $-0.05$]) versus $ -0.04$ in the main sample. Although this difference was statistically significant $(t = -3.42, p < .001)$, the effect size was very small $(d = 0.05)$.

\subsubsection{Human Ratings and Forecasting Accuracy}\label{subsec:HumansCor}

{\bf Forecast-level accuracy}. We first examined how strongly human ratings and EQM composite scores correlated with forecast-level accuracy. To avoid look-ahead bias, the EQM composite score for each Season 4 rationale was generated from a LASSO model trained on Seasons 1--3 only ($N = 28{,}542$), following the procedure in Section~\ref{subsec:exploratory_training}. Table~\ref{tab:human_eqm_correlations} reports the correlations. At the forecast level ($N = 5{,}190$), the EQM composite score correlated with our forecast-level accuracy measure, $BS_{\text{diff}}$, at $r = .23$, more than three times the correlation for average human ratings ($r = .07$); the difference was highly statistically significant ($p < .001$). The two measures were themselves correlated at $r = .17$, suggesting partial but limited overlap.

{\bf Forecaster-level accuracy}. We next examined whether the EQM advantage extends to forecaster-level accuracy. To do so, we trained the forecaster-level EQM composite model on Seasons 1--3 and applied it to Season 4 forecasters using the procedure in Appendix~\ref{subsec:exploratory_training}. We then restricted the Season 4 sample to forecasters with at least 10 rationales that each received three or more ratings  ($N = 141$), averaged both EQM composite scores and human ratings within each forecaster, and correlated each with forecaster-level accuracy, $ABS_{\text{norm}}$. Table \ref{tab:human_eqm_correlations} reports the results. The EQM composite score correlated with forecaster-level accuracy at $r = .50$ compared to $r = .40$ for the average human rating; the difference approached but did not reach conventional statistical significance ($p = .097$).\footnote{With a minimum of five rationales ($N = 242)$, the correlations were $r = .48$ for the EQM composite score and $r = .36$ for human ratings ($p = .028$);  with a minimum of one rationale ($N = 409)$, the results were $r = .42$ versus $r = .26$ ($p < .001$).}

\begin{table}[t!]
\centering
\begin{minipage}{0.92\textwidth}
\centering
\footnotesize
\caption{Correlations of the EQM composite score and average human ratings with accuracy}
\label{tab:human_eqm_correlations}
\renewcommand{\arraystretch}{1.1}
\setlength{\tabcolsep}{4pt}
\begin{tabular*}{\textwidth}{@{\extracolsep{\fill}}llccccc@{}}
\toprule
\textbf{Level} & \textbf{Metric} & \textbf{$n$} & \textbf{EQM} & \textbf{Human Rating} & \textbf{Between} & \textbf{EQM Better} \\
\midrule
Forecast
& $BS_{\text{diff}}$
& 5{,}190
& $r=.23,\ p<.001$
& $r=.07,\ p<.001$
& $r=.17,\ p<.001$
& $p<.001$ \\

Forecaster
& \ensuremath{ABS_{\text{norm}}}
& 141
& $r=.50,\ p<.001$
& $r=.40,\ p<.001$
& $r=.40,\ p<.001$
& $p=.097$ \\
\bottomrule
\end{tabular*}

\vspace{0.35em}
\scriptsize
\textit{Notes.} The forecast-level row reports correlations with $BS_{\text{diff}}$ among Season 4 rationales with human ratings. The forecaster-level row reports correlations with \ensuremath{ABS_{\text{norm}}} among forecasters with at least 10 questions for which the forecaster had a rationale with at least three human ratings. \emph{EQM} reports the correlation between the EQM composite score and the relevant accuracy measure. \emph{Human Rating} reports the correlation between average human rating and the relevant accuracy measure. \emph{Between} reports the correlation between the EQM composite score and average human rating. \emph{EQM Better} reports the $p$-value from a test of dependent correlations for whether EQMs outperformed average human ratings. The forecast-level EQM composite model was trained on Seasons 1--3 forecast-level data and applied to the subset of Season 4 rationales with human ratings; the forecaster-level EQM composite model was trained on Seasons 1--3 forecaster-level data and applied to Season 4 forecasters.
\end{minipage}
\end{table}

\begin{figure}[t!]
  \centering
  \includegraphics[width=\textwidth]{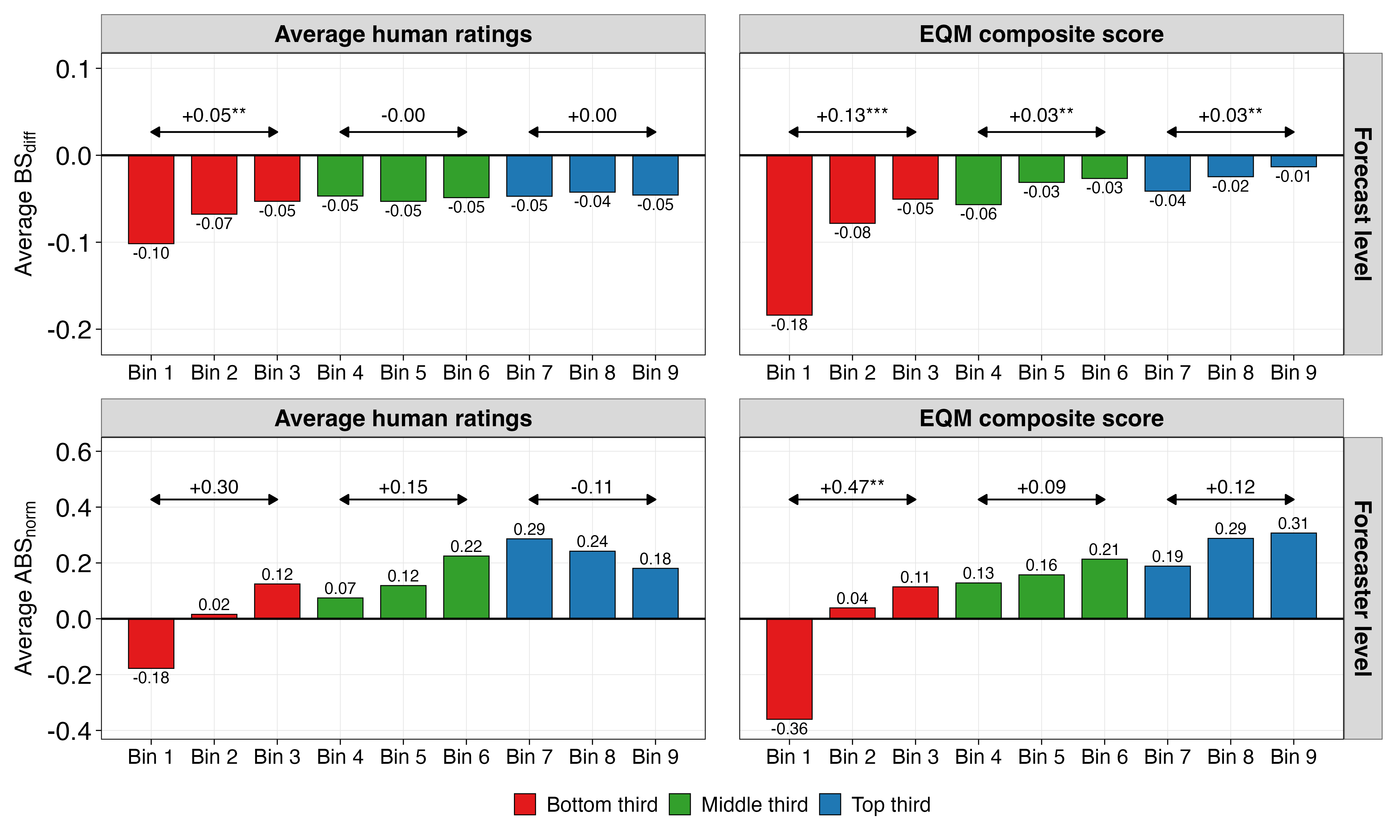}
  \caption{
        \emph{Mean accuracy across human-rating and EQM composite score bins}.
        Observations are sorted into nine equal-sized bins by the relevant measure. \emph{Left column}: bins formed by the average human rating. \emph{Right column}: bins formed by the EQM composite score. \emph{Top row}: forecast-level accuracy, measured by \ensuremath{BS_{\text{diff}}} ($N = 5{,}190$). \emph{Bottom row}: forecaster-level accuracy, measured by \ensuremath{ABS_{\text{norm}}} ($N = 141$). Positive values indicate better accuracy. Asterisks indicate $t$-tests comparing the outer bins within each tercile: * $p < .05$, ** $p < .01$, *** $p < .001$.
  }
  \label{fig:bin9_MOOF_both_levels}
\end{figure}

{\bf Where the signal is concentrated.} We next examined where in the score distribution human ratings and EQM composite scores were most informative. To do so, we sorted observations into nine equal-sized bins by each measure and examined average accuracy within each bin (Figure~\ref{fig:bin9_MOOF_both_levels}). 

At the forecast level, human ratings provided only weak differentiation. Average \(BS_{\text{diff}}\) rose from \(-0.10\) in Bin 1 to \(-0.05\) in Bin 3, a within-tercile difference of \(0.05\) (\(p = .001\)). However, there was no statistically significant difference in accuracy in  either the middle (\(p = .89\)) or top tercile (\(p = .92\)). By contrast, the EQM composite score displayed a sharper bottom-tercile increase, with the average \(BS_{\text{diff}}\) rising from \(-0.18\) in Bin 1 to \(-0.05\) in Bin 3, a within-tercile difference of \(0.13\) (\(p < .001\)). The corresponding within-tercile differences were smaller but still positive in both the middle (\(0.03\), \(p = .010\)) and top terciles (\(0.03\), \(p = .007\)).

At the forecaster level, both measures again showed largest gains in the bottom tercile. Average \(ABS_{\text{norm}}\) rose from \(-0.18\) in Bin 1 to \(0.12\) in Bin 3 for human ratings, a within-tercile difference of \(0.30\), although this difference was not statistically significant (\(p = .116\)). Human ratings continued to increase in the middle tercile (\(+0.15\), \(p = .128\)) but declined in the top tercile (\(-0.11\), \(p = .094\)). For the EQM composite score, the average \(ABS_{\text{norm}}\) rose from \(-0.36\) in Bin 1 to \(0.11\) in Bin 3 (\(0.47\), \(p = .009\)). Compared to the human ratings, the EQM composite score showed a consistently positive pattern, increasing in both the middle (\(0.09\), \(p = .285\)) as well as the top terciles (\(0.12\), \(p = .178\)), although these within-tercile differences were not statistically significant.

Overall, both measures were better at identifying weak forecasts and forecasters than at distinguishing among stronger ones. However, the EQM composite score identified the weakest more sharply than human ratings, and accuracy kept improving across higher EQM bins while it flattened or even reversed across human-rating bins.

\subsubsection{Correlates of Human Ratings}

The previous section showed that human ratings were less predictive of accuracy than EQM composite scores. A natural follow-up  question is what human raters respond to when evaluating rationales.

\begin{figure}[t!]
\centering
\includegraphics[width=0.95\linewidth]{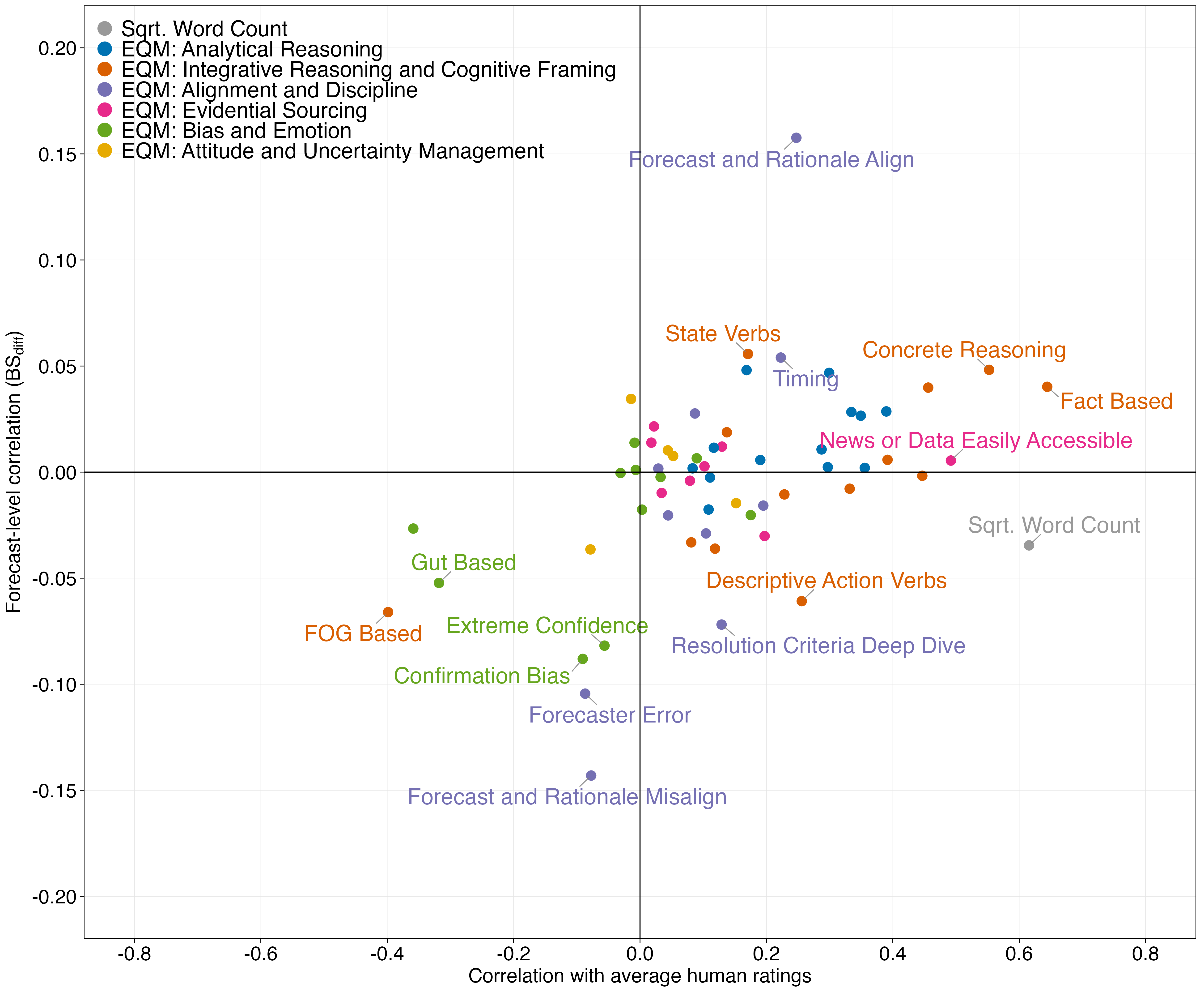}
\caption{
\emph{Pattern-level correlations with average human ratings and forecast-level accuracy}. Rated-rationale sample ($N = 5{,}190$).
Each point represents one EQM pattern, with colors indicating EQM family; \emph{Sqrt. Word Count} is shown separately in gray. \emph{$x$-axis}: correlation with average human ratings. \emph{$y$-axis}: correlation with forecast-level accuracy, measured by \ensuremath{BS_{\text{diff}}}. Vertical and horizontal reference lines mark zero correlation. Patterns in the upper-right quadrant are favored both by human raters and forecast-level accuracy, whereas patterns in the lower-left quadrant are disfavored by both.
}
\label{fig:eqm_rating_vs_accuracy}
\end{figure}

Compared to accuracy, average human ratings were substantially easier to predict from EQM patterns and rationale length. A LASSO model fit via the training paradigm in Appendix~\ref{subsec:within_sample_training} achieved a correlation of $r = .74$ with human ratings, with the coefficients shown in Appendix~\ref{app:lasso_human_ratings}. Individual EQM patterns also showed large bivariate correlations with human ratings. Figure~\ref{fig:eqm_rating_vs_accuracy} plots each EQM pattern's correlation with average human ratings ($x$-axis) against its correlation with forecast-level accuracy ($y$-axis). The square root of word count, denoted as \textit{Sqrt. Word Count}, is also included as a non-EQM comparison feature. Features in the upper-right quadrant were both rewarded by human raters and positively associated with forecast-level accuracy, while features in the lower-left quadrant were both penalized by raters and negatively associated with forecast-level accuracy.

At the pattern level, human raters were broadly directionally correct. Patterns associated with higher forecast-level accuracy, such as \textit{Fact Based}, \textit{Concrete Reasoning}, and \textit{Timing}, tended to receive higher ratings. Conversely, patterns associated with worse accuracy, such as \textit{FOG Based}, \textit{Confirmation Bias}, \textit{Simplification Bias}, and \textit{Gut Based}, tended to receive lower ratings. Across features, the correlation was $r = .45$, suggesting moderate directional alignment between correlates of human ratings and of forecast-level accuracy.

However, this directional alignment at the pattern level did not translate into strong overall correlations with accuracy: average human ratings correlated only weakly with forecast-level accuracy ($r=.07$). The gap appears to reflect differences in \emph{weighting}: raters did not weight features in proportion to their association with forecast-level accuracy. For example, human ratings were most strongly correlated with \textit{Fact Based} ($r=.64$) and \textit{Sqrt. Word Count} ($r=.62$; raw word count alone yielded $r=.48$). \textit{Fact Based} is also among the stronger individual correlates of forecast-level accuracy. Rationale length, as measured by \textit{Sqrt. Word Count}, by contrast, was essentially unrelated to forecast-level accuracy in Study 1 ($|r| < .01$) and slightly negative in this study ($r = -.03$), yet it was one of the strongest predictors of human ratings. Thus, raters appeared to weight rationale length almost as much as any EQM pattern, even though length carried little information about forecast-level accuracy.

To examine these relative weights more directly, we estimated two OLS regressions in the rated-rationale sample. Both regressions used the 50 EQM patterns and the square root of word count as predictors, but differed in the outcome: one predicted forecast-level accuracy, while the second predicted average human ratings. All predictors were standardized prior to fitting, and each coefficient vector was normalized by the sum of its absolute values, yielding relative weights that are comparable across the two models.

\begin{figure}[t!]
\centering
\includegraphics[width=0.95\linewidth]{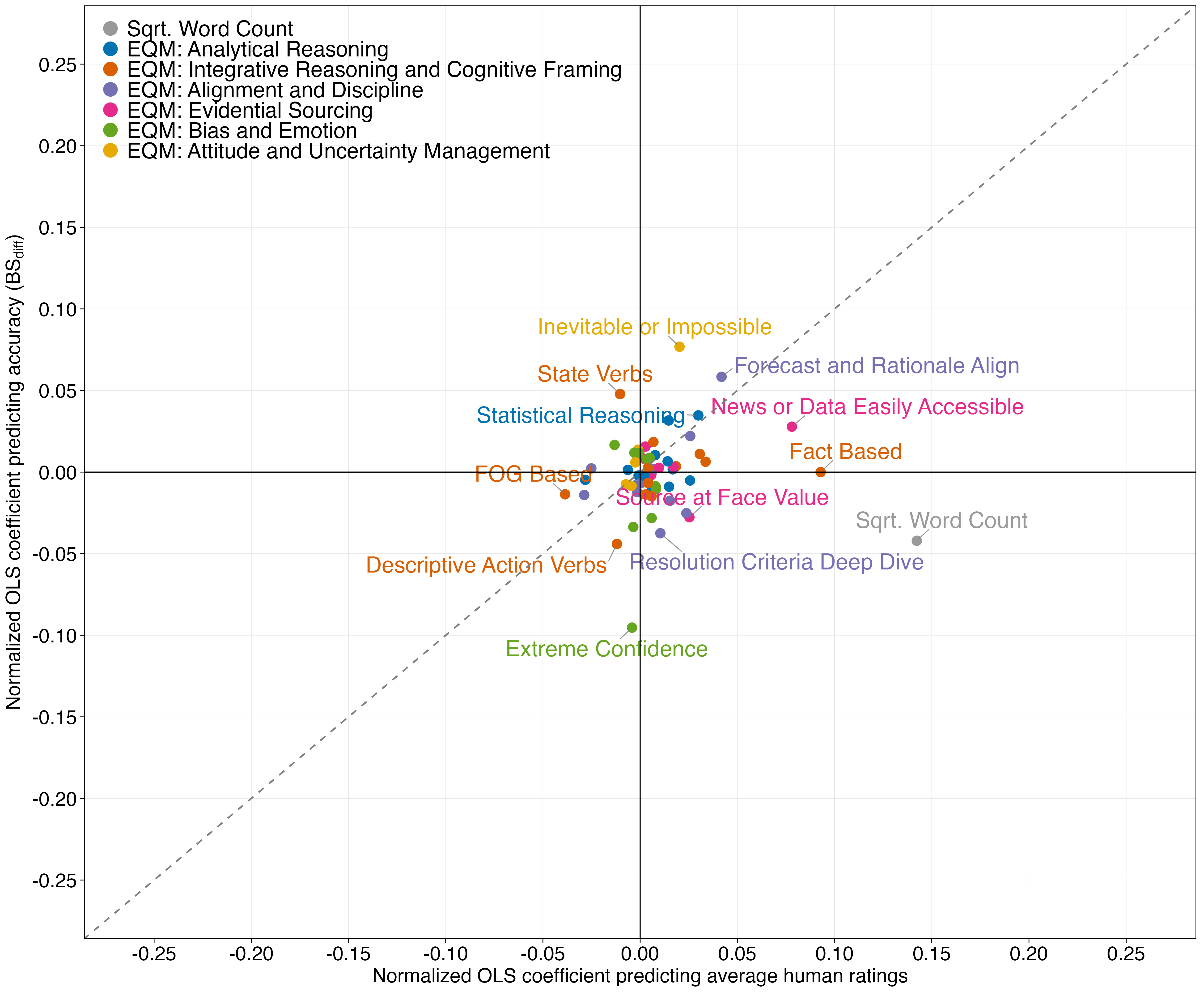}
\caption{
\emph{Normalized multivariate regression weights for average human ratings and forecast-level accuracy}.
We estimated two OLS regressions using the same predictors: the 50 EQM patterns and the square root of word count (\emph{Sqrt. Word Count}). Predictors were standardized before model fitting. Coefficients from each model were then normalized by dividing by the sum of their absolute values, making the two coefficient vectors comparable. \emph{$x$-axis}: normalized coefficient from the model predicting average human ratings. \emph{$y$-axis}: normalized coefficient from the model predicting forecast-level accuracy, measured by \ensuremath{BS_{\text{diff}}}. The dashed diagonal represents equal proportional weighting by raters and by the accuracy model; points off the diagonal represent a mismatch in weighting.
}
\label{fig:eqm_rating_vs_accuracy_ols}
\end{figure}

Figure~\ref{fig:eqm_rating_vs_accuracy_ols} plots the resulting normalized coefficients for average human ratings ($x$-axis) against the normalized coefficients for forecast-level accuracy ($y$-axis). The dashed diagonal represents equal proportional weighting by human ratings and forecast-level accuracy. The multivariate results sharpen the pattern previously observed in Figure~\ref{fig:eqm_rating_vs_accuracy}. \textit{Sqrt. Word Count} received the largest absolute weight (positive) in the human-rating model but a small negative weight in the accuracy model. \textit{Fact Based} also obtained a relatively large positive weight in the human-rating model but little weight in the accuracy model. This finding does not imply that factual reasoning is unimportant for accuracy; rather, raters overweighted \textit{Fact Based} relative to its marginal contribution to forecast-level accuracy, conditional on the EQM pattern set and rationale length. By contrast, \textit{Extreme Confidence} received the largest absolute weight (negative) in the accuracy regression but near-zero weight in the human ratings regression. The correlation between the two normalized coefficient vectors in Figure~\ref{fig:eqm_rating_vs_accuracy_ols} is much smaller, $r = .06$, far below the bivariate pattern-level correlation of $r = .45$.

Taken together, Figures~\ref{fig:eqm_rating_vs_accuracy} and~\ref{fig:eqm_rating_vs_accuracy_ols} suggest that human raters were directionally sensitive to many of the same rationale features that predicted forecast-level accuracy, but not well calibrated to their relative importance. Raters overweighted rationale length and fact-based presentation relative to their accuracy signal, while underweighting patterns such as \emph{Extreme Confidence} that carried substantial negative accuracy signal.

\subsubsection{Human Ratings Reliability}\label{subsec:Reliability}

Finally, we examined inter-rater reliability. Among pairs of raters who evaluated at least 100 common rationales (850 rater pairs), agreement levels were modest. On the original five-point scale, average Cohen’s $\kappa$ was .24 and average inter-rater correlation was $r = .33$. Collapsing the scale into three categories (1--2, 3, 4--5), comparable to the three-point EQM scale used for LLM scoring, yielded average $\kappa = .20$ and average $r = .26$. Collapsing further to a binary distinction between ratings at or below 3 versus above 3 produced still lower reliability, with average $\kappa = .12$ and average $r = .15$, and average percent agreement of 68.6\%. For context, as reported in Section~\ref{subsec:EQM_sensitivity}, GPT-4o and Gemini 2.5 Pro achieved $\kappa = .56$ and agreement of 82\% when scoring the EQM patterns on a two-point scale. Overall, these results indicate that perceived ``rationale quality''---operationalized here as usefulness and integration of \textit{CHAMPS KNOW}---was a noisy construct even among experienced forecasters operating within a shared forecasting environment.

\subsubsection{Conclusion}\label{subsec:HumansConclusion}

Study 3 yielded five findings. First, average human ratings correlated weakly with forecast-level accuracy at $r=.07$, compared with $r=.23$ for the EQM composite score. Second, the patterns most strongly weighted by human raters included surface features that carried little information about forecast-level accuracy: the square root of word count alone correlated with average human ratings at $r=.62$ but was essentially uncorrelated with forecast-level accuracy. Third, EQM composite scores were more predictive of forecaster-level accuracy ($r = .50$) than human ratings ($r = .40$). Fourth, both human ratings and EQM composite scores were more predictive of accuracy at the bottom of their respective distributions than at the top. Finally, inter-rater reliability was low (average Cohen's $\kappa=.24$ on the original five-point scale; binary-bin pairwise agreement of 68.6\%), substantially below the cross-model agreement among LLMs scoring EQM patterns.

These findings are specific to the \textit{CHAMPS KNOW} rubric and the particular pool of raters. A natural next question is whether they generalize to other rating frameworks or rater pools. Study 4 tests one piece of this generalization by applying both frameworks to an independent dataset with a different human-rating rubric.

\FloatBarrier

\subsection{Study 4: Do the Results Transfer Out of Sample?}

\label{subsec:team_dynamics}

\subsubsection{Introduction}

Studies 1–3 established that EQMs outperform both pre-LLM methods and human ratings within the ACE tournament. A natural concern is whether these results depend on the specific dataset, question domain, or elicitation format. Study 4 addresses this worry by testing whether ACE-trained EQM composite models transfer when applied, without retraining, to an independent forecasting dataset.

The out-of-sample dataset comes from the \emph{Team Dynamics} (TD) study, a multi-stage forecasting experiment conducted by the Forecasting Research Institute to examine how team deliberation affects forecasting accuracy \citep{barker2025teamdynamics}.\footnote{At the time of writing, only the cited pre-registration is publicly available. A working paper with full results is forthcoming.} For the present analysis, we focus on the Stage 2 individual forecasts and associated rationales, collected prior to any team interaction or updating. The TD study differs from the ACE tournament in several respects: different forecasting questions, different forecasters, a different elicitation format (quantile forecasts rather than probabilities), and a different human-rating rubric. In addition, the TD questions resolved in summer 2025, well after the October 2023 knowledge cutoff of the GPT-4o model used for scoring \citep{openai2024gpt4ocard}. The LLM therefore could not have memorized the realized outcomes of these questions during training, mitigating the risk that EQM scores predict accuracy by exploiting memorized resolutions instead of evaluating reasoning quality.

\subsubsection{Data and Scoring}

The full Stage 2 TD dataset contained 61,796 rationales from 3,143 forecasters across 24 unique questions.\footnote{Four questions required two forecasts at different time horizons; in those cases, a single rationale supported both forecasts, resulting in up to 20 rationales per forecaster.} The TD dataset also included human ratings of rationale quality, elicited on a 1--5 scale similarly to the ACE human ratings in Section~\ref{subsec:HumansvsEQM}, although TD ratings measured \textit{helpfulness} rather than integration of \emph{CHAMPS KNOW}.\footnote{Participants were asked in Stage 3 to rate rationales from Stage 2 on a five-point scale according to how helpful they found them for updating their own Stage 3 forecasts. Our analysis uses the Stage 2 forecasts only, but we include the Stage 3 ratings for our analyses involving human ratings.} Restricting attention to rationales with at least 10 words and at least three human ratings yielded 42,809 rationales with corresponding forecasts. The questions resolved in summer 2025, and all rationales were scored in December 2025 using GPT-4o.

Forecasts in TD were elicited as quantiles (25\%, 50\%, and 75\%) rather than probabilities as in the ACE tournament. Therefore, forecast accuracy was evaluated using the $S$-score, also known as pinball loss, for quantile forecasts \citep{gneiting2007strictly, jose2009evaluating}\footnote{The $S$-score is the average quantile loss across elicited quantiles. For quantile level $\alpha$, submitted quantile $q_\alpha$, and realized outcome $x$, the per-quantile $S$-score is given by $S_\alpha(q_\alpha,x)=\alpha\max(x-q_\alpha,0)+(1-\alpha)\max(q_\alpha-x,0)$. We average this quantity across quantile levels; lower values indicate more accurate forecasts, with zero loss when the submitted quantile equals the realized value.}. Lower $S$-scores indicate higher accuracy and a perfect forecast receives a score of 0. Because $S$-scores are in the units of the questions and thus on disparate scales, each forecast's $S$-score was converted into a within-question rank-based measure in which higher values correspond to better performance (1 = most accurate, 0 = least accurate). For questions with two forecasts tied to a single rationale, the two scores were averaged to produce a single forecast-level accuracy outcome. Accuracy ranks were computed using the full dataset ($N = 61{,}796$), but only the filtered sample (10 or more words and 3 or more ratings; $N = 42{,}809$) was retained for analysis. Forecaster-level accuracy was defined as the average of the rank-based measures over questions with rationales of 10 or more words and 3 or more ratings. As in Studies 1 and 3, we restricted forecaster-level analysis to forecasters with at least ten qualifying rationales ($N = 2{,}221$).

Because the TD study used quantile forecasts rather than probabilities, selected EQM pattern descriptions required minor adjustments. Specifically, the patterns \textit{Forecast and Rationale Align}, \textit{Forecast and Rationale Misalign}, \textit{Forecaster Error}, \textit{Adjusting Up}, and \textit{Adjusting Down} reference forecast probabilities in their descriptions (e.g., ``the outcome with the highest forecast probability'' or ``a baseline probability''). For Study 4, these descriptions were adapted to refer to quantile forecasts instead; see Appendix~\ref{app:td_pattern_adaptations} for details. The broader scoring logic remained unchanged. 

\subsubsection{Results: Out-of-Sample}

All EQM composite scores in this section were generated by LASSO models trained on ACE data and applied to the TD dataset \emph{without} retraining. 

{\bf Forecast level.} As shown in Table~\ref{tab:study4_transfer_results}, the ACE-trained EQM composite score correlated with forecast-level accuracy at $r = .16$, close to the $r = .19$ obtained in Study 1 (Table~\ref{tab:hypothesis_tests}). The average helpfulness rating correlated at $r = .19$, slightly but statistically significantly higher than the EQM composite score ($p < .001$). The correlation between EQM composite scores and average helpfulness ratings in this subset was $r = .26$, again indicating only partial overlap between the two indicators.

{\bf Forecaster level.} Among forecasters with at least ten rationales of ten or more words and at least three ratings per rationale ($N = 2{,}221$), the ACE-trained EQM composite score correlated with forecaster-level accuracy at $r = .46$, only somewhat below the $r = .51$ found in Study 1 (Table~\ref{tab:hypothesis_tests}). This modest attenuation---despite substantial differences in questions, forecasters, and elicitation format---suggests that the EQM framework captures reasoning patterns that generalize across forecasting settings. The averaged human rating per forecaster correlated with forecaster-level accuracy at only $r = .33$. Hence, the forecaster-level finding that EQM composite scores are more predictive than human ratings replicated in the TD dataset.

\begin{table}[t!]
\centering
\begin{minipage}{0.98\textwidth}
\centering
\footnotesize
\caption{Out-of-sample correlations with accuracy in the Team Dynamics dataset}
\label{tab:study4_transfer_results}
\renewcommand{\arraystretch}{1.1}
\setlength{\tabcolsep}{4pt}
\begin{tabular*}{\textwidth}{@{\extracolsep{\fill}}llllll@{}}
\toprule
\textbf{Level} & \textbf{$n$} & \textbf{Out-of-sample EQM} & \textbf{TD Human Ratings} & \textbf{Between} & \textbf{Better Measure} \\
\midrule
Forecast
& 42{,}809
& $r=.16$ (Study 1 $r=.19$)
& $r=.19$ (Study 3 $r=.07$)
& $r=.26$
& Human rating ($p<.001$) \\

Forecaster
& 2{,}221
& $r=.46$ (Study 1 $r=.51$)
& $r=.33$ (Study 3 $r=.40$)
& $r=.46$
& EQM ($p<.001$) \\
\bottomrule
\end{tabular*}

\vspace{0.5em}
\scriptsize
\textit{Notes.} Rows correspond to different accuracy measures and sample restrictions. \emph{Out-of-sample EQM} reports the correlation between the ACE-trained EQM composite score and TD accuracy, applied without retraining (see Appendix~\ref{subsec:exploratory_training}); the corresponding Study 1 EQM correlations are shown in parentheses. \emph{TD Human ratings} reports the correlation between average TD helpfulness ratings and TD accuracy; the corresponding Study 3 human-rating correlations are shown in parentheses. \emph{Between} reports the correlation between the EQM composite score and average TD human rating. \emph{Better Measure} reports which measure had the stronger correlation with accuracy, with the one-sided $p$-value from a dependent-correlation test in parentheses.
\end{minipage}
\end{table}

\begin{figure}[t!]
  \centering
  \includegraphics[width=\textwidth]{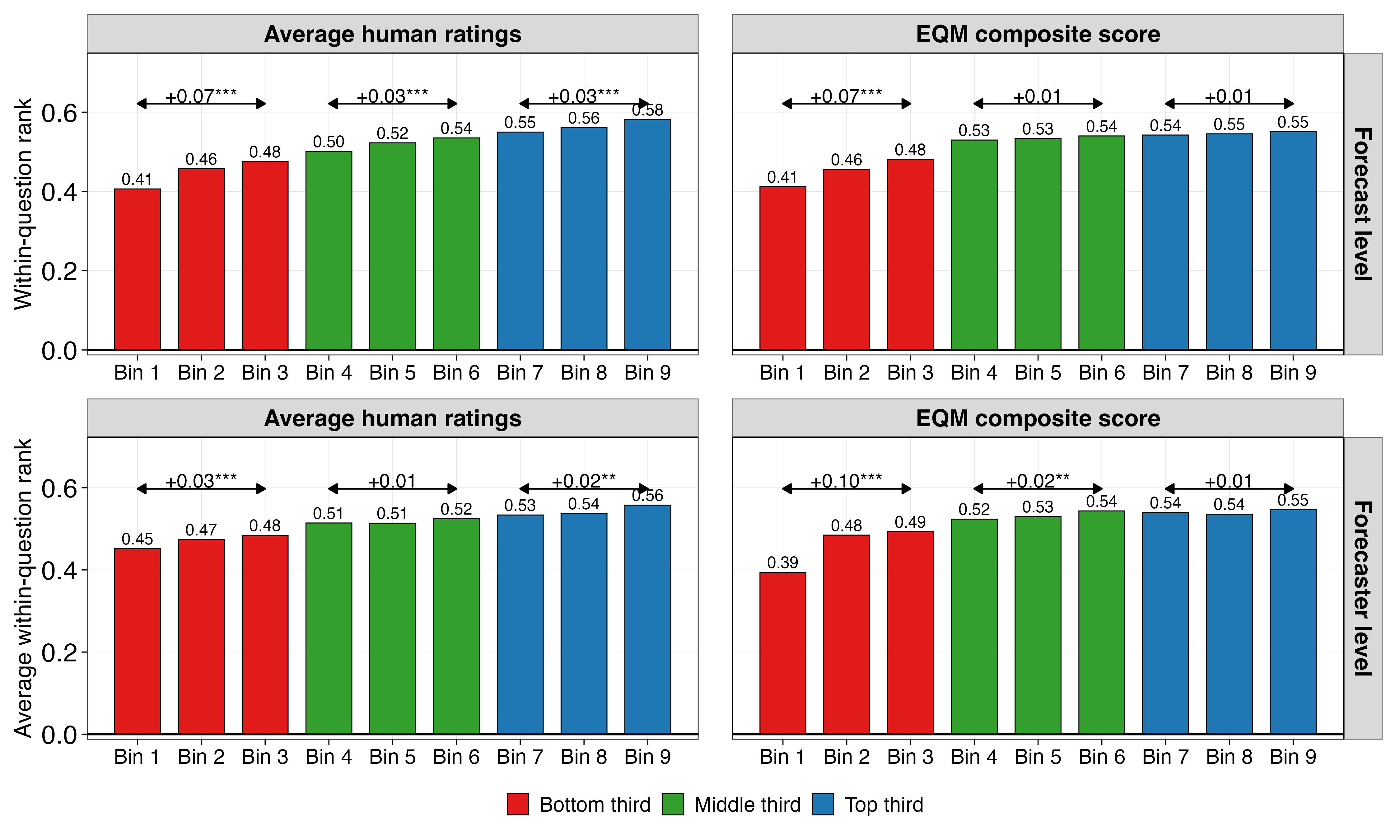}
  \caption{
    \emph{Mean accuracy across human-rating and EQM composite score bins in the Team Dynamics dataset}.
    Observations are sorted into nine equal-sized bins by the relevant measure. \emph{Left column}: bins formed by average human rating. \emph{Right column}: bins formed by the EQM composite score. \emph{Top row}: forecast-level accuracy, measured by within-question rank ($N = 42{,}809$). \emph{Bottom row}: forecaster-level accuracy, measured by average within-question rank ($N = 2{,}221$). Higher values indicate better accuracy. Asterisks indicate $t$-tests comparing the outer bins within each tercile: * $p < .05$, ** $p < .01$, *** $p < .001$.
  }
  \label{fig:bin9_TD_both_levels}
\end{figure}

{\bf Where the signal is concentrated.}  As in Study 3, we sorted observations into nine equal-sized bins by each measure and examined average accuracy within each bin (Figure~\ref{fig:bin9_TD_both_levels}). 

At the forecast level, both measures showed largest gains in the bottom tercile (a within-tercile increase of 0.07 for both, $p < .001$). Human ratings continued to separate forecasts in the middle (\(0.03\), \(p < .001\)) and top tercile (\(0.03\), \(p < .001\)). By contrast, the EQM composite score showed smaller increases in the middle (\(0.01\), \(p = .059\)) and top tercile (\(0.01\), \(p = .120\)) that were borderline but not statistically significant.

At the forecaster level, the asymmetric pattern was stronger for the EQM composite score. Average within-question rank increased by \(+0.10\) in the bottom tercile for the EQM composite score (\(p < .001\)). The EQM composite score also increased in the middle tercile (\(+0.02\), \(p = .007\)), but showed little additional gain in the top tercile (\(+0.01\), \(p = .375\)). Human ratings showed a smaller increase in the bottom tercile (\(+0.03\), \(p < .001\)), no statistically significant difference in the middle tercile (\(+0.01\), \(p = .189\)), and a modest increase in the top tercile (\(+0.02\), \(p = .002\)).

Overall, the TD results replicate the asymmetry found in Study 3: EQM composite scores were most informative at the lower end of the distribution. At the forecast level, human ratings provided steadier gains across the full range, but at the forecaster level, EQM composite scores more sharply identified the weakest performers.

{\bf Correlates of human ratings.} Finally, we tested whether the correlates of human ratings were consistent across our two studies. A model trained on the $N = 5{,}190$ averaged human ratings and EQM profiles from Section~\ref{subsec:HumansvsEQM} transferred strongly to the averaged TD helpfulness ratings, correlating at $r = .54$. As in Study 3, TD helpfulness ratings were strongly associated with word count ($r = .44$) and \emph{Sqrt. Word Count} ($r = .53$). In fact, \emph{Sqrt. Word Count} was the strongest individual correlate of the averaged human ratings in Study 4, followed by \textit{Fact Based} ($r = .43$). These findings reinforce the pattern from Study 3: human helpfulness judgments are easier to predict than forecast accuracy and are primarily associated with rationale length and fact-based reasoning.

\subsubsection{Conclusion}

Study 4 shows that the EQM framework transferred to a new forecasting environment with different questions, forecasters, elicitation format, and human-rating rubric. The TD questions resolved in summer 2025, well after the October 2023 training data cutoff of the GPT-4o model used for scoring \citep{openai2024gpt4ocard}, mitigating the concern that EQM scores exploit memorized resolutions. Out-of-sample correlations were similar to the corresponding ACE correlations at both levels of analysis (forecast level: $r=.16$ vs. $r=.19$ in ACE; forecaster level: $r=.46$ vs. $r=.51$ in ACE), and the asymmetric signal pattern replicated clearly in the TD data. Only minor pattern adjustments were needed to accommodate the quantile elicitation format, suggesting the framework is portable across forecasting settings.

Study 4 also qualifies one human-ratings finding from Study 3. In the TD dataset, human helpfulness ratings were more strongly correlated with forecast-level accuracy than EQM composite scores (EQMs: $r=.16$; human ratings: $r=.19$), whereas in ACE EQMs had a large advantage (EQMs: $r=.23$; human ratings: $r=.07$). At the forecaster level, however, EQMs remained more predictive of accuracy than human ratings.

\FloatBarrier

\section{Conclusion}\label{sec:conclusions}

This manuscript studies how written explanations link to judgment accuracy. Using forecasting tournaments as the empirical setting \citep{tetlock2014tournaments,tetlock2015superforecasting}, we introduce \emph{Explanation Quality Markers} (EQMs), a theory-guided framework that uses large language models to score interpretable reasoning patterns in text. Across four studies, EQM composite scores are more strongly correlated with forecasting accuracy than pre-LLM text-analysis methods. Notably, EQMs identify low-accuracy forecasts and forecasters more reliably than they identify the highest-accuracy ones, and the comparison with human ratings further suggests that what makes an explanation look good is not always what makes it informative about accuracy.

{\bf Limitations.} Several limitations should be kept in mind. First, written rationales are an imperfect proxy for reasoning. Forecasters may omit reasoning that is tacit, difficult to verbalize, or simply left out due to time constraints. Conversely, rationales may include post hoc justifications that do not reflect the actual reasoning process. These considerations limit what can be inferred about reasoning from text alone.

Second, our findings are correlational. We show that EQM scores are associated with realized accuracy, not that the reasoning patterns measured by EQMs causally improve judgment. A stronger test would require an experiment. For example, in a controlled setting, researchers could train forecasters to use techniques described by EQM patterns (e.g., \textit{Statistical Reasoning}) and ask forecasters to record their reasoning in rationales. Researchers could then measure the treatment effects on both accuracy and EQM pattern scores. We approximate such a study in Appendix~\ref{app:training-eqm} using an experimental training condition in the ACE tournament. We find that a randomized probability training shifted EQM patterns in the expected directions, increasing scores for \emph{Statistical Reasoning}, \emph{Fact Based}, and \emph{Best Practices}.

Finally, the main analyses used geopolitical forecasting data from a structured tournament setting, and the transfer study (Study 4), while encouraging, remains within the forecasting domain. Generalizing EQMs to other domains therefore remains an open question.

{\bf Future extensions.} EQMs are best viewed as a portable measurement framework rather than a finished instrument and can be adapted in several ways. One extension is for researchers or practitioners to revise the pattern set for a new domain or forecasting setting. A further possibility is algorithmic search for new patterns, in which LLMs discover recurring reasoning behaviors from subsets of written judgments and translate them into EQM patterns; we are exploring this approach in a related research project.

Another potential extension is to adapt EQMs to new forms of written analytic judgment such as  intelligence assessments, policy briefs, or medical recommendations. Some of these settings may still involve judgments that can ultimately be evaluated for accuracy. Others may place greater weight on persuasion or explainability, which could serve as alternative target outcomes.

This broader range of settings underscores the central promise of EQMs. Many consequential judgments are already accompanied by written explanations, but those explanations are rarely measured systematically. EQMs provide a way to turn such text into explicit, testable features of judgment quality.

\newpage

\appendix

\section{Supplementary Methods}\label{appendix:Methods}
\subsection{Correspondence Between the Pre-Registration and the Present Paper}\label{appendix:Pre-reg}

This appendix briefly describes the correspondence between the pre-registration and the present paper.

\subsubsection{Core elements that remained the same}

The pre-registered study was submitted on 10 March 2025 and is largely represented here as Study~1, which preserves the approach of human-derived LLM-scored patterns (denoted herein as the EQM approach, and as the ``New-School'' approach in the pre-registration) and the comparison with the pre-LLM models (denoted the ``Old-School'' approach in the pre-registration). The paper retains the same ACE dataset, the same main filtering rules, the same two primary dependent variables, and the same basic model-comparison framework of the pre-registration. Specifically, the analysis uses ACE forecasting data, restricts attention to first rationales, applies the minimum 10-word threshold, and focuses on binary questions. The two main dependent variables remain forecast-level accuracy relative to the local crowd (\textit{BS}$_{\text{diff}}$) and forecaster-level normalized seasonal accuracy (\textit{ABS}$_{\text{norm}}$). The main statistical comparison also remains the same: EQM composite scores are compared against the pre-LLM benchmark using the held-out LASSO prediction framework and dependent-correlation tests described in the pre-registration. 

The pre-registration records a near-final version of pre-LLM features used in the analyses and states that approximately 60 SME-derived patterns would be scored, corresponding to what we refer to here as the EQM pattern set. It also reserved the right to adjust the feature set up to the point at which the full study sample was scored. To develop the LLM batch-scoring pipeline for the EQM pattern set, we tested the pipeline on 10{,}000 rationales, representing fewer than 20\% of the rationales in the main sample. The final EQM scores for the full study sample were returned by the LLM on 20 March 2025 and used for the analyses reported here.

\subsubsection{Elements that were added or changed}

Several features of the present paper extend beyond the original pre-registration.

First, two authors were added after the pre-registration: Simas Ku\v{c}inskas and Nadja Flechner. These two authors assisted with both the analyses and writing.

Second, the pre-registration envisioned a single paper comparing three ``schools'' of approaches: Old-School, New-School, and Future-School. The present paper focuses on the New-School/EQM approach and its comparison to the Old-School/pre-LLM approach. The Future-School approach---that is, the automated discovery of LLM-derived hypotheses and patterns---is no longer included in this paper and is instead developed in a forthcoming companion paper that is under development. Relatedly, this paper does not include the originally proposed comparisons involving the superset of all three schools.

Third, the pre-registration explicitly allowed follow-up exploratory analyses if the New-School/EQM approach outperformed the Old-School/pre-LLM approach. The present paper makes use of that provision, adding several robustness and methodology analyses to Study 1 that were not part of the original pre-registration, including inter-model reliability checks, prompt-sensitivity analyses, single-pattern scoring ablations, alternative benchmark constructions, binarized scoring checks, and a nonlinear XGBoost comparison. In addition, Studies~2--4 were added after the main Study~1 results and should therefore be understood as exploratory extensions rather than part of the original confirmatory core.

Fourth, a minor data-processing deviation occurred in the construction of the local-crowd benchmark. The pre-registration proposed assigning the first forecast on each question a naive local-crowd value of 0.5. In the implemented analysis, these first forecasts were instead omitted because no prior local-crowd forecast was yet available. This exclusion affects only 27 forecast-rationale pairs---less than 0.05\% of the 55{,}463 observations in Study~1---and does not materially change the results.

Finally, the in-sample held-out LASSO framework used for Study~1 matches the pre-registration in substance in Appendix~\ref{subsec:within_sample_training}, but the paper adds across-season and across-dataset out-of-sample LASSO applications for exploratory purposes in Appendix~\ref{subsec:exploratory_training}. These include season-over-season prediction within ACE and transfer from ACE to the Team Dynamics dataset.

\clearpage

\subsection{Data Filtering and Scoring Details}
\label{appendix:FilteringScoring}

This appendix outlines our pre-registered data filtering and scoring logic. We employ three main filters, as described in Section~\ref{subsec:data_filtering}: rationales must be at least 10 words, rationales must be the first provided rationale per forecaster on each question, and the question must be binary. The decision to focus on first rationales and to exclude rationales shorter than 10 words is because very short rationales with only a few words are incapable of explaining reasoning, and many short rationales in the dataset were selected from a menu of stock phrases (e.g., ``Affirming prior forecast'' appears over 37,000 times in the dataset). In total, the ten most frequent short phrases accounted for approximately 175,000 out of 424,764 total rationales, all under 10 words, making a 10-word threshold an effective means of excluding non-informative entries. 

First rationales are preferred because they tend to be more reflective of forecasters' underlying analytical strategies. In contrast, follow-up rationales were typically more procedural, and frequently cited the passage of time. Finally, filtering to binary questions---as opposed to multinomial questions---simplifies scoring.

In calculating \ensuremath{BS_{\text{diff}}}, the goal of the local crowd forecast was not to construct the most accurate possible aggregate forecast, as there are many reasonable ways to define an aggregate. Instead, the local crowd was intended to provide a rolling benchmark for question difficulty. The local crowd forecast included all forecasts, not only those associated with rationales, because filtering on rationales was performed after the local crowd forecast was calculated. We also retained only each forecaster's final forecast on a given day, so that multiple same-day updates by the same forecaster did not disproportionately influence the local crowd.\footnote{Cases in which a forecaster's forecast from a previous day entered the local crowd for one of their later rationales were very rare, occurring only 35 times out of the 55,463 rationales in Study 1, or 0.06\% of the time.} Finally, when 11 prior forecasts were not available, we used the median of the available prior forecasts, excluding the first forecast for each question from the analysis because no local crowd forecast was available.\footnote{This departs slightly from the pre-registration, which proposed assigning the first forecast on each question a naive local-crowd value of 0.5. This exclusion removes only 27 forecast-rationale pairs---or less than 0.05\% of the 55,463 data points from Study 1---which does not materially change the results.}

For forecaster-level analyses, we use \ensuremath{ABS_{\text{norm}}}, a normalized season-level accuracy measure. The measure is constructed in three steps. First, for each question, we compute each forecaster's daily Brier score over the lifetime of the question. If the forecaster had submitted a prediction by a given day, the daily Brier score is based on the forecaster's most recent forecast. If the forecaster had not yet submitted a forecast, the forecaster receives the median daily Brier score across participating forecasters for that day. These daily Brier scores are then averaged over the lifetime of the question to produce the forecaster's question-level Brier score.

Second, we normalize each forecaster's question-level Brier score relative to the distribution of forecaster scores on the same question:

\begin{equation}
BS_{\text{norm}} =
\frac{\text{Average Score} - \text{Forecaster's Score}}
{\text{Standard Deviation}} .
\end{equation}

Because lower Brier scores indicate better accuracy, subtracting the forecaster's score from the question average makes higher values correspond to better performance. Positive values of \ensuremath{BS_{\text{norm}}} therefore indicate above-average accuracy on a given question, while negative values indicate below-average accuracy.

Third, for each forecaster-season, we average \ensuremath{BS_{\text{norm}}} across all questions on which the forecaster participated, yielding \ensuremath{ABS_{\text{norm}}}. The end-to-end construction of \ensuremath{ABS_{\text{norm}}} mimics preceding analyses on ACE tournament data (see, e.g., \cite{mellers2014strategies}). We compute \ensuremath{ABS_{\text{norm}}} only for forecasters who made forecasts on at least 10 different questions in a season, each accompanied by a rationale of at least 10 words. This restriction reduces the influence of forecast-level noise and ensures that the forecaster-level measure reflects repeated forecasting activity rather than isolated judgments \citep{karvetski2021ijf}.

We note that \ensuremath{ABS_{\text{norm}}} rewards updating forecasts, since it is calculated over a question's lifetime, while \ensuremath{BS_{\text{diff}}} does not capture the benefit of updating since it reflects only the first forecast.

\clearpage

\subsection{Prompt for Scoring Forecasting Rationales}
\label{appendix:NSPrompt}

An example prompt to score all 60 EQM patterns appears in the box below. It reproduces the content of the prompt sent to the model for one forecast-rationale pair; line breaks, boldface, and the itemized layout have been added for legibility, and long passages are truncated as marked. The complete set of 60 pattern names and definitions, which the prompt lists in full, is given in Appendix~\ref{appendix:NSPatterns}.

\begin{tcolorbox}[breakable, colback=white, colframe=black!55, colbacktitle=black!8,
  coltitle=black, fonttitle=\bfseries, boxrule=0.5pt, left=6pt, right=6pt, top=4pt, bottom=4pt,
  title={Example scoring prompt for one forecast-rationale pair}]
\small\raggedright
\setstretch{1.05}
\setlength{\parindent}{0pt}\setlength{\parskip}{5pt}

Below we have the following data components of a geopolitical forecast: (1)~a forecasting question, (2)~the specific answer options, (3)~the formal resolution criteria, (4)~the date and probability forecast of a forecaster, and (5)~the forecaster's rationale. We would like to score the rationale on a few dozen patterns, with scores for each pattern being \textbf{0} (no evidence of pattern present in rationale), \textbf{1} (there is some evidence of pattern present in rationale, but pattern incomplete or evidence is mixed), or \textbf{2} (clear, unambiguous evidence of pattern present in rationale). The pipe-delimited patterns (pattern name: description of how to identify each pattern) include:

\begin{description}[leftmargin=1.5em, font=\normalfont\bfseries, itemsep=3pt, parsep=0pt, topsep=2pt]
\item[Statistical Reasoning:] The forecaster looks to past occurrences or data and uses logical/statistical reasoning to inform his forecast. This could be looking at past occurrences, forming a comparison class and deriving a baserate, using basic statistics, or building a statistical model.
\item[Statistical Polling:] The forecaster incorporates insights from statistical polling or surveys to form his forecast.
\item[\normalfont\itshape $\vdots$] \textit{The remaining patterns are listed in the same name-then-description format; all 60 appear in Appendix~\ref{appendix:NSPatterns}.}
\item[Dispositional Factors:] The forecaster focuses on dispositional factors of key actors that make the event more or less likely.
\item[Grit:] The forecaster demonstrates either a passion for forecasting, persistence with a difficult task, or perseverance in the face of extreme uncertainty or shortcoming.
\end{description}

In the output, please return scores for all 60 patterns, with the pattern field using the exact title/case of the pattern name as above, and the score field should be a single number (0, 1, or 2), without any additional output. Also, return a one sentence description that explains the forecaster's reasoning for the question, including specific details and not just broad reasoning patterns.

Here are the input data:

\begin{description}[leftmargin=1.5em, font=\normalfont\bfseries, itemsep=4pt, parsep=0pt, topsep=2pt]
\item[(1) Forecasting question:] ``Will six-party talks with North Korea resume before 1 January 2014?''
\item[(2) Answer options:] If the U.S.\ does not announce beforehand that it is delaying the transfer of wartime operational command of the South Korean military to South Korea beyond December 1, 2015: (a)~Yes, (b)~No.
\item[(3) Formal resolution criteria:] ``In order for a `yes' realization to occur, the `condition' must be realized before the `outcome.' The six parties are North Korea, South Korea, the United States, Japan, China, and Russia. For this question to resolve in the affirmative, representatives of all six governments must meet simultaneously as part of an event officially described as a resumption of six-party talks. [additional detail truncated for publication] `Before' should be interpreted to mean at or prior to the end (23:59:59~ET) of the previous day.''
\item[(4) Forecast:] On 12/20/13, a forecaster gave a forecast of 2\%.
\item[(5) Rationale:] ``Time is rapidly running to actually have the talks. Also, the US has continued to make discussion of nukes part of the discussion and Kim has not signaled willingness to do that yet.''
\end{description}
\end{tcolorbox}

\clearpage

\subsection{EQM Pattern Set}
\label{appendix:NSPatterns}

\noindent This appendix presents the full descriptions and directional hypotheses for the 60 EQM patterns, presented in six families and corresponding tables.

\renewcommand{\thetable}{A\arabic{table}}
\setcounter{table}{0}

\begin{table}[H]
\footnotesize
\caption{Bias and Emotion EQM Patterns}
\label{tab:bias_and_narrative_reasoning}
\renewcommand{\arraystretch}{1.25}
\begin{tabularx}{\textwidth}{|>{\raggedright\arraybackslash}p{3cm}|X|>{\raggedright\arraybackslash}p{2cm}|}
\hline
\textbf{Pattern Name} & \textbf{Full Description} & \textbf{Association} \\
\hline
Clustering Illusion & The forecaster finds patterns in data using either minimal data points or suggests a pattern without reasonably testing for the pattern. Examples include assuming that recent or isolated events are part of a larger trend, using just a few data points to make a robust statistical argument, or eyeballing graphs without digging into the data to confirm. & Negative \\
Confirmation Bias & The forecaster seems to favor a hypothesis or outcome, and then presents some piece of evidence (e.g., news story, data point) that confirms the hypothesis or outcome without searching for evidence for other hypotheses or outcomes. & Negative \\
Extreme Confidence & The forecaster expresses a high degree of confidence in his forecast, stating the events "will" happen or "will not" happen, or that the outcome is nearly certain. & Negative \\
Speculative Terms & The forecaster speculates on what might happen using terms like "could" or "might". & Negative \\
Gut Based & The forecaster relies on instinctive or intuitive reasoning without citing evidence or data to support their conclusion. This includes statements like "I feel," "I sense," or "it seems to me," which suggest a forecast based on personal instinct rather than analysis or factual support. & Negative \\
Inside View & The forecaster analyzes only the dynamics of the current question, without taking an outside view and considering how past events or precedent could provide additional insight. & Negative \\
Personal Anecdote & The forecaster uses a personal anecdote to help inform his forecast. & Negative \\
Simplification Bias & The forecaster is over-simplifying a complex forecasting dynamic into a simplistic viewpoint. & Negative \\
Positive Emotion & The forecaster expresses positive emotion (e.g., excitement) with regard to the topic, question, or a particular outcome. & Negative \\
Negative Emotion & The forecaster expresses negative emotion (e.g., dread) with regard to the topic, question, or a particular outcome. & Negative \\
Political Preference & The forecaster expresses a political preference that relates to one of the question outcomes. & Negative \\
\hline
\multicolumn{3}{@{}p{\textwidth}@{}}{\scriptsize
\textit{Notes.} Rows correspond to individual EQM patterns. \emph{Pattern Name} provides the name of each EQM pattern. \emph{Full Description} provides the text-based definition that was provided to the LLM to score the pattern. \emph{Association} describes the hypothesized sign of the correlation between the pattern and accuracy (both \ensuremath{BS_{\text{diff}}} and \ensuremath{ABS_{\text{norm}}}). These patterns of the Bias and Emotion family capture influences that may distort probabilistic judgment and draw from the heuristics-and-biases literature \citep{kahneman1973prediction,tversky1974judgment}. Patterns such as \textit{Clustering Illusion}, \textit{Confirmation Bias}, and \textit{Simplification Bias} reflect failures in interpreting evidence or identifying statistical patterns \citep{fischhoff1977confidence,nickerson1998confirmation}. Other patterns capture intuition-driven reasoning, including \textit{Gut Based} reasoning, \textit{Personal Anecdote}, and the \textit{Inside View}, where forecasters analyze the current situation without reference to broader historical cases or statistical base rates \citep{kahneman1979, kahneman2011thinking}. Additional patterns capture affective framing in language, including \textit{Positive Emotion}, \textit{Negative Emotion}, and \textit{Political Preference}. Finally, \textit{Extreme Confidence} and \textit{Speculative Terms} describe patterns associated with overconfidence or unsupported conjecture \citep{moore2017confidence}. Across these patterns, stronger expressions of bias, emotional framing, or intuition-driven reasoning are hypothesized to be negatively associated with forecasting accuracy.
} \\
\hline
\end{tabularx}
\end{table}

\clearpage

\begin{table}[t!]
\footnotesize
\caption{Analytical Reasoning EQM Patterns}
\label{tab:analytic_and_heuristic_reasoning_tools}
\renewcommand{\arraystretch}{1.25}
\begin{tabularx}{\textwidth}{|>{\raggedright\arraybackslash}p{3cm}|X|>{\raggedright\arraybackslash}p{2cm}|}
\hline
\textbf{Pattern Name} & \textbf{Full Description} & \textbf{Association} \\
\hline
Analogies & The forecaster uses an analogy to another event or forecasting question to form his forecast. & Positive \\
Best Practices & The forecaster applies one of the following strategies: 1) analyzing past data trends, averages, or variations when predicting upcoming values, 2) considering the last occurrence of an event in cases where timing matters, or 3) using historical precedents while balancing arguments for potential differences in the current context. & Positive \\
Causal Reasoning & The forecaster identifies at least one current or future key driver that he believes will have an influence on the outcome of the question and clearly integrates the driver(s) into his forecast. Note, this is not speculative reasoning, but clear cause-and-effect type reasoning including detailed knowledge of system dynamics or known limits/boundaries on what is feasible given the question at hand. & Neutral \\
Domain Expertise & The forecaster demonstrates advanced domain or technical expertise that relates to the question. & Positive \\
Historic Expertise & The forecaster demonstrates advanced historical knowledge that relates to the question. & Positive \\
Fermi Breakdown & The forecaster breaks the main forecasting problem down into sub-problems, then solves each subproblem, and aggregates to reach a final forecast. & Positive \\
Scenario Reasoning & The forecaster presents multiple potential future outcomes, considering different scenarios that could impact the question’s resolution, indicating detailed, structured exploration of more than one plausible path. & Positive \\
Second Level Reasoning & The forecaster tries to reason beyond the most immediate or nearer term actions, and considers second-order effects such as the response to a potential action, or the response to the response of a potential action. & Positive \\
Statistical Causal Blend & The forecaster evaluates the forecast from a statistical perspective AND also considers at least one current or future causal driver when forming his forecast. & Positive \\
Statistical Polling & The forecaster incorporates insights from statistical polling or surveys to form his forecast. & Positive \\
Statistical Reasoning & The forecaster looks to past occurrences or data and uses logical/statistical reasoning to inform his forecast. This could be looking at past occurrences, forming a comparison class and deriving a baserate, using basic statistics, or building a statistical model. & Positive \\
Statistical Sensitivity Analysis & The forecaster includes some form of sensitivity analysis within his statistical reasoning. & Positive \\
Wildcards & The forecaster explicitly mentions rare but consequential events that would greatly change the expected outcome of the question. & Positive \\
\hline
\multicolumn{3}{@{}p{\textwidth}@{}}{\scriptsize
\textit{Notes.} Rows correspond to individual EQM patterns. \emph{Pattern Name} provides the name of each EQM pattern. \emph{Full Description} provides the text-based definition that was provided to the LLM to score the pattern. \emph{Association} describes the hypothesized sign of the correlation between the pattern and accuracy (both \ensuremath{BS_{\text{diff}}} and \ensuremath{ABS_{\text{norm}}}). These patterns of the Analytical Reasoning family capture structured analytical strategies used in forecasting. Several patterns reflect approaches commonly recommended in forecasting practice, including the use of comparison classes, base rates, and historical analogies \citep{tetlock2005expert,goldstein2009fast}. Patterns such as \textit{Analogies}, \textit{Best Practices}, \textit{Historic Expertise}, and \textit{Statistical Reasoning} reflect the use of empirical reference points or historical precedent when forming forecasts. Other patterns capture structured analytical decomposition or exploration of uncertainty. \textit{Fermi Breakdown} reflects breaking complex problems into smaller estimable components. \textit{Scenario Reasoning}, \textit{Second Level Reasoning}, and \textit{Wildcards} capture structured exploration of alternative futures and second-order effects, which stem largely from superforecasting best practices \citep{tetlock2015superforecasting}. To compare with \textit{Statistical Reasoning}, we included \textit{Causal Reasoning} and \textit{Statistical Causal Blend}, which capture the integration of statistical evidence with causal explanations or updating. Prior work suggests causal reasoning is often cognitively easier and therefore more prevalent than statistical reasoning, though not necessarily more accurate \citep{kahneman2011thinking}. Accordingly, \textit{Causal Reasoning} was assigned a neutral hypothesis, whereas \textit{Statistical Causal Blend} was hypothesized to be positively associated with forecasting accuracy.
} \\
\hline
\end{tabularx}
\end{table}

\clearpage

\begin{table}[t!]
\footnotesize
\caption{Alignment and Discipline EQM Patterns}
\label{tab:forecasting_mechanics_and_disciplined_practice}
\renewcommand{\arraystretch}{1.25}
\begin{tabularx}{\textwidth}{|>{\raggedright\arraybackslash}p{3cm}|X|>{\raggedright\arraybackslash}p{2cm}|}
\hline
\textbf{Pattern Name} & \textbf{Full Description} & \textbf{Association} \\
\hline
Forecast and Rationale Align & The expressed forecast and the forecaster's rationale align in two ways: 1) the outcome expressed in the rationale as most likely has the highest forecast probability, and 2) the level of uncertainty expressed in the rationale matches the magnitude of the forecast. & Positive \\
Forecast and Rationale Misalign & The expressed forecast and the forecaster's rationale are misaligned in that 1) the outcome expressed in the rationale as most likely does not have the highest forecast probability, or 2) the level of uncertainty expressed in the rationale differs from the magnitude of the forecast. & Negative \\
Forecaster Error & The forecaster commits an error in translating his rationale into a forecast (e.g., says small chance of event occurring, then gives large forecast, or vice versa). & Negative \\
Levels of Confidence & The forecaster makes an elaborate argument and expresses a level of confidence within each structural point of the argument, possibly describing features or assumptions of the argument where the forecaster is less certain. & Positive \\
Resolution Criteria Deep Dive & The forecaster takes a deep dive into the resolution criteria of the question, as a lawyer might, to determine exactly what may and may not count towards an affirmative resolution. & Positive \\
Timing & The forecaster mentions the question's resolution timeframe or the time left for the question to resolve, and integrates this timing into his forecast or decision to adjust his forecast. & Positive \\
Conditions to Update & The forecaster outlines future conditions that will trigger an update to his forecast. & Positive \\
Updating & The forecaster is updating or affirming a previous forecast. & Positive \\
Adjusting Down & The forecaster acknowledges a baseline probability (either derived from a data point, statistical means, or a forecast from a teammate) but chooses to adjust (for a factor OTHER than time) this baseline probability DOWN for his forecast to reflect a LESSER likelihood of the event. & Positive \\
Adjusting Up & The forecaster acknowledges a baseline probability (either derived from a data point, statistical means, or a forecast from a teammate) but chooses to adjust (for a factor OTHER than time) this baseline probability UP for his forecast to reflect a GREATER likelihood of the event. & Negative \\
\hline
\multicolumn{3}{@{}p{\textwidth}@{}}{\scriptsize
\textit{Notes.} Rows correspond to individual EQM patterns. \emph{Pattern Name} provides the name of each EQM pattern. \emph{Full Description} provides the text-based definition that was provided to the LLM to score the pattern. \emph{Association} describes the hypothesized sign of the correlation between the pattern and accuracy (both \ensuremath{BS_{\text{diff}}} and \ensuremath{ABS_{\text{norm}}}). These patterns of the Alignment and Discipline family capture disciplined forecasting mechanics and internal consistency between a forecaster's reasoning and stated probability judgment. Several patterns reflect good forecasting hygiene emphasized in tournament research, including frequent updating, calibration of confidence, attention to resolution criteria, and explicit consideration of timing \citep{mellers2014strategies,tetlock2015superforecasting}. In particular, prior work finds that forecasters who update their beliefs incrementally and responsively tend to achieve higher forecasting accuracy \citep{atanasov2020small}. Other patterns capture alignment between narrative reasoning and numerical forecasts. \textit{Forecast and Rationale Align} reflects internal coherence between the forecaster's explanation and the probability estimate, whereas \textit{Forecast and Rationale Misalign} and \textit{Forecaster Error} capture inconsistencies that may arise when translating reasoning into a probabilistic judgment. Finally, the \textit{Adjusting Up} and \textit{Adjusting Down} patterns capture directional adjustments from a baseline probability. Prior research on judgmental adjustment suggests upward adjustments are more prone to optimism bias and overreaction, whereas downward adjustments tend to be more cautious and better calibrated \citep{fildes2009effective}. Accordingly, \textit{Adjusting Down} was hypothesized to be positively associated with forecasting accuracy, while \textit{Adjusting Up} was hypothesized to be negatively associated.
} \\
\hline
\end{tabularx}
\end{table}

\clearpage

\begin{table}[t!]
\footnotesize
\caption{Evidential Sourcing EQM Patterns}
\label{tab:source_evaluation_and_information_handling}
\renewcommand{\arraystretch}{1.25}
\begin{tabularx}{\textwidth}{|>{\raggedright\arraybackslash}p{3cm}|X|>{\raggedright\arraybackslash}p{2cm}|}
\hline
\textbf{Pattern Name} & \textbf{Full Description} & \textbf{Association} \\
\hline
Authority Based & The forecaster bases his forecast on a statement from a perceived expert or authority. & Positive \\
News or Data Easily Accessible & The forecaster derives insight from a news story or data source that would likely appear on page 1 of a search engine (e.g., from the major news sources like NYTimes, abcnews, NPR, Reuters, etc.). & Positive \\
News or Data Remote & The forecaster derives insight from a news story or data source that would NOT appear on page 1 of a search engine, but rather requires some advanced searching/curation. & Positive \\
Source at Face Value & The forecaster uses information or data from a source without critical analysis, immediately incorporating it into their forecast. This includes uncritical references to news articles or data points followed by accepting it as true without questioning its reliability or potential biases. & Positive \\
Source Credibility and Reliability & The forecaster presents information or data from a source/news story and analyzes the source and conclusions in terms of reliability and credibility. & Positive \\
Teaming Agree & The forecaster is expressing agreement with a point made by a teammate forecaster. & Positive \\
Teaming Disagree & The forecaster is disagreeing with, expressing doubt or uncertainty about, or otherwise correcting a point made by a teammate forecaster. & Positive \\
Teaming Insight & The forecaster is drawing the majority of his insight from a teammate and using this as a basis for his forecast. & Positive \\

\hline
\multicolumn{3}{@{}p{\textwidth}@{}}{\scriptsize
\textit{Notes.} Rows correspond to individual EQM patterns. \emph{Pattern Name} provides the name of each EQM pattern. \emph{Full Description} provides the text-based definition that was provided to the LLM to score the pattern. \emph{Association} describes the hypothesized sign of the correlation between the pattern and accuracy (both \ensuremath{BS_{\text{diff}}} and \ensuremath{ABS_{\text{norm}}}). These patterns of the Evidential Sourcing family capture whether forecasters incorporate external information into their rationales, the kinds of sources they use, and whether those sources are evaluated critically. The general hypothesis for this family as it relates to accuracy is that using information is better than relying on unsupported assertion, so the Evidential Sourcing patterns were assigned a positive association. The distinction between \textit{Source at Face Value} and \textit{Source Credibility and Reliability} is a refinement to reflect whether information is simply incorporated or explicitly evaluated in terms of credibility and reliability \citep{karvetski2020jdm, mandel2023metainformational,kelly2025source}. \textit{Authority Based}, \textit{News or Data Easily Accessible}, and \textit{News or Data Remote} capture the type and accessibility of the information referenced. The teaming patterns capture whether forecasters simply agree with, challenge, or wholly incorporate and rely on teammates' reasoning in collaborative forecasting environments \citep{tetlock2015superforecasting, horowitz2019teams}.

} \\
\hline
\end{tabularx}
\end{table}

\clearpage

\begin{table}[t!]
\footnotesize
\caption{Integrative Reasoning and Cognitive Framing EQM Patterns}
\label{tab:cognitive_framing_and_language}
\renewcommand{\arraystretch}{1.25}
\begin{tabularx}{\textwidth}{|>{\raggedright\arraybackslash}p{3cm}|X|>{\raggedright\arraybackslash}p{2cm}|}
\hline
\textbf{Pattern Name} & \textbf{Full Description} & \textbf{Association} \\
\hline
Abstract Reasoning & The forecaster's reasoning is more abstract, concerned with the "why", is goal-oriented, distal, or transcends the particularities of the given situation. & Negative \\
Concrete Reasoning & The forecaster's reasoning is more concrete, concerned with the "how", is proximal, or contextualizes the situation. & Positive \\
Descriptive Action Verbs & The forecaster uses descriptive action verbs (physical or observable action performed by the subject) when describing the actions of key actors. & Positive \\
Interpretive Action Verbs & The forecaster uses interpretive action verbs (verbs convey subjective or inferred meaning rather than purely describing physical action) when describing the actions of key actors. & Neutral \\
State Verbs & The forecaster uses state verbs (a condition, state of being, emotion, or perception) when describing the actions or state of key actors. & Negative \\
Situational Factors & The forecaster focuses on situational factors that make the event more or less likely. & Positive \\
Dispositional Factors & The forecaster focuses on dispositional factors of key actors that make the event more or less likely. & Negative \\
Fact Based & The majority of the forecaster's rationale consists of fact-based statements, where each fact has verifiable details, references to actions or events, measurable data or outcomes, historical facts or timelines, or cited sources or reports. & Positive \\
FOG Based & The majority of the forecaster's rationale consists of fact-deficient, obfuscating generalities (FOG), where each FOG statement consists of emotional or evocative language, broad generalizations, vague attributions of intent or behavior, speculative language without supporting evidence, or unspecified references to urgency or importance. & Negative \\
Dialectical Reasoning & The forecaster acknowledges multiple viewpoints or factors that conflict in terms of their conclusion (using terms like "however", "but", "nevertheless", "although", "on the other hand", etc.). & Positive \\
Elaborative Reasoning & The forecaster acknowledges multiple viewpoints or factors that align to the same conclusion (using terms like "therefore", "moreover", etc.). & Positive \\
First Person & The forecaster writes at least some of his rationale in a first-person perspective, describing an action he is taking, expressing or implying the views as his own, or otherwise using first-person pronouns like "I", "I'll", "I'm", "me", "my", etc. & Positive \\
Third Person & The forecaster reasons about the relevant political parties, organizations, individuals, or other entities using third-person pronouns (e.g., "he will", "she did", "they might", "because of them", "their intention", "it", "this group", "these people", "those in power"). & Negative \\
\hline
\multicolumn{3}{@{}p{\textwidth}@{}}{\scriptsize
\textit{Notes.} Rows correspond to individual EQM patterns. \emph{Pattern Name} provides the name of each EQM pattern. \emph{Full Description} provides the text-based definition that was provided to the LLM to score the pattern. \emph{Association} describes the hypothesized sign of the correlation between the pattern and accuracy (both \ensuremath{BS_{\text{diff}}} and \ensuremath{ABS_{\text{norm}}}). These patterns of the Integrative Reasoning and Cognitive Framing family present several linguistic spectra. First, drawing on construal-level theory \citep{trope2010construal}, reasoning may range from \textit{Concrete Reasoning} (proximal, contextualized explanations focused on how events unfold) to \textit{Abstract Reasoning} (more distal explanations emphasizing why events occur or broader goals). Second, verbs describing actors’ behavior vary in abstraction, progressing from \textit{Descriptive Action Verbs} (observable actions) to \textit{Interpretive Action Verbs} (actions interpreted through inference) to \textit{State Verbs} (describing beliefs, intentions, or internal states). Third, explanations may emphasize \textit{Situational Factors} (contextual conditions affecting outcomes) or \textit{Dispositional Factors} (traits or intentions of actors). Greater concreteness and contextual proximity were hypothesized to be positively associated with accuracy, while greater abstraction was hypothesized to be negatively associated with accuracy, as reflected by the hypotheses. The \textit{Fact Based} versus \textit{FOG Based} pattern distinction captures whether reasoning relies primarily on verifiable details and evidence or on fact-deficient, obfuscating generalities. The \textit{Dialectical} and \textit{Elaborative} reasoning patterns capture whether forecasters acknowledge multiple perspectives that conflict or reinforce the same conclusion \citep{conway2008twoways, karvetski2021ijf}. These patterns are related to the concept of integrative complexity, but are intended only as preliminary indicators rather than a full implementation of the traditional seven-point integrative complexity scale. Prior work in forecasting rationales has also examined opposing perspectives including \textit{First Person} versus \textit{Third Person} language \citep{zong2020measuring,karvetski2021ijf}, finding the former to be associated with more accurate forecasters and the latter with less accurate forecasters.

} \\
\hline
\end{tabularx}
\end{table}

\clearpage

\begin{table}[t!]
\footnotesize
\caption{Attitude and Uncertainty Management EQM Patterns}
\label{tab:uncertainty_and_belief_updating}
\renewcommand{\arraystretch}{1.25}
\begin{tabularx}{\textwidth}{|>{\raggedright\arraybackslash}p{3cm}|X|>{\raggedright\arraybackslash}p{2cm}|}
\hline
\textbf{Pattern Name} & \textbf{Full Description} & \textbf{Association} \\
\hline
Actively Open Minded & The forecaster shows a willingness to seek out and reflect on contrary evidence and an openness to changing his mind in the face of contrary evidence. & Positive \\
Grit & The forecaster demonstrates either a passion for forecasting, persistence with a difficult task, or perseverance in the face of extreme uncertainty or shortcoming. & Positive \\
Guessing & The forecaster states that his forecast is primarily based on a guess. & Negative \\
Inevitable or Impossible & The forecaster expresses either inevitability or impossibility with regards to the outcome of the question. & Neutral \\
Wallowing in Uncertainty & The forecaster laments that the situation is too complex to give an accurate forecast. & Negative \\
\hline
\multicolumn{3}{@{}p{\textwidth}@{}}{\scriptsize
\textit{Notes.} Rows correspond to individual EQM patterns. \emph{Pattern Name} provides the name of each EQM pattern. \emph{Full Description} provides the text-based definition that was provided to the LLM to score the pattern. \emph{Association} describes the hypothesized sign of the correlation between the pattern and accuracy (both \ensuremath{BS_{\text{diff}}} and \ensuremath{ABS_{\text{norm}}}). These patterns of the Attitude and Uncertainty Management family capture forecasters' attitudes toward uncertainty, revision, and sustained engagement with difficult forecasting problems. \textit{Actively Open-Minded} reasoning draws on the broader literature on actively open-minded thinking, which emphasizes willingness to consider contrary evidence and revise one's views \citep{stanovich2000reasoning,baron2008thinking}. \textit{Grit} reflects persistence and sustained effort in the face of difficulty \citep{duckworth2016grit}. Both actively open-mindedness and grit are discussed as traits associated with stronger forecasters in \citet{tetlock2015superforecasting}. By contrast, \textit{Guessing} and \textit{Wallowing in Uncertainty} capture resignation or lack of analytic commitment and were hypothesized to be negatively associated with forecasting accuracy. \textit{Inevitable or Impossible} captures strong directional language about certainty and was assigned a neutral hypothesis because such language may reflect either justified confidence or overstatement.
} \\
\hline
\end{tabularx}
\end{table}
\clearpage

\subsection{Composite Scoring Details}\label{sec:composite_scoring}

\subsubsection{Within-dataset composite scoring}\label{subsec:within_sample_training}
This appendix provides further details to Section~\ref{subsec:predictive_modeling} in terms of how the EQM and Pre-LLM patterns were converted into a single composite score via LASSO regression. The composite is constructed as the LASSO-predicted value of accuracy. However, we refer to it as a composite score because we use it not only to predict accuracy but also as a standalone measure of explanation quality that can be compared with other metrics, such as human ratings.

For predicting \ensuremath{BS_{\text{diff}}} in Study 1, we randomly partitioned the $N = 368$ forecasting questions into five mutually exclusive, exhaustive, and approximately equal-sized sets. This was done to avoid data leakage across questions and minimize overfitting concerns. For predicting \ensuremath{ABS_{\text{norm}}}, 1,770 forecaster-season combinations were partitioned similarly into five mutually exclusive, exhaustive, and approximately equal-sized sets. 

We used a two-stage cross-validation procedure. First, in each iteration, one of the five partitions was held out and the remaining four partitions were retained for model fitting. Within the four retained partitions, we standardized the data according to the mean and standard deviations among the four partitions and further performed repeated random splits (also five times) into a training set (approximately 70\%) and a testing set (approximately 30\%) to tune the LASSO regularization parameter ($\lambda$). A grid search was conducted in each split, and the $\lambda$ that maximized the correlation between predicted and observed outcomes in the testing set was recorded. The median $\lambda$ across all five repetitions  with different training/testing sets was selected as the optimal $\lambda$ for the iteration, and this optimal $\lambda$ was used to fit a final LASSO model over the four retained partitions and then to predict the outcomes (whether \ensuremath{BS_{\text{diff}}} or \ensuremath{ABS_{\text{norm}}}) in the held-out partition. These predictions serve as our composite scores, providing a single metric of perceived quality that incorporates the input from all scores, whether from pre-LLM methods or EQMs. This process was repeated in total five times, each time holding out a different partition, resulting in five sets of held-out predictions for each school. Because the partitioning and splitting procedures were identical across the set of patterns, any differences in correlative performance can be attributed solely to the signal from the sets of predictor variables. 

\subsubsection{Across-dataset composite scoring}
\label{subsec:exploratory_training}

The LASSO training regime above seeks to avoid overfitting using partitions. But, even with partitioning to control for data leakage, it is still generating composite scores within the same sample. For exploratory analyses, we aim to test the composite LASSO models in both season-over-season fashion within the ACE dataset---where the earlier season is the training/testing data and the focal season the out-of-sample held-out data---and in an across-dataset fashion, where the model is fit on the ACE dataset and then applied to forecasting data from an entirely new study (e.g., as done in Study 4). For these applications, we used a similar $\lambda$ estimation procedure as above, by splitting the entirety of the in-sample training dataset into five partitions and finding the optimal $\lambda$ per partition as above. Rather than using the four partitions to make predictions on the fifth as done in-sample, the median among the five optimal $\lambda$ values across the five partitions was used when applying the model to the new dataset (whether the next season or an entirely new dataset). For example, when applying the model season-over-season, the first season is split into five partitions, and the optimal ($\lambda$) is found for each. Then the median ($\lambda$) is used to fit the LASSO on the entirety of the first season's data, and the fitted model is applied to make predictions with the second season's data to generate composite scores for the second season.

\clearpage

\section{Supplementary Results}
\label{appendix:supp_results}

\subsection{LASSO Coefficients}
\label{app:lasso_coefficients}

Tables \ref{tab:lasso_forecast_eqm}--\ref{tab:lasso_forecaster_prellm} report the non-zero coefficients from the Study 1 LASSO models used to construct the EQM and pre-LLM composite scores at the forecast and forecaster levels.

\begin{table}[H]
\centering
\begin{minipage}{0.92\textwidth}
\centering
\footnotesize
\caption{Study 1 forecast-level EQM LASSO coefficients}
\label{tab:lasso_forecast_eqm}
\renewcommand{\arraystretch}{1.1}
\setlength{\tabcolsep}{4pt}
\begin{tabular}{lc}
\toprule
\textbf{Pattern} & \textbf{Coefficient} \\
\midrule
Intercept & $-0.045$ \\
Extreme Confidence & $-0.019$ \\
Confirmation Bias & $-0.010$ \\
Speculative Terms & $-0.007$ \\
Simplification Bias & $-0.005$ \\
Forecast and Rationale Misalign & $-0.005$ \\
Descriptive Action Verbs & $-0.004$ \\
Causal Reasoning & $-0.003$ \\
Abstract Reasoning & $-0.003$ \\
Elaborative Reasoning & $-0.001$ \\
Updating & $-0.001$ \\
Gut Based & $-0.001$ \\
Statistical Reasoning & $0.001$ \\
Timing & $0.001$ \\
Teaming Insight & $0.001$ \\
State Verbs & $0.002$ \\
Teaming Agree & $0.003$ \\
Concrete Reasoning & $0.003$ \\
Inevitable or Impossible & $0.011$ \\
Forecast and Rationale Align & $0.023$ \\
\bottomrule
\end{tabular}
\vspace{0.5em}
\begin{flushleft}
\scriptsize
\textit{Notes.} \emph{Pattern} gives the EQM pattern name. \emph{Coefficient} gives the LASSO regression value. Only EQMs with nonzero coefficients are shown. Positive coefficients indicate EQMs associated with higher \ensuremath{BS_{\text{diff}}} values (more accurate), whereas negative coefficients indicate EQMs associated with lower \ensuremath{BS_{\text{diff}}} values (less accurate). The intercept is included for completeness.
\end{flushleft}

\end{minipage}
\end{table}

\begin{table}[H]
\centering
\begin{minipage}{0.92\textwidth}
\centering
\footnotesize
\caption{Study 1 forecast-level pre-LLM LASSO coefficients}
\label{tab:lasso_forecast_prellm}
\renewcommand{\arraystretch}{1.1}
\setlength{\tabcolsep}{4pt}
\begin{tabular}{lc}
\toprule
\textbf{Feature} & \textbf{Coefficient} \\
\midrule
Intercept & $-0.045$ \\
Clout & $-0.004$ \\
conflict & $-0.004$ \\
focusfuture & $-0.003$ \\
word\_cnt & $-0.002$ \\
Tone & $-0.002$ \\
auxverb & $-0.002$ \\
Authentic & $-0.001$ \\
ipron & $-0.001$ \\
cause & $-0.001$ \\
tentat & $-0.001$ \\
tone\_pos & $-0.001$ \\
ethnicity & $-0.001$ \\
money & $-0.001$ \\
illness & $-0.001$ \\
fulfill & $-0.001$ \\
space & $-0.001$ \\
feeling & $-0.001$ \\
focuspresent & $-0.001$ \\
discrep & $0.001$ \\
emo\_neg & $0.001$ \\
prosocial & $0.001$ \\
socrefs & $0.001$ \\
male & $0.001$ \\
lack & $0.001$ \\
Apostro & $0.001$ \\
adverb & $0.002$ \\
affiliation & $0.002$ \\
allure & $0.002$ \\
adj & $0.003$ \\
number & $0.004$ \\
CC\_RF & $0.005$ \\
negate & $0.006$ \\
\bottomrule
\end{tabular}
\vspace{0.5em}
\begin{flushleft}
\scriptsize
\textit{Notes.} \emph{Feature} describes the pre-LLM model name. \emph{Coefficient} gives the LASSO regression value. Only pre-LLM models with nonzero coefficients are shown. Positive coefficients indicate pre-LLM models associated with higher \ensuremath{BS_{\text{diff}}} values (more accurate), whereas negative coefficients indicate pre-LLM models associated with lower \ensuremath{BS_{\text{diff}}} values (less accurate). The intercept is included for completeness.
\end{flushleft}

\end{minipage}
\end{table}

\begin{table}[H]
\centering
\begin{minipage}{0.92\textwidth}
\centering
\footnotesize
\caption{Study 1 forecaster-level EQM LASSO coefficients}
\label{tab:lasso_forecaster_eqm}
\renewcommand{\arraystretch}{1.1}
\setlength{\tabcolsep}{4pt}
\begin{tabular}{lc}
\toprule
\textbf{Pattern} & \textbf{Coefficient} \\
\midrule
Simplification Bias & $-0.036$ \\
Forecast and Rationale Misalign & $-0.031$ \\
Inside View & $-0.027$ \\
FOG Based & $-0.019$ \\
Confirmation Bias & $-0.017$ \\
Abstract Reasoning & $-0.010$ \\
Causal Reasoning & $-0.010$ \\
Speculative Terms & $-0.008$ \\
Updating & $-0.001$ \\
Statistical Reasoning & $0.001$ \\
Teaming Insight & $0.001$ \\
Actively Open Minded & $0.003$ \\
First Person & $0.004$ \\
Statistical Polling & $0.004$ \\
Resolution Criteria Deep Dive & $0.007$ \\
Wallowing in Uncertainty & $0.011$ \\
Inevitable or Impossible & $0.014$ \\
Teaming Agree & $0.025$ \\
Timing & $0.032$ \\
Forecast and Rationale Align & $0.046$ \\
Intercept & $0.132$ \\
\bottomrule
\end{tabular}
\vspace{0.5em}
\begin{flushleft}
\scriptsize
\textit{Notes.} \emph{Pattern} gives the EQM pattern name. \emph{Coefficient} gives the LASSO regression value. Only EQMs with nonzero coefficients are shown. Positive coefficients indicate EQMs associated with higher \ensuremath{ABS_{\text{norm}}} values (more accurate), whereas negative coefficients indicate EQMs associated with lower \ensuremath{ABS_{\text{norm}}} values (less accurate). The intercept is included for completeness.
\end{flushleft}

\end{minipage}
\end{table}

\begin{table}[H]
\centering
\begin{minipage}{0.92\textwidth}
\centering
\footnotesize
\caption{Study 1 forecaster-level pre-LLM LASSO coefficients}
\label{tab:lasso_forecaster_prellm}
\renewcommand{\arraystretch}{1.1}
\setlength{\tabcolsep}{4pt}
\begin{tabular}{lc}
\toprule
\textbf{Feature} & \textbf{Coefficient} \\
\midrule
ethnicity & $-0.024$ \\
they & $-0.015$ \\
focusfuture & $-0.012$ \\
cause & $-0.010$ \\
Clout & $-0.005$ \\
feeling & $-0.004$ \\
politic & $-0.003$ \\
money & $-0.002$ \\
want & $-0.001$ \\
QMark & $0.002$ \\
DIAL & $0.004$ \\
comm & $0.004$ \\
time & $0.006$ \\
Conversation & $0.009$ \\
number & $0.009$ \\
prosocial & $0.010$ \\
allure & $0.012$ \\
AllPunc & $0.017$ \\
CC\_RF & $0.048$ \\
Intercept & $0.132$ \\
\bottomrule
\end{tabular}
\vspace{0.5em}
\begin{flushleft}
\scriptsize
\textit{Notes.} \emph{Feature} describes the pre-LLM model name. \emph{Coefficient} gives the LASSO regression value. Only pre-LLM models with nonzero coefficients are shown. Positive coefficients indicate pre-LLM models associated with higher \ensuremath{ABS_{\text{norm}}} values (more accurate), whereas negative coefficients indicate pre-LLM models associated with lower \ensuremath{ABS_{\text{norm}}} values (less accurate). The intercept is included for completeness.
\end{flushleft}

\end{minipage}
\end{table}

\clearpage

\subsection{OLS Coefficients}
\label{app:ols_coefficients}

Table \ref{tab:ols_eqm_coefficients} reports the corresponding OLS coefficients for the Study 1 EQM patterns. 

\begin{table}[H]
\centering
\begin{minipage}{0.92\textwidth}
\centering
\caption{Study 1 EQM OLS coefficients}
\label{tab:ols_eqm_coefficients}

{\fontsize{9}{7.2}\selectfont
\renewcommand{\arraystretch}{0.92}
\setlength{\tabcolsep}{3pt}
\begin{tabular}{lcc}
\toprule
\textbf{Pattern} & \textbf{Forecast Level} & \textbf{Forecaster Level} \\
\midrule
Statistical Reasoning & $0.000$ & $-0.001$ \\
Statistical Polling & $-0.002$ & $0.006$ \\
Causal Reasoning & $-0.009$ & $-0.065$ \\
Statistical Causal Blend & $0.002$ & $-0.001$ \\
Elaborative Reasoning & $-0.004$ & $-0.028$ \\
Dialectical Reasoning & $-0.001$ & $0.000$ \\
Timing & $0.002$ & $0.026$ \\
Teaming Agree & $0.004$ & $0.018$ \\
Teaming Insight & $0.001$ & $0.006$ \\
Updating & $-0.003$ & $-0.017$ \\
Adjusting Up & $-0.002$ & $0.002$ \\
Adjusting Down & $0.000$ & $0.006$ \\
Extreme Confidence & $-0.024$ & $-0.019$ \\
Wallowing in Uncertainty & $0.000$ & $0.019$ \\
Simplification Bias & $-0.005$ & $-0.043$ \\
Historic Expertise & $0.000$ & $0.000$ \\
Domain Expertise & $-0.001$ & $0.000$ \\
Fact Based & $0.003$ & $-0.020$ \\
FOG Based & $0.000$ & $-0.040$ \\
News or Data Easily Accessible & $0.006$ & $0.034$ \\
News or Data Remote & $0.000$ & $-0.010$ \\
Source at Face Value & $-0.008$ & $-0.028$ \\
Conditions to Update & $-0.002$ & $-0.009$ \\
Scenario Reasoning & $0.000$ & $0.007$ \\
Confirmation Bias & $-0.011$ & $-0.022$ \\
Clustering Illusion & $-0.001$ & $-0.005$ \\
Inside View & $-0.001$ & $-0.032$ \\
Authority Based & $0.002$ & $0.009$ \\
Negative Emotion & $0.000$ & $0.006$ \\
Speculative Terms & $-0.009$ & $-0.033$ \\
Inevitable or Impossible & $0.017$ & $0.046$ \\
Gut Based & $-0.004$ & $0.005$ \\
Guessing & $0.003$ & $0.007$ \\
Wildcards & $0.002$ & $0.009$ \\
Analogies & $0.001$ & $0.012$ \\
Forecast and Rationale Misalign & $-0.006$ & $-0.035$ \\
Forecast and Rationale Align & $0.025$ & $0.054$ \\
Resolution Criteria Deep Dive & $0.000$ & $0.016$ \\
Actively Open Minded & $0.001$ & $0.019$ \\
First Person & $0.002$ & $0.001$ \\
Third Person & $-0.001$ & $-0.012$ \\
Second Level Reasoning & $0.002$ & $0.016$ \\
Best Practices & $0.000$ & $-0.012$ \\
Abstract Reasoning & $-0.004$ & $-0.017$ \\
Concrete Reasoning & $0.007$ & $0.053$ \\
Descriptive Action Verbs & $-0.006$ & $-0.013$ \\
Interpretive Action Verbs & $0.000$ & $0.012$ \\
State Verbs & $0.005$ & $0.010$ \\
Situational Factors & $0.000$ & $0.022$ \\
Dispositional Factors & $0.004$ & $0.014$ \\
word\_cnt & $-0.005$ & $-0.026$ \\
\bottomrule
\end{tabular}
\vspace{0.5em}
\begin{flushleft}
\scriptsize
\textit{Notes.} \emph{Pattern} gives the EQM pattern name; \texttt{word\_cnt} is included as a non-EQM comparison row. \emph{Forecast Level} gives the OLS regression value for forecast-level accuracy (\ensuremath{BS_{\text{diff}}}). \emph{Forecaster Level} gives the OLS regression value for forecaster-level accuracy (\ensuremath{ABS_{\text{norm}}}). Positive coefficients indicate EQM patterns associated with higher accuracy, whereas negative coefficients indicate EQM patterns associated with lower accuracy. Variables were standardized prior to fitting.
\end{flushleft}
}
\end{minipage}
\end{table}

\clearpage

\subsection{Correlations Among Accuracy-Relevant EQM Patterns}
\label{appendix:pattern_correlation_plots}

\begin{figure}[t!]
  \centering
  \includegraphics[width=\textwidth]{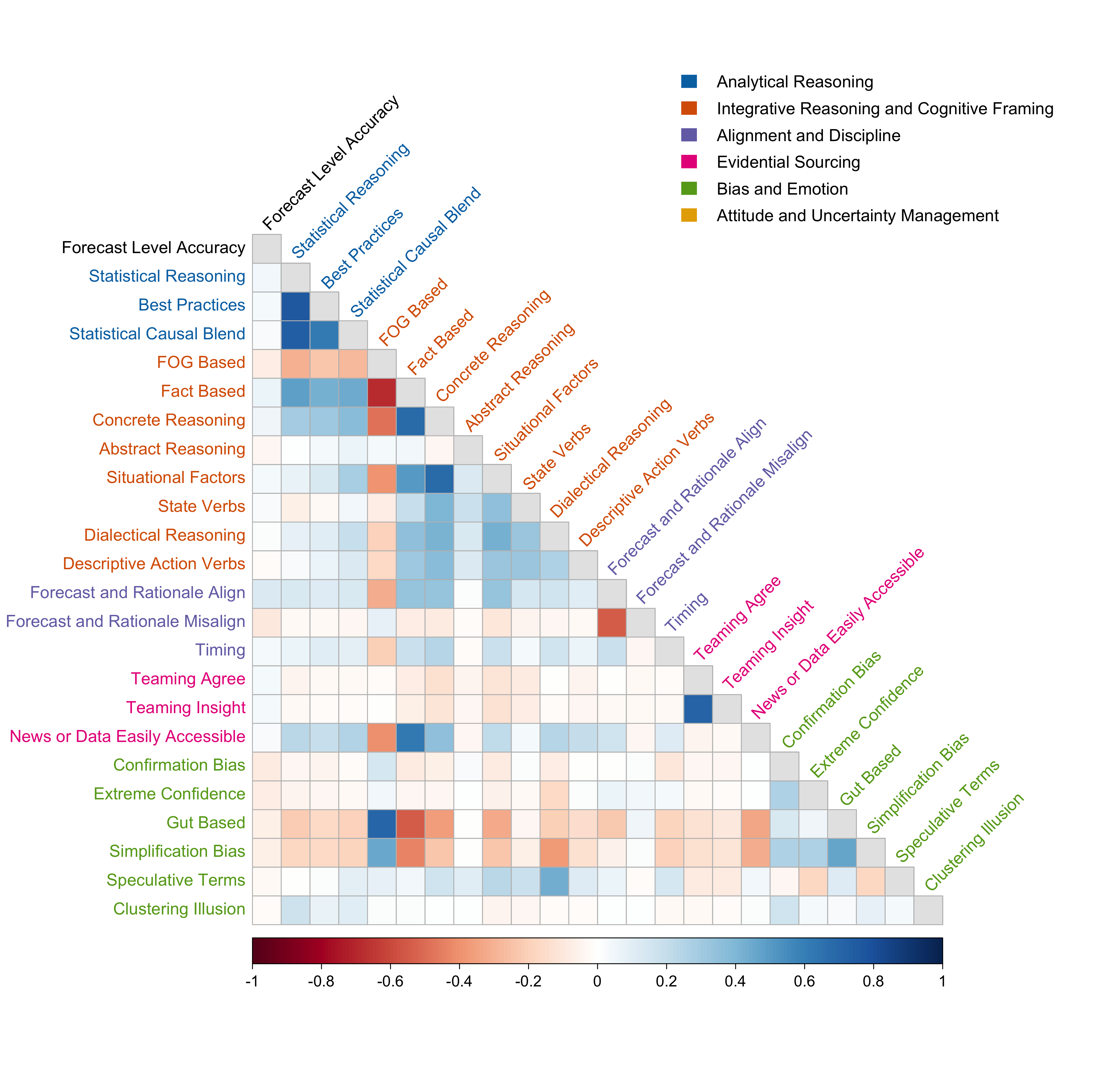}
  \caption{
    \emph{Correlations among accuracy-relevant EQM patterns at the forecast level}.
    Patterns are included if they had a positive or negative directional hypothesis and were negatively or positively significantly correlated with forecast-level accuracy, measured by \ensuremath{BS_{\text{diff}}}, at \(p < .001\) in hypothesized ways. The first column shows correlations between each EQM pattern and forecast-level accuracy (\ensuremath{BS_{\text{diff}}}). Remaining cells show pairwise correlations among retained EQM patterns. Colors indicate correlation direction and magnitude, with blue denoting positive correlations and red denoting negative correlations. Pattern names are colored by EQM family.
  }
  \label{fig:eqm_pattern_cors_forecast}
\end{figure}

\begin{figure}[t!]
  \centering
  \includegraphics[width=\textwidth]{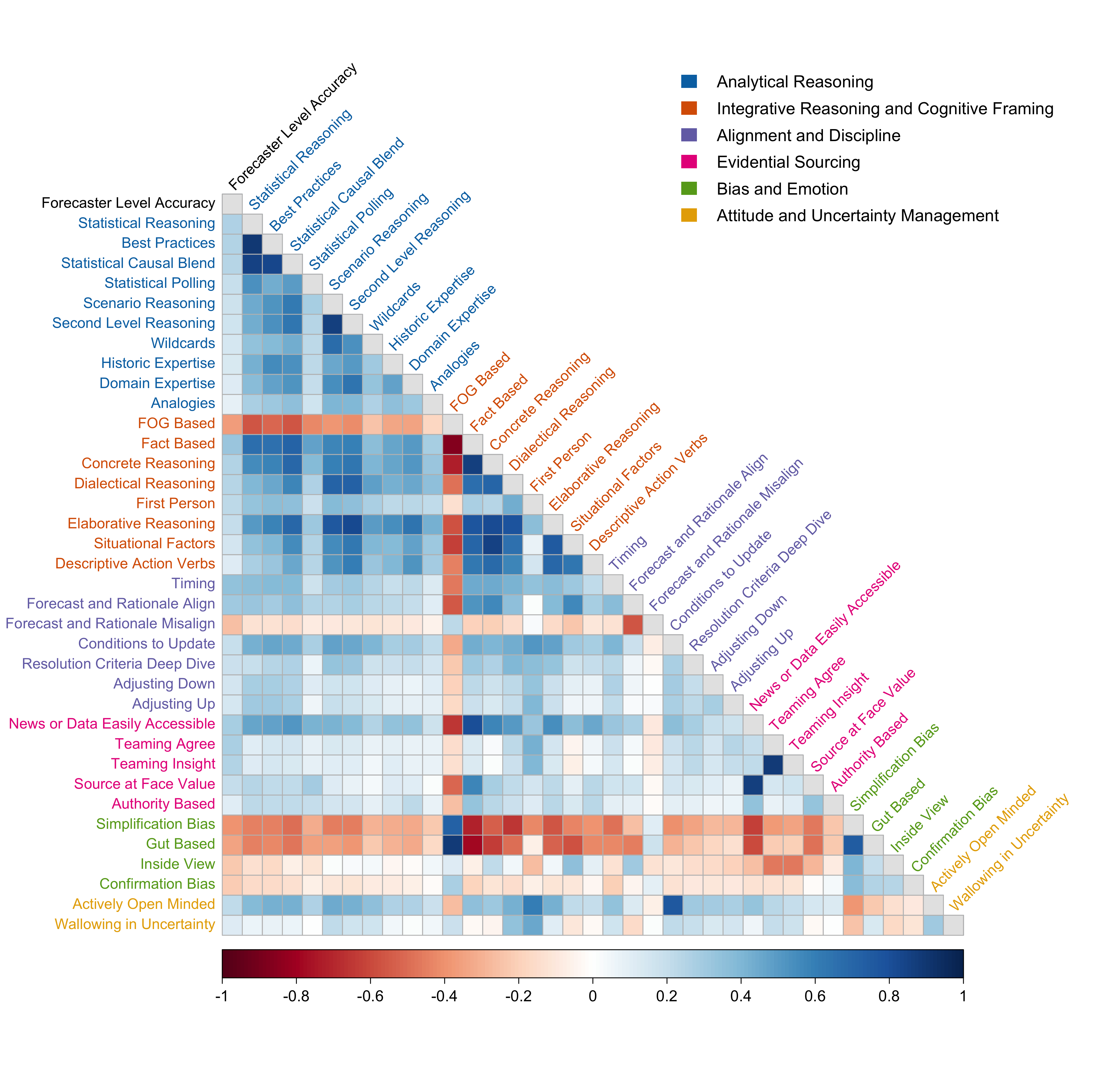}
  \caption{
    \emph{Correlations among accuracy-relevant EQM patterns at the forecaster level}.
    Patterns are included if they had a positive or negative directional hypothesis and were negatively or positively significantly correlated with forecaster-level accuracy, measured by \ensuremath{ABS_{\text{norm}}}, at \(p < .001\) in hypothesized ways. The first column shows correlations between each EQM pattern and forecaster-level accuracy (\ensuremath{ABS_{\text{norm}}}). Remaining cells show pairwise correlations among retained EQM patterns. Colors indicate correlation direction and magnitude, with blue denoting positive correlations and red denoting negative correlations. Pattern names are colored by EQM family.
  }
  \label{fig:eqm_pattern_cors_forecaster}
\end{figure}

In the main paper, the EQM patterns were used with LASSO regression to produce composite scores to correlate with accuracy, and the patterns were analyzed individually with accuracy in Section~\ref{subsec:hypothesis_testing}. In this appendix, their correlation with each other is presented. Figures~\ref{fig:eqm_pattern_cors_forecast} and~\ref{fig:eqm_pattern_cors_forecaster} show correlations among the EQM patterns that were negatively or positively significantly associated with accuracy at \(p < .001\) in the hypothesized ways. The forecast-level plot includes 23 patterns significantly associated with \(BS_{\text{diff}}\), while the forecaster-level plot includes 36 patterns significantly associated with \(ABS_{\text{norm}}\).

At the forecast level, several relationships aligned with the conceptual structure of the EQM system. \textit{Fact Based} and \textit{FOG Based} were negatively correlated, as predicted, reflecting the distinction between evidence-rich and fact-deficient rationales. The Integrative Reasoning and Cognitive Framing family also showed visible clustering, suggesting that these patterns captured related aspects of how forecasters framed and structured their explanations. Two bias-related patterns, \textit{Simplification Bias} and \textit{Gut Based}, were among the strongest negative correlates of forecast-level accuracy and were positively correlated with \textit{FOG Based}, consistent with the idea that low-evidence, intuition-driven, and oversimplified rationales tended to co-occur.

The forecaster-level plot showed similar structure, with stronger clustering across families. More analytical and evidence-based patterns tended to correlate positively with one another, whereas bias-related patterns tended to correlate negatively with these accuracy-enhancing patterns. Overall, these correlations behaved in theoretically expected ways, increasing confidence that the EQM patterns function as a coherent Gestalt system rather than as isolated markers.

\clearpage

\subsection{Reliability Analysis}
\label{appendix:reliability}

This appendix provides further details to the reliability analysis presented in Section~\ref{subsec:EQM_sensitivity}. We computed inter-model reliability between GPT-4o and Gemini~2{.}5~Pro for all EQM patterns across the set of 55,463 rationales. Reliability was evaluated using two complementary metrics. First, we calculated the Pearson correlation between the two models' scores (i.e., the 0, 1, or 2 outputs) for each pattern. Second, we also computed Cohen’s $\kappa$ using squared weighting for the corresponding score classifications. Interpreting these metrics jointly provides a more complete characterization of agreement as discussed in McHugh~\citeyearpar{mchugh2012kappa}.

\begin{figure}[t!]
    \centering
    \includegraphics[width=0.95\linewidth]{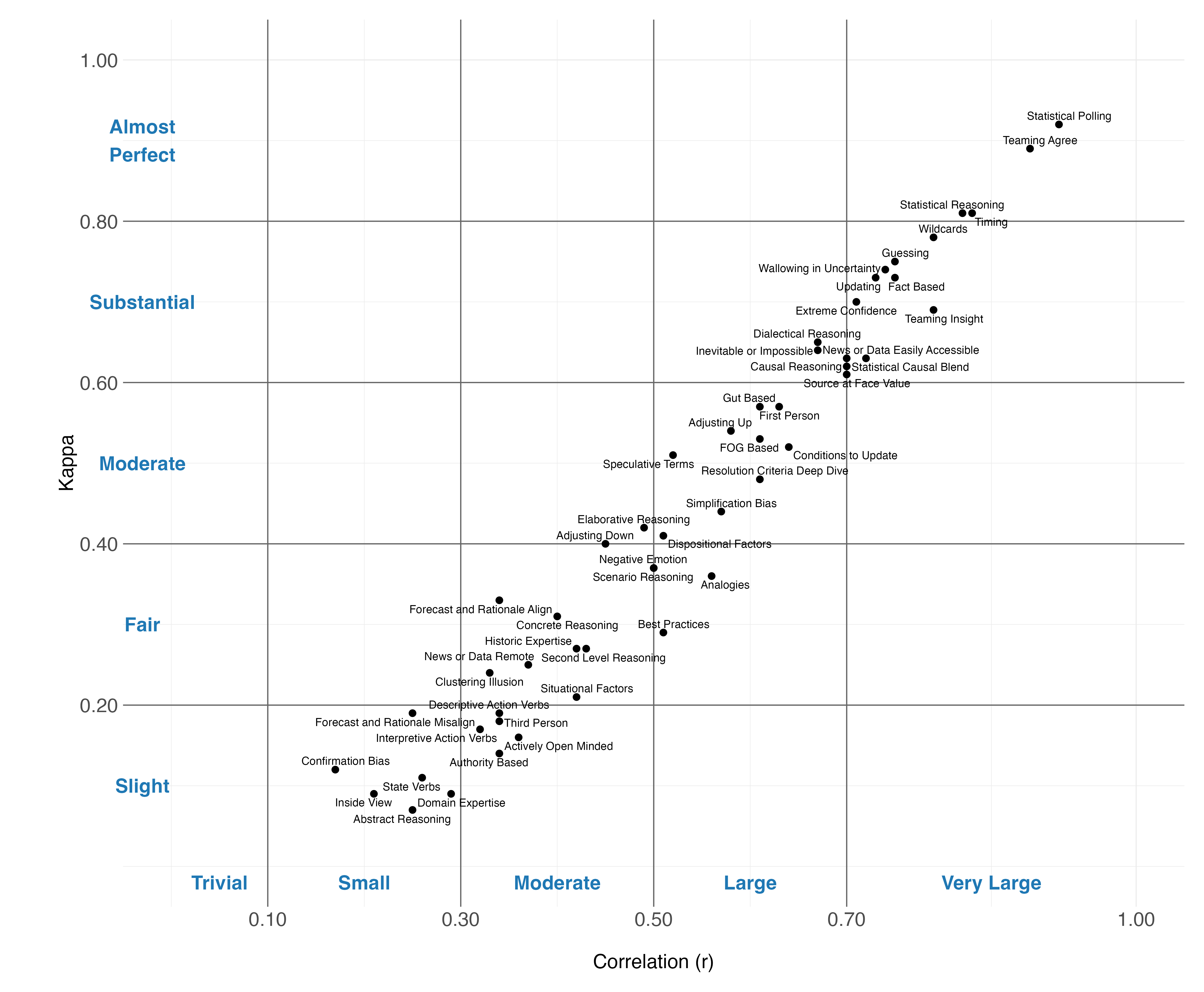}
    \caption{\emph{Inter-model reliability between GPT-4o and Gemini~2{.}5~Pro for each EQM pattern}. \emph{Horizontal axis}: Pearson correlation $r$. \emph{Vertical axis}: Cohen’s $\kappa$. Horizontal reference bands reflect the agreement categories of Landis and Koch~\citeyearpar{landis1977measurement}, and vertical bands reflect effect size conventions from Cohen~\citeyearpar{cohen1988statistical}.}
    \label{fig:intermodel_reliability}
\end{figure}

Figure~\ref{fig:intermodel_reliability} displays the $r-\kappa$ pairs for each individual EQM pattern.  We also included the conventional benchmarks in Cohen~\citeyearpar{cohen1988statistical} for effect sizes for \textit{small}, \textit{medium}, and \textit{large}, with additional distinctions for \textit{trivial} and \textit{very large}, and the qualitative agreement categories proposed in Landis and Koch~\citeyearpar{landis1977measurement} for $\kappa$. A subset of patterns---including \textit{Statistical Polling}, \textit{Teaming Agree}, and \textit{Statistical Reasoning}---exhibit both high correlations ($r > .70$) and substantial-to-almost-perfect agreement ($\kappa > .60$). These patterns appear to be recognized similarly by both models, suggesting strong cross-model stability. By contrast, other patterns show only slight agreement ($\kappa < .20$, $r < .30$), indicating lower categorical alignment.

Taken together, the results indicate that GPT-4o and Gemini~2{.}5~Pro exhibit broadly comparable but not interchangeable scoring behavior. This heterogeneity highlights the potential value of model-level triangulation and averaging of pattern scores when using the EQM pattern set, and these cross-model checks complement work on LLM-as-judge reliability and bias \citep{zheng2023judging,schroeder2025llmjudge}.

\clearpage

\subsection{Sensitivity Checks on Prompt Variations}
\label{appendix:sensitivity}

To assess the robustness of the EQM scoring methodology, we conducted a series of sensitivity checks by systematically altering the prompt provided to the LLM. Prior work documents that prompt design can substantially shift the distribution of LLM annotation outputs in social-science applications \citep{abraham2025prompt,atreja2025prompt}, and that LLM outputs can be sensitive to the order in which items are presented in the prompt \citep{halterman2026codebook,wang2024notfair,jiang2023cape}; the variations below probe related sensitivities in the EQM pipeline.
Each variation tested whether changes in prompt structure, content, or scoring format meaningfully altered the resulting pattern scores. All variations described here preserve the core batch-scoring design, in which all 60 patterns are scored in a single API call; the ablation that scores each pattern individually is reported separately in Appendix~\ref{subsec:single_pattern_scoring}.
We compared six variations against the baseline:

\begin{itemize}
    \item \texttt{baseline}: The original EQM prompt and scoring format described in Section~\ref{sec:pattern_scoring}.
\item \texttt{pattern\_shuffled}: The order of the 60 patterns within the prompt was randomized for each rationale.
\item \texttt{str\_enum}: The structured-output enum for the score was changed from numerical digits (0, 1, 2) to words (``zero'', ``one'', ``two'').
\item \texttt{no\_pattern}: The pattern name was removed from the structured-output response format. In the baseline, the LLM returns each pattern name alongside its score; in this variation, the LLM returns only the scores.
\item \texttt{no\_description}: The baseline prompt also asks the LLM to return a one-sentence description explaining the forecaster's reasoning. This description was removed in the \texttt{no\_description} setting.
\item \texttt{no\_pattern\_no\_desc}: Both \texttt{no\_pattern} and \texttt{no\_description} applied jointly.
\item \texttt{binary}: The prompt and response format were changed from a three-point scale (0, 1, 2) to a binary scale (0, 1). For comparison, scores from the other settings were collapsed by merging 1 and 2 into 1.
\end{itemize}

Table~\ref{tab:sensitivity_checks_gpt4} reports pairwise agreement across these settings for GPT-4o; Table~\ref{tab:sensitivity_checks_gemini} replicates the analysis for Gemini 2.5 Flash.

For GPT-4o (Table~\ref{tab:sensitivity_checks_gpt4}), the methodology was highly robust to superficial prompt variations. Altering the pattern order (\texttt{pattern\_shuffled}), changing the enum format (\texttt{str\_enum}), or removing the one-sentence description request (\texttt{no\_description}) all yielded high agreement with the \texttt{baseline} setting (89.5\% to 94.5\% identical scores). Robustness decreased modestly with more structural changes; removing the pattern name from the output format (\texttt{no\_pattern}) reduced identical agreement to 82.0\%.

Gemini 2.5 Flash (Table~\ref{tab:sensitivity_checks_gemini}) shows a flatter profile. On the structural \texttt{no\_pattern} change, Gemini retains 87.7\% identical agreement with the \texttt{baseline}, compared with 82.0\% for GPT-4o. On the superficial variations (\texttt{pattern\_shuffled}, \texttt{str\_enum}, \texttt{no\_description}), Gemini agreement stays within a narrow 86.0\%--88.0\% band, 3.5--6.8 percentage points below GPT-4o's 89.5\%--94.5\%. Gemini is thus more uniform across variations but peaks lower than GPT-4o. Together, these results indicate that the batch EQM scoring pipeline is stable against superficial prompt perturbations across both LLMs tested.

\begin{table}[t!]
\centering
\begin{minipage}{\textwidth}
\centering
\caption{Pairwise comparison of prompt variations (GPT-4o)}
\label{tab:sensitivity_checks_gpt4}
\footnotesize
\renewcommand{\arraystretch}{1.3}
\begin{tabularx}{\textwidth}{@{}l *{7}{>{\centering\arraybackslash}X}@{}}
\toprule
 & \rotatebox{60}{\scriptsize\textbf{\texttt{baseline}}} & \rotatebox{60}{\scriptsize\textbf{\texttt{no\_pattern}}} & \rotatebox{60}{\scriptsize\textbf{\texttt{pattern\_shuffled}}} & \rotatebox{60}{\scriptsize\textbf{\texttt{str\_enum}}} & \rotatebox{60}{\scriptsize\textbf{\texttt{no\_pattern\_no\_desc}}} & \rotatebox{60}{\scriptsize\textbf{\texttt{no\_description}}} & \rotatebox{60}{\scriptsize\textbf{\texttt{binary}}} \\
\midrule
\textbf{\texttt{baseline}} & & 82.0 (11.4 / 6.6) & 89.5 (9.4 / 1.0) & 93.5 (5.9 / 0.6) & 82.3 (11.2 / 6.6) & 94.5 (5.0 / 0.5) & 95.28 \\
\midrule
\textbf{\texttt{no\_pattern}} & & & 80.2 (13.0 / 6.9) & 83.0 (7.8 / 9.2) & 95.4 (1.3 / 3.2) & 83.8 (9.9 / 6.3) & 84.57 \\
\midrule
\textbf{\texttt{\makecell[l]{pattern\_\\shuffled}}} & & & & 87.7 (10.7 / 1.6) & 80.5 (12.8 / 6.7) & 89.8 (9.2 / 1.0) & 92.40 \\
\midrule
\textbf{\texttt{str\_enum}} & & & & & 83.1 (7.8 / 9.2) & 92.7 (6.2 / 1.1) & 95.13 \\
\midrule
\textbf{\texttt{\makecell[l]{no\_pattern\_\\no\_desc}}} & & & & & & 84.1 (9.8 / 6.2) & 84.57 \\
\midrule
\textbf{\texttt{\makecell[l]{no\_\\description}}} & & & & & & & 94.78 \\
\bottomrule
\end{tabularx}

\vspace{0.35em}
\scriptsize
\textit{Notes.} Each three-part entry shows agreement percentages between the row and column prompt variants, formatted as \textbf{Identical (Diff by 1 / Diff by 2)}. The \texttt{binary} column shows a single agreement percentage. The single-pattern scoring ablation is reported in Appendix~\ref{subsec:single_pattern_scoring}.
\end{minipage}
\end{table}

\clearpage

\begin{table}[t!]
\centering
\begin{minipage}{\textwidth}
\centering
\caption{Pairwise comparison of prompt variations (Gemini 2.5 Flash)}
\label{tab:sensitivity_checks_gemini}
\footnotesize
\renewcommand{\arraystretch}{1.3}
\begin{tabularx}{\textwidth}{@{}l *{7}{>{\centering\arraybackslash}X}@{}}
\toprule
 & \rotatebox{60}{\scriptsize\textbf{\texttt{baseline}}} & \rotatebox{60}{\scriptsize\textbf{\texttt{no\_pattern}}} & \rotatebox{60}{\scriptsize\textbf{\texttt{pattern\_shuffled}}} & \rotatebox{60}{\scriptsize\textbf{\texttt{str\_enum}}} & \rotatebox{60}{\scriptsize\textbf{\texttt{no\_pattern\_no\_desc}}} & \rotatebox{60}{\scriptsize\textbf{\texttt{no\_description}}} & \rotatebox{60}{\scriptsize\textbf{\texttt{binary}}} \\
\midrule
\textbf{\texttt{baseline}} & & 87.7 (9.8 / 2.6) & 86.0 (10.6 / 3.4) & 88.0 (10.0 / 2.1) & 86.9 (10.0 / 3.1) & 87.7 (10.0 / 2.3) & 92.76 \\
\midrule
\textbf{\texttt{no\_pattern}} & & & 85.7 (10.4 / 3.9) & 87.8 (9.8 / 2.4) & 86.7 (10.0 / 3.4) & 87.7 (9.7 / 2.5) & 92.31 \\
\midrule
\textbf{\texttt{\makecell[l]{pattern\_\\shuffled}}} & & & & 86.4 (10.2 / 3.4) & 85.8 (9.6 / 4.6) & 86.9 (9.6 / 3.6) & 90.82 \\
\midrule
\textbf{\texttt{str\_enum}} & & & & & 87.3 (9.8 / 3.0) & 88.4 (9.5 / 2.1) & 91.90 \\
\midrule
\textbf{\texttt{\makecell[l]{no\_pattern\_\\no\_desc}}} & & & & & & 88.4 (8.8 / 2.8) & 91.16 \\
\midrule
\textbf{\texttt{\makecell[l]{no\_\\description}}} & & & & & & & 92.13 \\
\bottomrule
\end{tabularx}

\vspace{0.35em}
\scriptsize
\textit{Notes.} Each three-part entry shows agreement percentages between the row and column prompt variants, formatted as \textbf{Identical (Diff by 1 / Diff by 2)}. The \texttt{binary} column shows a single agreement percentage. The single-pattern scoring ablation is reported in Appendix~\ref{subsec:single_pattern_scoring}.
\end{minipage}
\end{table}

\subsection{Ablation: Single-Pattern Scoring}\label{subsec:single_pattern_scoring}

As noted in Section~\ref{sec:EQMpatterns}, the EQM framework scores all 60 patterns jointly in a single API call, so that the co-presence of the full pattern set provides contrastive context for disambiguating related constructs. To ablate this design choice, we also scored each pattern individually in a separate API call (60 calls per rationale), which we denote the \texttt{single} setting.

Scoring the full ACE pool of $55{,}463$ rationales under \texttt{single} would have required roughly $3.3$ million API calls, compared with the $55{,}463$ calls used for the \texttt{baseline} pipeline (Section~\ref{sec:pattern_scoring}). To keep the ablation tractable, we drew a uniform random sample of $500$ rationales from the ACE pool and re-scored that sample under every prompt setting analyzed in this subsection, including the \texttt{baseline} and the Gemini variants. All counts and agreement rates reported here are computed on this common subset, so the settings are compared on the same rationales rather than on independently drawn samples.

Comparing \texttt{single} against \texttt{baseline} on this subset, overall agreement was 59.6\% identical scores for GPT-4o (with 26.4\% differing by 1 and 14.0\% differing by 2) and 75.0\% for Gemini 2.5 Flash (with 14.8\% differing by 1 and 10.2\% differing by 2). This divergence---the largest across all configurations tested---is consistent with the hypothesis that co-scoring related patterns provides disambiguating context (for example, the contrast between \textit{Fact Based} and \textit{FOG Based}), and that removing this context changes scoring behavior substantially. The shift was much larger for GPT-4o than for Gemini 2.5 Flash.

This finding is broadly consistent with prior observations that LLM scoring of one evaluation feature can benefit from co-presence of related feature definitions as auxiliary helpers \citep{fu2024gptscore, liu2024xeval}. Our setting differs in that all 60 EQM patterns are scored jointly as equal targets in a single zero-shot prompt rather than as helpers to a designated target pattern.

To investigate the divergence further, we inspected the per-pattern score distributions under \texttt{single} and identified three patterns whose single-pattern scoring was particularly anomalous: \textit{Forecast and Rationale Misalign}, \textit{Simplification Bias}, and \textit{Forecaster Error} (full descriptions in Appendix~\ref{appendix:NSPatterns}). Under \texttt{single}, GPT-4o provided scores of 2 on a large share of the 500 rationales for each pattern: 497 for \textit{Forecast and Rationale Misalign}, 341 (plus 153 scores of 1) for \textit{Simplification Bias}, and 473 for \textit{Forecaster Error}. This concentration did not appear under \texttt{baseline}; the anomaly was specific to single-pattern scoring. With the full pattern set, co-presence of related and contrastive patterns moderates these extremes. Single-pattern scoring is internally self-consistent but qualitatively different from batch scoring; batch context is therefore load-bearing in the EQM pipeline.

\clearpage

\subsection{Additional Details for Section~\ref{subsec:CaseStudy}}

This appendix provides additional details on the methods and results for Study 2. 

\subsubsection{Additional Study 2 Accuracy Scoring and Imputation Details}\label{appendix:Study2AccuracyImpute}

In Study 2, at the forecast level, accuracy was measured using \ensuremath{BS_{\text{diff}}}, as defined in Section~\ref{subsec:accuracy_measures}. At the forecaster level, we wanted to look season over season. In cases where a forecaster was new to a given season and lacking previous season forecasting data, or where previous season data was sparse, we decided to use a participation-adjusted seasonal average of \ensuremath{ABS_{\text{norm}}} so that zero retained a natural interpretation and could also serve as an imputation value when prior-season accuracy was unavailable. Specifically, for each forecaster-season, we first calculated \ensuremath{ABS_{\text{norm}}} across the questions on which the forecaster participated, and then multiplied that average by \(n/N\), where \(n\) is the number of binary questions forecast by that forecaster in that season and \(N\) is the total number of binary questions in that season. This construction treats unattempted questions as average performance. As a result, a forecaster with no participation in a given season receives a value of zero, and lightly participating forecasters are pulled toward zero rather than appearing artificially extreme.

For the behavioral measures, we computed forecaster-season averages for forecast frequency (average number of forecasts per question) and update size (average absolute difference between successive forecasts within a question). For cases where a forecaster made only a single forecast on a question, we imputed the value by modeling the relationship between forecast frequency and average update size, and extrapolating to the case where only a single forecast was made on a question\footnote{To impute the update size for questions with only a single forecast, we restricted to questions receiving at least two total forecasts (one update), but no more than five total forecasts (four updates)  yielding $N = 85{,}400$ question-forecaster pairs, and fit a regression with average update size as the dependent variable and forecast count as the predictor variable. The relationship was negative and statistically significant ($\beta_{1} = -.02$, $p < .001$, $\beta_{0} = .22$, $p < .001$), implying higher frequency counts tended to be correlated with smaller average update sizes. We used the regression to then predict the average update size for questions with only a single forecast, which yielded a value of 0.203.}. This preserves continuity at the lower bound for single-shot forecasters. When prior-season information was missing, prior-season forecaster accuracy was set to zero, prior-season forecast frequency to one, and prior-season update size to this regression-imputed lower-bound value.

For the forecaster-level EQM benchmark, we estimated the LASSO in season \(t-1\) using the participation-adjusted forecaster-season accuracy measure as the dependent variable and the averaged EQM profile as predictors, with the number of questions included as an additional predictor variable. The fitted coefficients from season \(t-1\) were then applied to the current-season forecaster profiles in season \(t\). Thus, the forecaster-level EQM composite score in the season-over-season analysis was an out-of-season projection based on the current season's averaged rationale profiles. When forecasters submitted forecasts in the current season but did not provide enough usable rationales to fully measure their EQM profile under the ten-word threshold, the missing EQM components were imputed to the corresponding current-season EQM mean values before scoring.

\subsubsection{Study 2 correlations with different thresholding}\label{appendix:ThresholdingCorrelations}

The main text reports Study 2 results under a moderate thresholding regime requiring 5 forecasts in the prior season and 5 in the current season with rationales of 10 or more words, which is viewed as the most practically relevant balance between coverage (e.g., number of forecasters) and measurement stability (e.g., forecasters not requiring imputation). Here we summarize the corresponding results under the none (0/0) and high (10/10) threshold regimes and compare them to the moderate case.

The none thresholding case (0/0) allows forecasters with no prior scored questions or very sparse current-season rationale data, and therefore requires the greatest degree of imputation. The high thresholding condition (10/10) restricts attention to forecasters with robust prior histories and current-season rationale coverage, thereby reducing noise but also narrowing the sample. The moderate condition (5/5), featured in the main text, represents a middle ground between these two extremes.

\begin{figure}[p]
\centering
\includegraphics[width=\textwidth]{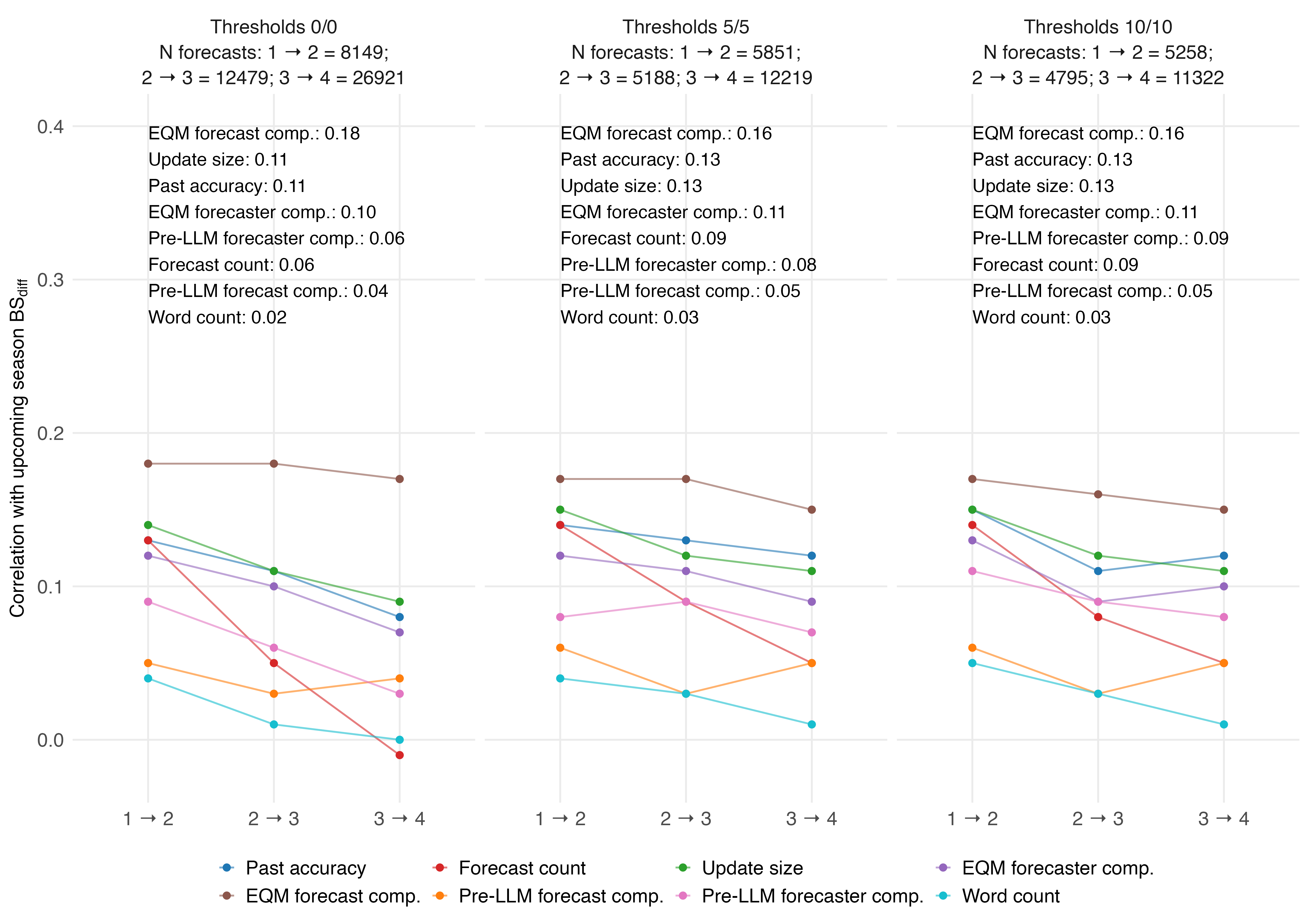}
\caption{\emph{Season-over-season correlations of indicators with forecast-level accuracy under alternative participation thresholds.} This figure extends Table~\ref{tab:sos_summary} by showing the same season-over-season correlations under three threshold choices. Each point reports the Pearson correlation between the indicated metric and current-season forecast-level accuracy, measured by \ensuremath{BS_{\text{diff}}}, for an adjacent season pair. \emph{Left panel}: no minimum threshold for prior-season scored questions or current-season forecasts with rationales of ten or more words. \emph{Middle panel}: the main specification used in Table~\ref{tab:sos_summary}, requiring at least five scored questions in the prior season and at least five current-season forecasts with rationales of ten or more words. \emph{Right panel}: a stricter specification requiring at least ten scored questions in the prior season and at least ten current-season forecasts with rationales of ten or more words. Panel subtitles report the number of forecast-level observations for each season transition. Text within each panel reports the mean correlation across the three season transitions for each indicator. Forecast-level indicators are computed for the individual forecast. Forecaster-level indicators are computed from prior-season data for past accuracy and from current-season data for forecast count, update size, EQM forecaster composite, and pre-LLM forecaster composite. For forecast-level composite scores, the LASSO model was fit on previous-season forecast-level data and applied to the current forecast's EQM pattern or pre-LLM feature scores. For forecaster-level composite scores, the LASSO model was fit on previous-season averaged forecaster-level data and applied to current-season averaged EQM pattern or pre-LLM feature scores.}
\label{fig:sos_forecast_accuracy_thresholds}
\end{figure}

At the forecast level, as presented in Figure~\ref{fig:sos_forecast_accuracy_thresholds}, the main substantive pattern is highly stable across thresholds. EQM forecast-level composite scores remain the strongest predictor across the none, moderate, and high regimes, including when compared with prior accuracy. The pre-LLM forecast- and forecaster-level composite scores remain among the weakest signals across all three thresholds, with only word count trailing. In other words, changing the threshold has relatively little effect on the substantive conclusion for individual forecasts: EQM is the strongest available signal, but the absolute magnitude of forecast-level prediction remains modest.

\begin{figure}[p]
\centering
\includegraphics[width=\textwidth]{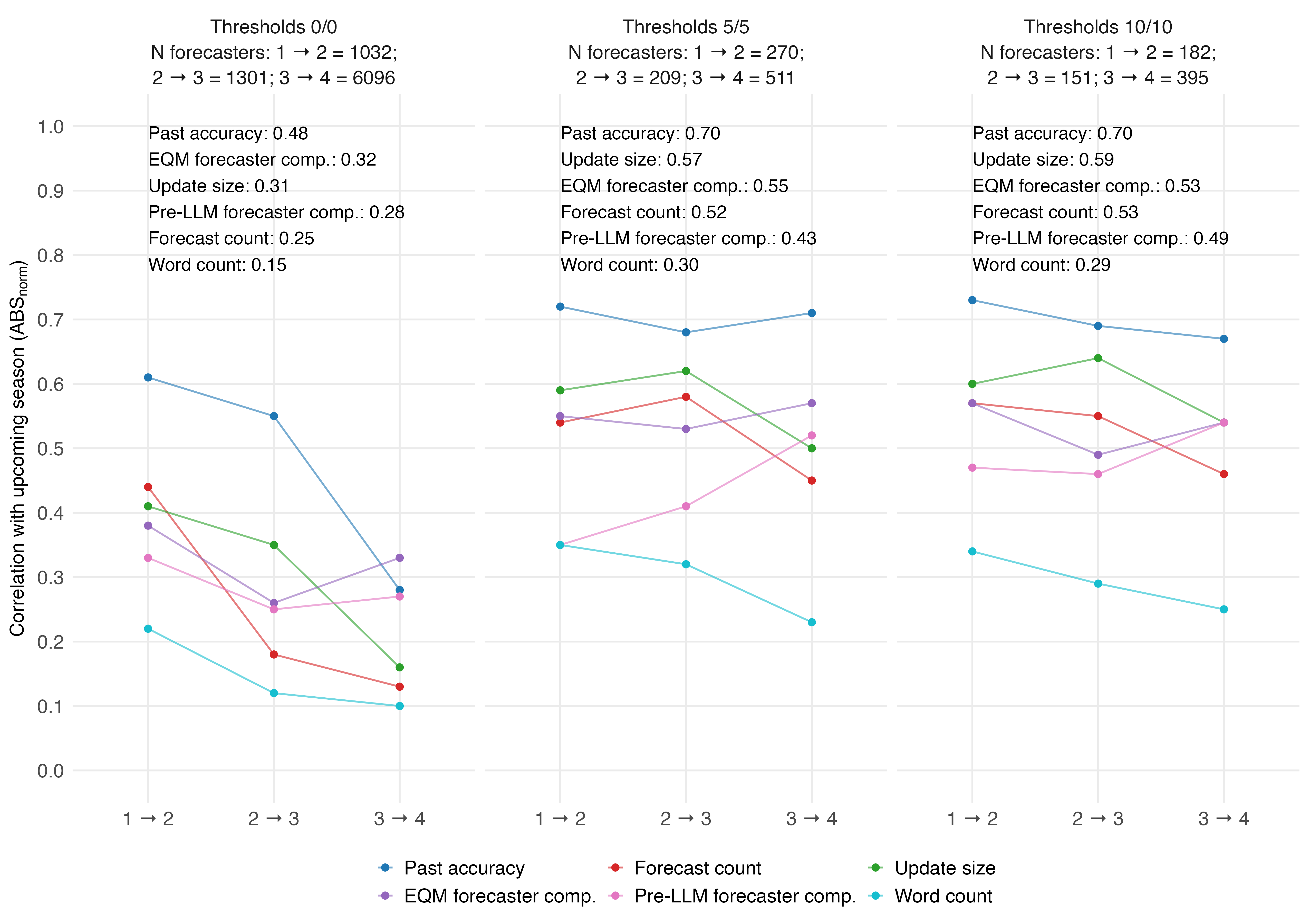}
\caption{\emph{Season-over-season correlations of indicators with forecaster-level accuracy under alternative participation thresholds.} This figure extends Table~\ref{tab:forecaster_sos_5_5} by showing the same season-over-season correlations under three threshold choices. Each point reports the Pearson correlation between the indicated metric and current-season participation-adjusted forecaster-level accuracy, measured by $ABS_{\text{norm}} \times (n/N)$, for an adjacent season pair. \emph{Left panel}: no minimum threshold for prior-season scored questions or current-season forecasts with rationales of ten or more words. \emph{Middle panel}: the main specification used in Table~\ref{tab:forecaster_sos_5_5}, requiring at least five scored questions in the prior season and at least five current-season forecasts with rationales of ten or more words. \emph{Right panel}: a stricter specification requiring at least ten scored questions in the prior season and at least ten current-season forecasts with rationales of ten or more words. Panel subtitles report the number of forecasters for each season transition. Text within each panel reports the mean correlation across the three season transitions for each indicator. Past accuracy is the forecaster's prior-season participation-adjusted normalized accuracy. Forecast count is the average number of forecasts per question in the current season. Update size is the average absolute size of forecast updates in the current season. EQM forecaster composite is the composite score derived from current-season EQM patterns averaged at the forecaster level. Pre-LLM forecaster composite is the composite score derived from current-season pre-LLM features averaged at the forecaster level. Word count is the average number of words in the forecaster's rationales. For the forecaster-level composite scores, the LASSO model was fit on previous-season averaged forecaster-level data and applied to current-season averaged EQM pattern or pre-LLM feature scores.}
\label{fig:sos_forecaster_accuracy_thresholds}
\end{figure}

At the forecaster level, as presented in Figure~\ref{fig:sos_forecaster_accuracy_thresholds}, thresholding matters more. Moving from none (0/0) to moderate (5/5) produces a substantial increase in correlations for the prior accuracy and rationale-based signals, while moving from moderate to high (10/10) produces comparatively little additional improvement. This pattern likely reflects the fact that the none condition requires imputing both past history and current-season profile information for sparse participants, whereas both the moderate and high conditions average over at least several questions and thereby reduce noise. Across all three thresholding regimes, prior accuracy remains the strongest predictor of current-season forecaster accuracy. EQM remains competitive with other behavioral indicators and continues to outperform the pre-LLM composite scores, but it does not exceed prior accuracy at the forecaster level.

\subsubsection{Study 2 aggregation results with different thresholding}\label{appendix:ThresholdingAggregations}

\begin{figure}[t!]
\centering
\includegraphics[width=\linewidth]{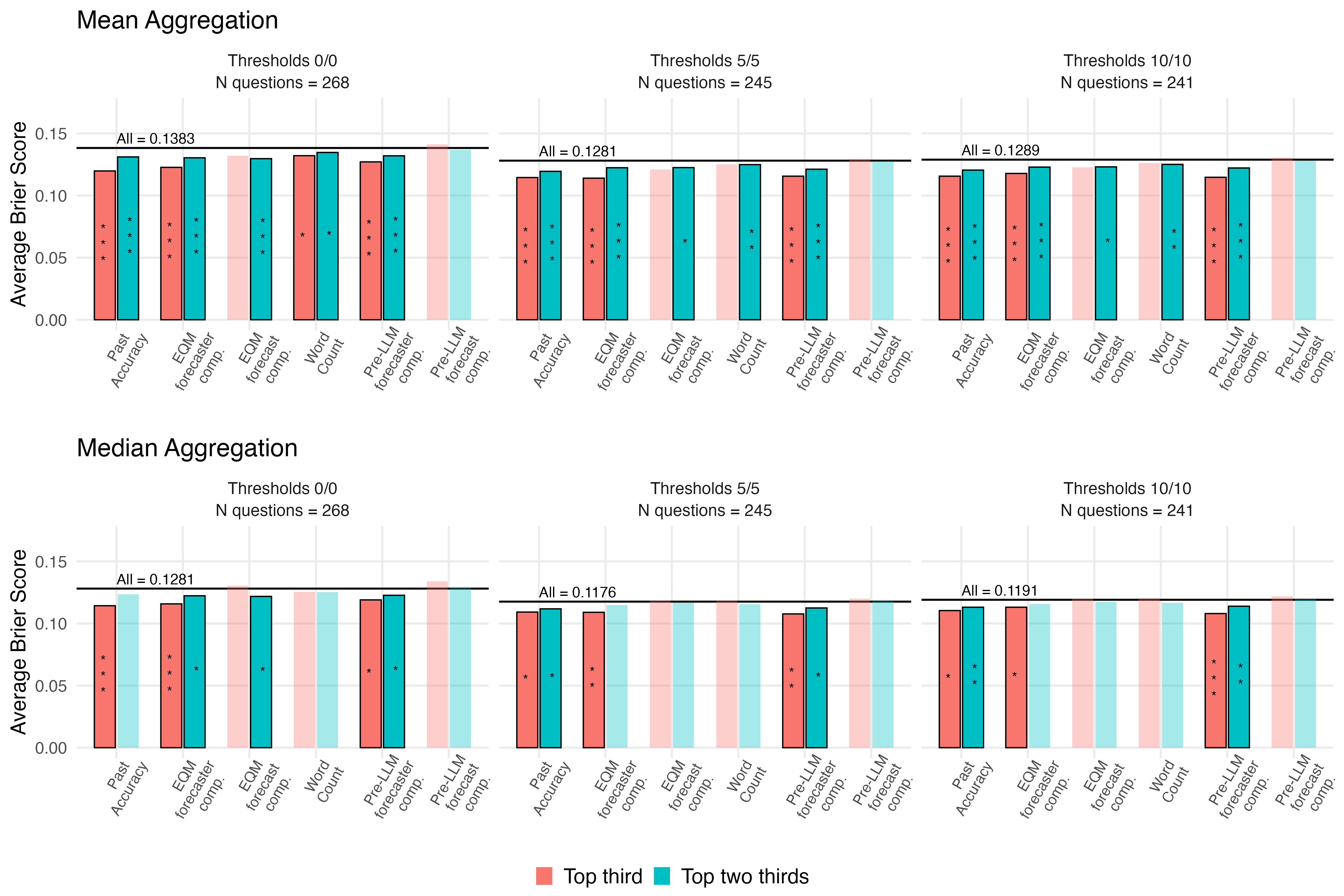}
\caption{\emph{Aggregate Brier scores by indicator, selection rule, aggregation method, and participation threshold.} This figure extends Figure~\ref{fig:aggregation_selection_combined} by showing the aggregation results under three threshold choices. \emph{Top row}: mean aggregation. \emph{Bottom row}: median aggregation. Within each row, \emph{left panel} shows the no-threshold specification (0/0), \emph{middle panel} shows the main specification (5/5), and \emph{right panel} shows the stricter specification (10/10), where the first number is the minimum number of scored questions in the prior season and the second number is the minimum number of current-season forecasts with rationales of ten or more words. Panel subtitles report the number of questions in each sample. Within each panel, the horizontal line labeled ``All'' marks the baseline Brier score obtained by aggregating all thirty forecasts on each question. Bars show the average Brier score after retaining either the top third (ten forecasts) or the top two-thirds (twenty forecasts) within each question, ranked by the indicator on the $x$-axis. Lower scores indicate better aggregate accuracy. Indicators include past accuracy, EQM forecaster composite, EQM forecast composite, word count, pre-LLM forecaster composite, and pre-LLM forecast composite. Asterisks indicate paired $t$-tests comparing each selection-rule aggregate with the all-forecast baseline: * $p < .05$, ** $p < .01$, *** $p < .001$. Bars significant at the 5\% level are shown at full opacity with black outlines; non-significant bars are shown with reduced opacity.}
\label{fig:aggregation_selection_appendix}
\end{figure}

The aggregation analysis shows a similar thresholding pattern, as presented in Figure~\ref{fig:aggregation_selection_appendix}. Moving from none to moderate generally improves aggregate accuracy, likely because more accurate forecasters also tend to be more active and because the moderate threshold removes many sparse, noisier contributors. Moving from moderate to high yields comparatively smaller changes. Median aggregation improves on the baseline aggregation using all thirty forecasts across thresholding regimes, likely by reducing the influence of outlier forecasts. The strongest gains typically come from rank-ordering by forecaster-level signals, especially past accuracy and forecaster-level EQM, rather than by forecast-level composite scores.

Forecast-level EQM reaches significance only in some thresholding and subsetting conditions. This mirrors the earlier finding that forecast-level EQM is especially useful for identifying weaker individual forecasts, but is less consistently effective for improving aggregate forecasts through rank-based selection. By contrast, forecaster-level EQM remains more useful for aggregation because it captures persistent contributor quality more effectively.

Taken together, the thresholding results reinforce the interpretation in the main text. The moderate thresholding regime captures the core substantive findings while avoiding much of the noise introduced by the none condition and without materially changing the conclusions relative to the high condition. At the forecast level, EQM remains the strongest signal across all three thresholds. At the forecaster level, prior accuracy remains dominant, with EQM functioning as a useful secondary signal. In the aggregation setting, forecaster-level ranking signals are consistently more useful than forecast-level ones, and the largest thresholding-related improvement occurs when moving from none to moderate rather than from moderate to high.

\clearpage

\subsection{LASSO Coefficients for Human Ratings}
\label{app:lasso_human_ratings}

Table~\ref{tab:lasso_human_ratings} reports the non-zero coefficients from the LASSO model predicting average human ratings for Section~\ref{subsec:HumansCor}.

\begin{table}[H]
\centering
\begin{minipage}{0.92\textwidth}
\centering
\footnotesize
\caption{LASSO coefficients predicting average human ratings}
\label{tab:lasso_human_ratings}
\renewcommand{\arraystretch}{1.1}
\setlength{\tabcolsep}{4pt}
\begin{tabular}{lc}
\toprule
\textbf{Pattern} & \textbf{Coefficient} \\
\midrule
FOG Based & $-0.025$ \\
Forecaster Error & $-0.004$ \\
Wildcards & $0.002$ \\
Second Level Reasoning & $0.005$ \\
Domain Expertise & $0.005$ \\
Elaborative Reasoning & $0.013$ \\
Timing & $0.016$ \\
Best Practices & $0.022$ \\
Forecast and Rationale Align & $0.026$ \\
Situational Factors & $0.027$ \\
Concrete Reasoning & $0.035$ \\
News or Data Easily Accessible & $0.083$ \\
Fact Based & $0.133$ \\
Square-root word count & $0.179$ \\
Intercept & $2.789$ \\
\bottomrule
\end{tabular}

\vspace{0.35em}
\scriptsize
\textit{Notes.} \emph{Pattern} gives the EQM pattern name; the square root of word count is included as a non-EQM comparison row. \emph{Coefficient} gives the LASSO regression value. Positive coefficients indicate patterns associated with higher average human ratings, whereas negative coefficients indicate patterns associated with lower average human ratings. The intercept is included for completeness. The resulting composite scores correlated with average human ratings at $r=.74$.
\end{minipage}
\end{table}

\clearpage

\subsection{Pattern Adaptations for the Team Dynamics Quantile Forecasts}
\label{app:td_pattern_adaptations}

The EQM pattern set used for the Team Dynamics study was identical to the ACE pattern set except for five patterns whose original descriptions referred directly to probability forecasts for binary outcomes. Because the Team Dynamics study elicited quantile forecasts rather than probability forecasts, these descriptions were minimally adapted to refer to baseline values, median forecasts, quantile width, and quantile-specific errors. Table~\ref{tab:td_pattern_adaptations} reports the exact ACE description and the corresponding Team Dynamics description for each adapted pattern.

\begin{table}[H]
\centering
\begin{minipage}{0.96\textwidth}
\centering
\footnotesize
\caption{EQM pattern-description changes for the Team Dynamics study}
\label{tab:td_pattern_adaptations}
\renewcommand{\arraystretch}{1.15}
\setlength{\tabcolsep}{4pt}
\begin{tabularx}{\textwidth}{l X X}
\toprule
\textbf{Pattern} & \textbf{ACE Description} & \textbf{Team Dynamics Description} \\
\midrule

Adjusting Down
& The forecaster acknowledges a baseline probability (either derived from a data point, statistical means, or a forecast from a teammate) but chooses to adjust (for a factor OTHER than time) this baseline probability DOWN for his forecast to reflect a LESSER likelihood of the event.
& The forecaster acknowledges a baseline value (either derived from a data point, statistical means, or a forecast from a teammate) but chooses to adjust this baseline value DOWN in providing his median forecast to account for a perceived downward trend. \\

Adjusting Up
& The forecaster acknowledges a baseline probability (either derived from a data point, statistical means, or a forecast from a teammate) but chooses to adjust (for a factor OTHER than time) this baseline probability UP for his forecast to reflect a GREATER likelihood of the event.
& The forecaster acknowledges a baseline value (either derived from a data point, statistical means, or a forecast from a teammate) but chooses to adjust this baseline value UP in providing his median forecast to account for a perceived upward trend. \\

Forecast and Rationale Align
& The expressed forecast and the forecaster's rationale align in two ways: 1) the outcome expressed in the rationale as most likely has the highest forecast probability, and 2) the level of uncertainty expressed in the rationale matches the magnitude of the forecast.
& The expressed forecast and the forecaster's rationale align in two ways: 1) the outcome expressed in the rationale as most likely matches the median forecast, and 2) the level of uncertainty expressed in the rationale matches the width of the quantiles. \\

Forecast and Rationale Misalign
& The expressed forecast and the forecaster's rationale are misaligned in that 1) the outcome expressed in the rationale as most likely does not have the highest forecast probability, or 2) the level of uncertainty expressed in the rationale differs from the magnitude of the forecast.
& The expressed forecast and the forecaster's rationale are misaligned in that 1) the outcome expressed in the rationale as most likely does not correspond to the median forecast, or 2) the level of uncertainty expressed in the rationale differs from the width of the quantiles. \\

Forecaster Error
& The forecaster commits an error in translating his rationale into a forecast (e.g., says small chance of event occurring, then gives large forecast, or vice versa).
& The forecaster commits an error in translating his rationale into a forecast (e.g., says small chance of resolution value being large, then gives large forecast, or vice versa; quantiles are not monotonic or units are clearly off/mismatched). \\

\bottomrule
\end{tabularx}

\vspace{0.35em}
\scriptsize
\textit{Notes.} Five EQM pattern descriptions shown in the table were changed for the Team Dynamics study. \emph{Pattern} provides the pattern name. \emph{ACE Description} provides the original pattern description used within the ACE tournament for binary forecasting questions. \emph{Team Dynamics Description} provides the adjusted pattern description used in Study 4 with the Team Dynamics quantile forecasts. The remaining EQM pattern names and descriptions were unchanged from the ACE implementation.
\end{minipage}
\end{table}

\clearpage

\section{Probability Training and EQM Reasoning Patterns}
\label{app:training-eqm}

The main analyses in this paper are correlational. This appendix takes a step toward a causal interpretation by exploiting the randomized probability-training intervention in the ACE tournament \citep{mellers2014strategies}.

{\bf Design.} During the ACE tournament, a subset of forecasters was randomly assigned to receive 
probability training. The training module covered reference-class reasoning, averaging across estimates and models, time series extrapolation, and avoidance of common judgmental traps such as overconfidence, confirmation bias, and base-rate neglect 
\citep{mellers2014strategies}. Because probability training was assigned at the forecaster level, we conduct the analysis at the forecaster level. We restricted the analysis to new-to-season forecasters in Seasons~2--4, comparing those assigned to probability 
training ($N = 214$) with those assigned to no training ($N = 199$). For each 
forecaster, we calculated season-level averages of their EQM pattern scores, rationale length, and forecaster-level accuracy ($\mathit{ABS}_{\text{norm}}$).

Trained forecasters wrote longer rationales on average (42.5 words vs.\ 29.0 words, 
$p < .001$, Cohen's $d = 0.40$). Because longer rationales mechanically create more opportunities 
for EQM patterns to be detected, all EQM models below control for average word count. 
Specifically, for each of the 50 retained EQM patterns, we estimated:
\begin{equation*}
  \text{EQM}_i = \beta_0 + \beta_1\,\text{Train}_i + \beta_2\,\text{Words}_i + \varepsilon_i,
\end{equation*}
where $i$ denotes a forecaster, $\text{EQM}_i$ is the standardized pattern score, $\text{Train}_i$ is a binary treatment indicator, $\text{Words}_i$ is the forecaster's average word count, and $\beta_1$ is the treatment effect of training, expressed in standard-deviation units. We corrected for multiple 
comparisons using the Benjamini--Hochberg false discovery rate procedure across all 
50 EQM patterns that were tested.\footnote{Because average word count may itself be affected by training, this specification should be interpreted as a word-count-adjusted contrast rather than a clean estimate of the total causal effect of training on EQM scores. The main qualitative result is not driven by this adjustment: when average word count is omitted from the regression, the same five EQMs remain significant after FDR correction---\textit{Statistical Reasoning}, \textit{Statistical Causal Blend}, \textit{Best Practices}, \textit{News or Data Easily Accessible}, and \textit{Fact Based}. Several additional patterns also become FDR-significant in the unadjusted specification, including \textit{Dialectical Reasoning}, \textit{Scenario Reasoning}, \textit{Elaborative Reasoning}, \textit{Source at Face Value}, \textit{Situational Factors}, \textit{Descriptive Action Verbs}, \textit{Statistical Polling}, \textit{Third Person}, and \textit{Conditions to Update}, as well as EQM patterns such as \textit{FOG Based}, \textit{Gut Based}, and \textit{Simplification Bias} that impact accuracy negatively.}

{\bf Accuracy results.} The accuracy effects of training were not statistically significant in our new-to-season forecaster-level sample. In Season~2, trained 
forecasters were directionally more accurate than untrained forecasters 
($\mathit{ABS}_{\text{norm}}$: $-0.07$ vs.\ $-0.46$; $d = 0.34$, $p = .15$), 
consistent with \citet{mellers2014strategies}. In Seasons~3 and~4, however, the 
differences were small and not statistically significant. Pooling across 
Seasons~2--4, the trained and untrained groups performed nearly identically 
($-0.12$ vs.\ $-0.13$; $d = 0.01$, $p = .89$).

\begin{figure}[H]
    \centering
    \includegraphics[width=0.95\textwidth]{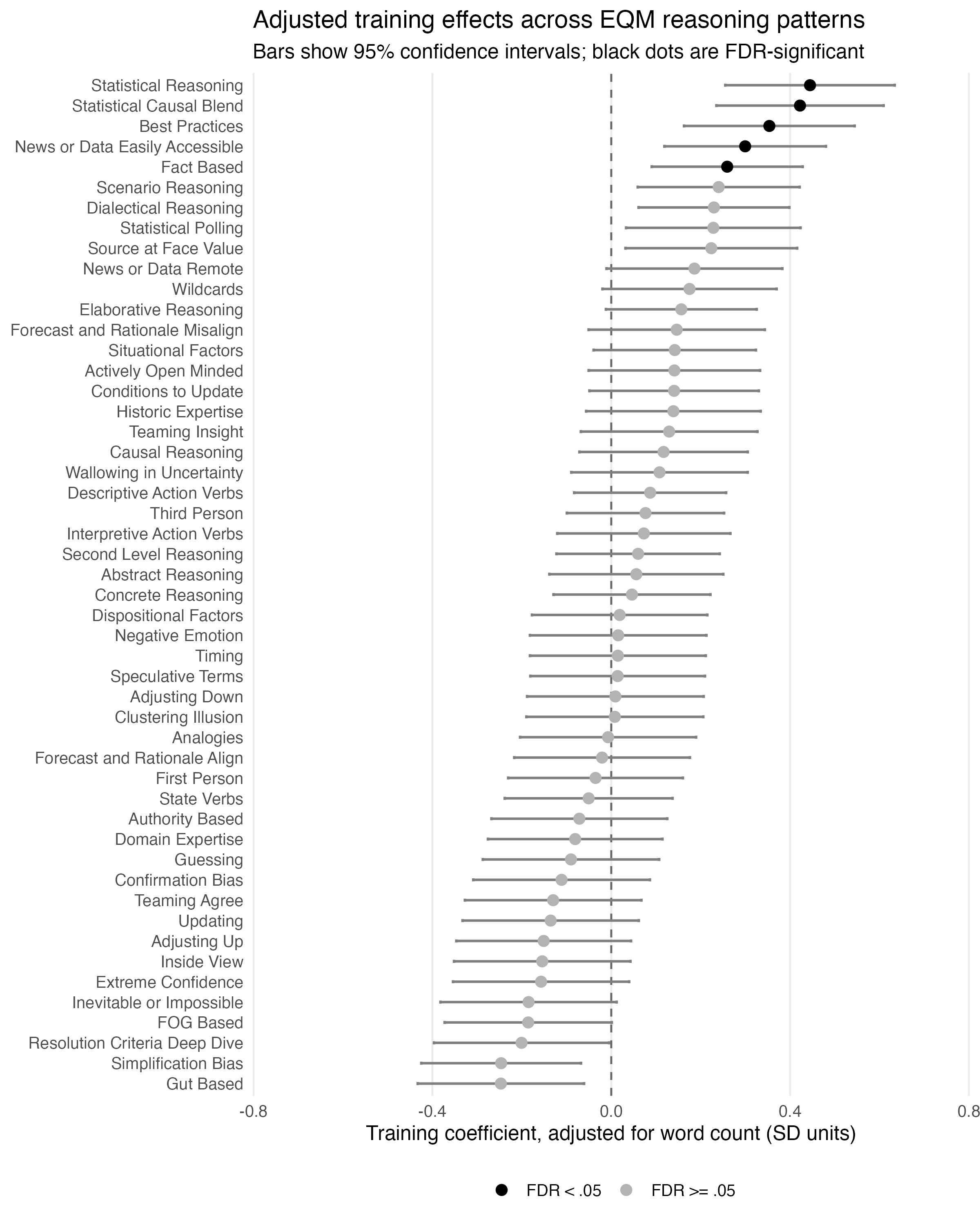}
    \caption{\emph{Adjusted training effects across EQM reasoning patterns}. Points show the estimated coefficient on probability training from separate forecaster-level models for each EQM pattern, controlling for average word count. Bars show 95\% confidence intervals. Black dots indicate effects that remain significant after Benjamini-Hochberg FDR correction.}
    \label{fig:training-eqm-effects}
\end{figure}

{\bf EQM results.} The estimated treatment effects of training on EQM patterns are shown in Figure~\ref{fig:training-eqm-effects}. Training had statistically significant positive 
effects, surviving FDR correction, on five patterns: \textit{Statistical Reasoning}, 
\textit{Statistical Causal Blend}, \textit{Best Practices}, \textit{News or Data 
Easily Accessible}, and \textit{Fact Based}. These are precisely the reasoning 
strategies emphasized in the training module. Training was also associated with lower 
scores on \textit{Gut Based} and \textit{Simplification Bias}, though these negative 
effects did not survive FDR correction.

Because training was randomly assigned, these results provide evidence that EQMs 
respond to exogenous changes in reasoning practices. This finding also assuages the concern that EQM patterns are merely proxies for forecaster engagement or effort. The pattern shifts are also 
substantively interpretable: the training taught specific analytical techniques, and 
EQMs detected exactly those techniques. That said, training did not reliably improve 
accuracy in this sample. A properly pre-specified 
experiment linking training, EQM scores, and accuracy would be a valuable future research project.

\clearpage

\bibliographystyle{plainnat}
\bibliography{sn-bibliography}
\end{document}